\documentclass[journal]{IEEEtran}

\usepackage{amsmath,amssymb,amsfonts}
\usepackage{algorithmic}
\usepackage{graphicx}
\usepackage{textcomp}
\def\BibTeX{{\rm B\kern-.05em{\sc i\kern-.025em b}\kern-.08em
    T\kern-.1667em\lower.7ex\hbox{E}\kern-.125emX}}

\usepackage{multicol}

\usepackage{subcaption}
\usepackage{caption,setspace}
\usepackage{float}
\usepackage{url}
\usepackage[colorlinks=false,
            allbordercolors={0 0 0}, 
            pdfborderstyle={/S/U/W 1},
            citebordercolor={1 1 1},
            linkbordercolor={1 1 1}]{hyperref}
            
\usepackage{fancyhdr}
\usepackage{cite}

\begin{document}

\title{EvalAttAI: A Holistic Approach to Evaluating Attribution Maps in Robust and Non-Robust Models}

\author{
Ian~E.~Nielsen$^{1}$
Ravi~P.~Ramachandran$^{1}$
Nidhal~Bouaynaya$^{1}$
Hassan~M.~Fathallah-Shaykh$^{2}$
Ghulam~Rasool$^{3}$
}

\maketitle

\renewcommand{\thefootnote}{\space}
\footnotetext[1]{$^{1}$
Rowan University, 201 Mullica Hill Rd, Glassboro, NJ 08028 USA (e-mails: \{nielseni6, bouaynaya, ravi\}@rowan.edu)}
\footnotetext[2]{$^{2}$
University of Alabama at Birmingham, 1720 2nd Ave S, Birmingham, AL 35294 USA (e-mail: hfshaykh@uabmc.edu)}
\footnotetext[3]{$^{3}$
Machine Learning Department of Moffitt Cancer Center and Research Institute, 12902 USF Magnolia Drive, Tampa, FL 33612 USA (email: Ghulam.Rasool@moffitt.org)
}
\footnotetext[4]{This work was partly supported by the National Science Foundation awards ECCS-1903466, OAC-2008690, and OAC-2234836. I. E. Nielsen is supported by the US Department of Education through a Graduate Assistance in Areas of National Need (GAANN) program Award Number P200A180055.}

\pagestyle{fancy} 
\fancyhead[C]{Nielsen et al.: EvalAttAI: A Holistic Approach to Evaluating Attribution Maps in Robust and Non-Robust Models}



\begin{abstract}
The expansion of explainable artificial intelligence as a field of research has generated numerous methods of visualizing and understanding the black box of a machine learning model. Attribution maps are generally used to highlight the parts of the input image that influence the model to make a specific decision.  On the other hand, the robustness of machine learning models to natural noise and adversarial attacks is also being actively explored. This paper focuses on evaluating methods of attribution mapping to find whether robust neural networks are more explainable. We explore this problem within the application of classification for medical imaging. Explainability research is at an impasse. There are many methods of attribution mapping, but no current consensus on how to evaluate them and determine the ones that are the best. Our experiments on multiple datasets (natural and medical imaging) and various attribution methods reveal that two popular evaluation metrics, Deletion and Insertion, have inherent limitations and yield contradictory results. We propose a new explainability faithfulness metric (called EvalAttAI) that addresses the limitations of prior metrics. Using our novel evaluation, we found that Bayesian deep neural networks using the Variational Density Propagation technique were consistently more explainable when used with the best performing attribution method, the Vanilla Gradient. However, in general, various types of robust neural networks may not be more explainable, despite these models producing more visually plausible attribution maps.

\end{abstract}




\section{Introduction}
\IEEEPARstart{W}{ith} the recent explosion of black box machine learning models over the past few years, there has been a great demand to explain how these models work so that users can understand and trust the results \cite{confalonieri2021historical, minh2022explainable, epifano2020towards}. This is especially important in mission-critical and life-saving applications, such as clinical diagnosis and medical decision-making \cite{zhang2022applications, van2022explainable, ahmed2022failure}. Explainable AI (XAI) seeks to give a user more trust in the model explanation \cite{nielsen2022robust}. The problem is that we do not know if we can trust the explanations either. In order for XAI to be viable for real world applications there must be trust, thus there have been many attempts to evaluate these explanations \cite{hooker2019benchmark, nourelahi2022explainable, mamalakis2022investigating, hama2022deletion, zhou2021evaluating, phan2022deepface, nielsen2022robust}. 

These application areas also demand robust machine learning models that resist natural noise in the input data and malicious or adversarial attacks \cite{dera2021premium, carannante2021trustworthy, Dera:MLSP2019}. In real-world tasks, the data can often come with noise and artifacts that the machine learning model might not have previously seen \cite{Dera:IEEERadar2019}. There might even be nefarious actors trying to confuse the model with adversarial examples \cite{dera2021premium, waqas2021exploring}. This requires that models be built and trained to resist noise and attacks \cite{waqas2021brain, carannante2021trustworthy}. There are several approaches to building robust machine learning models, including training on noisy datasets and Bayesian models \cite{dera2021premium, Dera:MLSP2019, Dera:IEEERadar2019, waqas2021exploring, ahmed2022failure, carannante2021trustworthy}. The first approach only focuses on model training and data processing, while the second alters the neural network architecture and introduces probability distribution functions over learnable parameters \cite{dera2021premium}. Robust models have been shown to produce more visually plausible explanations \cite{nourelahi2022explainable, tsipras2019robustness}. However, the quantitative evaluation of the faithfulness of these explanations is a challenging task. There are many criteria for evaluating explanations. In this paper, we propose a much more direct approach (called EvalAttAI) that eliminates the errors introduced by existing approaches to evaluating the faithfulness of explanations. The contributions in this paper are as follows:
\begin{itemize}
    \item We develop a new method of evaluating the faithfulness of an attribution map, which is much different from existing approaches and avoids the errors present in current methods. Our proposed method is denoted as Evaluating Attributions by Adding Incrementally (EvalAttAI).
    \item We compare our method to the state-of-the-art in the literature and show why ours is a more fair and accurate measure of faithfulness of model explanations. 
    \item We relate the concepts of faithfulness and robustness by showing whether robust models produce more faithful explanations when evaluated using our proposed EvalAttAI method.
\end{itemize}

The article is organized as follows. In Section II, we provide definitions of various terms that are used in the article and present various explainability faithfulness evaluation methods, including our newly proposed, EvalAttAI. Section III provides an overview of our methods and simulation experiments. In Section IV, we present results and discuss the same in Section V before concluding in Section VI.

\section{Explainability Metrics and Definitions} 

To evaluate the explainability method defined for a machine learning model, quantify its faithfulness, and link it to the robustness of the machine learning model, we must understand different concepts, terms, and metrics that are used in the literature. This section provides an overview of these important concepts.

\subsection{Robustness}

A model is said to be robust if its performance shows little decrease when the distribution of the test data is different from that used during the training. A model can be robustified (made more robust) by training it on a large variety of data that includes various types of noisy and adversarial images.

\subsection{Attribution Maps}

Explainability methods for image processing applications creates a pixel-wise array of attribution scores that indicate each feature's (a pixel of the input image in most cases) importance. This is known as an attribution map or an explainability map \cite{ancona2019gradient}.

\subsection{Sensitivity}

Sensitivity of an explanation method (i.e., attribution map generation method, such as Vanilla Gradient) is defined to be how much an explainability map changes when small perturbations are applied to the input of the model \cite{nauta2022anecdotal}. 
Attribution methods that produce explanations with large sensitivity scores are likely to produce dramatically different explanations when noise is introduced at the input of the model. This means that, in general, robust models tend to perform better than their non-robust counterparts when the attribution map is evaluated for sensitivity \cite{nielsen2022robust}. Generally, attribution methods with lower sensitivity produce more similar and consistent results in the presence of imperceptible noise than methods with higher sensitivity. More consistent explanations may be beneficial in many applications, but it does not reveal how important the pixels in the attribution map actually are to the model \cite{nielsen2022robust}. This motivates the examination of other metrics for evaluating explainability methods.

\subsection{Plausibility}

The first instinct is to visually examine an explainability map to see whether the attribution makes sense to a human observer \cite{nielsen2022robust, nauta2022anecdotal}. This is called plausibility \cite{nauta2022anecdotal}, which tells the users how visually convincing the explanation is to a human. For example, an attribution map that has many important pixels within the object of interest would be more plausible than an explanation that focuses on features that are irrelevant to a human. A plausibility metric has a subjective component, since humans define what are the important features according to their own visual perception. There are multiple methods which give a plausibility score corresponding to the number of important pixels that fall within a human drawn region containing the object  \cite{zhang2018top, zhou2016learning, raatikainen2022weighting}. It has been shown that robust and adversarially trained models tend to produce more plausible explanations \cite{nourelahi2022explainable, nielsen2022robust, zhang2019interpreting, chalasani2020concise}. However, plausibility can be misleading \cite{nielsen2022robust}. The explanations that look more reasonable to a human might misrepresent the features that the trained machine learning model uses for its internal processing and decision-making. Therefore, it is important to introduce more quantitative metrics that can help us evaluate and understand the ``goodness'' or ``faithfulness'' of an explanation. 

\subsection{Faithfulness and Fidelity}

The faithfulness of an explanation is defined to be a measure of how accurate the explanation (an attribution map in our case) is to the model itself. An explanation that is very faithful will show the user what is \emph{truly} most important to the model. In other words, faithfulness tells us the extent to which pixels deemed to be important in the attribution map are \emph{actually} important to the model.

In many recent works, fidelity and faithfulness were considered synonymous \cite{markus2021role, nguyen2020quantitative, velmurugan2021evaluating}. We argue that fidelity metrics are used to measure the faithfulness of the attribution map (or the explanation). Many different techniques have been proposed in the literature to measure the faithfulness or fidelity \cite{markus2021role, nguyen2020quantitative, velmurugan2021evaluating, yeh2019fidelity, ge2021counterfactual, petsiuk2018rise, hama2022deletion, nourelahi2022explainable, zhou2021evaluating, phan2022deepface, samek2016evaluating}. However, there is no current consensus on the best way to evaluate faithfulness of attribution maps.

The \emph{Fidelity} metric measures the correlation between the pixel attribution scores and the drop in prediction or accuracy scores when the pixel is altered or removed. This shows how well the attribution map ranks the pixels by their importance. The most common implementations of fidelity metrics involve replacing important pixels with another pixel value (most often black or the image mean) \cite{petsiuk2018rise, hama2022deletion, nourelahi2022explainable, zhou2021evaluating, samek2016evaluating, phan2022deepface}. This introduces error since the model has not been trained to understand arbitrary pixels being introduced in the input image. There have been attempts to rectify this by retraining the model for possible changes in the pixels \cite{hooker2019benchmark}. However, methods based on retraining introduce another type of error, the explanation (attribution map) is no longer being evaluated using the same model parameters (that is, the model has been modified due to retraining). Later in this work, we show that our proposed EvalAttAI eliminates the errors present in these existing methods. The reason is linked to the fact that EvalAttAI does not remove pixels, but rather perturbs pixels to a small, almost imperceptible degree.

\begin{figure*}[htpb]
     \centering
     \begin{subfigure}[h]{0.85\textwidth}
         \centering
         \includegraphics[width=\textwidth]{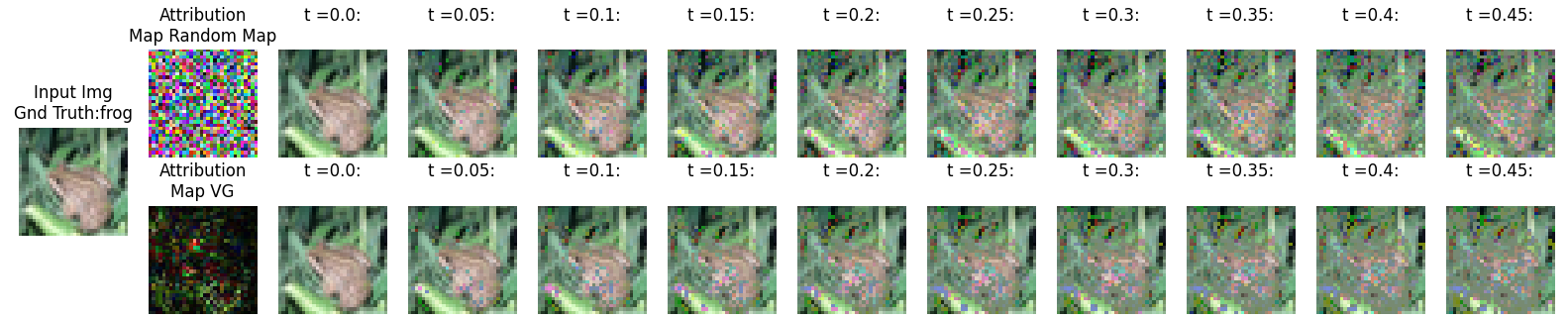}
         \caption{}
         \label{fig:del_methods}
     \end{subfigure}
     \hfill
     \begin{subfigure}[h]{0.85\textwidth}
         \centering
         \includegraphics[width=\textwidth]{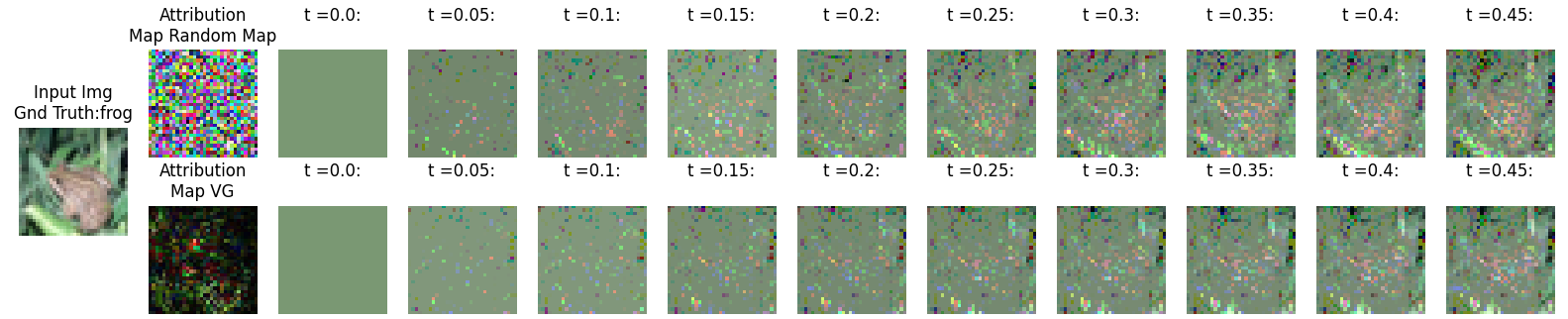}
         \caption{}
         \label{fig:ins_methods}
     \end{subfigure}
     \hfill
        \caption{Two different metrics (Insertion and Deletion) for assessing various attribution maps are presented. The variable $t$ indicates the percentage of pixels deleted/inserted where $t = 0$ refers to zero pixel and $t = 1$ refers to all pixels. Removed pixels are replaced by the channel mean. (a) Deletion and (b) Insertion examples using Vanilla Gradient (VG) and random Gaussian noise as the attribution maps.}
        \label{fig:del_ins_explained}
\end{figure*}

\subsection{Model Robustness and Attribution Faithfulness, Fidelity and Plausibility}

Using faithfulness, fidelity and plausibility, we will seek to shed more light on whether robust models are more explainable. As mentioned earlier, robust models have been shown to be more plausible. However, there is no consensus on whether the attribution maps of robust models are more faithful or whether there is a standard approach to evaluating faithfulness. 

\subsection{Attribution Evaluation Methods}

The most prominent faithfulness evaluation metrics currently include Insertion \cite{samek2016evaluating} and Deletion \cite{phan2022deepface, petsiuk2018rise, samek2016evaluating}. These methods are based on replacing pixels, and recently it was reported that their results appear contradictory \cite{nourelahi2022explainable}. That is, when Deletion performed well, Insertion did not and vice versa.

\subsubsection{Deletion}
Deletion is the most widely used method of evaluating faithfulness of attribution maps and has many variations \cite{phan2022deepface, petsiuk2018rise, samek2016evaluating}. Deletion involves (1) incrementally removing pixels, (2) replacing removed pixel spaces with an arbitrary value, and (3) evaluating the resulting change in model predictions. The order in which the pixels are chosen is based on attribution scores of these pixels. The faithfulness of the attribution map is evaluated based on the change in the model prediction. However, when replacing pixels with an arbitrary value, an error is introduced. The replaced pixels create a sharp contrast in the image, which the model was never trained to process. There have been attempts to fix this by retraining the model for each increment \cite{hooker2019benchmark}. However, retraining introduces new errors since the attributions are being evaluated on a newly trained model. EvalAttAI avoids this issue by perturbing the pixels rather than removing them and replacing with arbitrary values.


\subsubsection{Insertion}
Insertion is another method of evaluating the faithfulness of attribution maps. This method uses the same process as Deletion, but starts with all pixels removed from the input image and incrementally inserts the most important ones back. All the issues that pertain to Deletion also apply to Insertion. Even worse, these two appear to produce contradictory results showing that these methods may not be evaluating the same thing, that is, the faithfulness of the attribution methods \cite{nourelahi2022explainable}.

\subsubsection{\lowercase{c}-Eval}

The c-Eval approach \cite{vu2021ceval} evaluates attribution maps by perturbing the most important pixels using an adversarial attack and recording how this perturbation affects model predictions. However, not all pixels in an adversarial attack have the same impact on the output. Therefore, selecting only the most important pixels to perturb, is prone to error. Moreover, the adversarial attacks are designed to be applied to all pixels in the image, so only applying these to some pixels breaks the continuity. EvalAttAI alleviates this by applying the perturbation to all pixels. EvalAttAI also takes a much more direct approach than c-Eval by perturbing the pixels using attribution scores, rather than adversarial attacks that are unrelated to the attribution maps and explainability.



\subsection{ Evaluating Attributions by Adding Incrementally (E\lowercase{val}A\lowercase{tt}AI)}

Our approach to evaluating the faithfulness of explainability methods is based on adding a scaled attribution map to the input image incrementally. We show that EvalAttAI can serve as a new metric to significantly improve user trust in machine learning models. This is especially critical in medical and clinical applications. 

In this work, we focus on gradient-based explainability methods to show the applicability of EvalAttAI. However, EvalAttAI can be used for all other attribution generation methods without any changes. In general, for gradient-based methods, the attribution scores are calculated by taking the gradient of the activation score (logit or soft-max class probability values) with respect to the input features via backpropagation \cite{ancona2019gradient}. Since we will be perturbing pixels to decrease activation scores, it makes sense to look at gradient-based methods. Moreover, gradient-based attribution methods are derived directly from the loss function, which is used to train the model. Thus, these methods can be considered appropriate and effective for our proposed evaluation metric. We consider that the Vanilla Gradient method of explainability \cite{simonyan2014deep} will be the most faithful, since it works directly on the trained model without any further modification or changes \cite{nielsen2022robust}.

\begin{figure}[htpb]
    \centering
    \includegraphics[width=1.00\columnwidth]{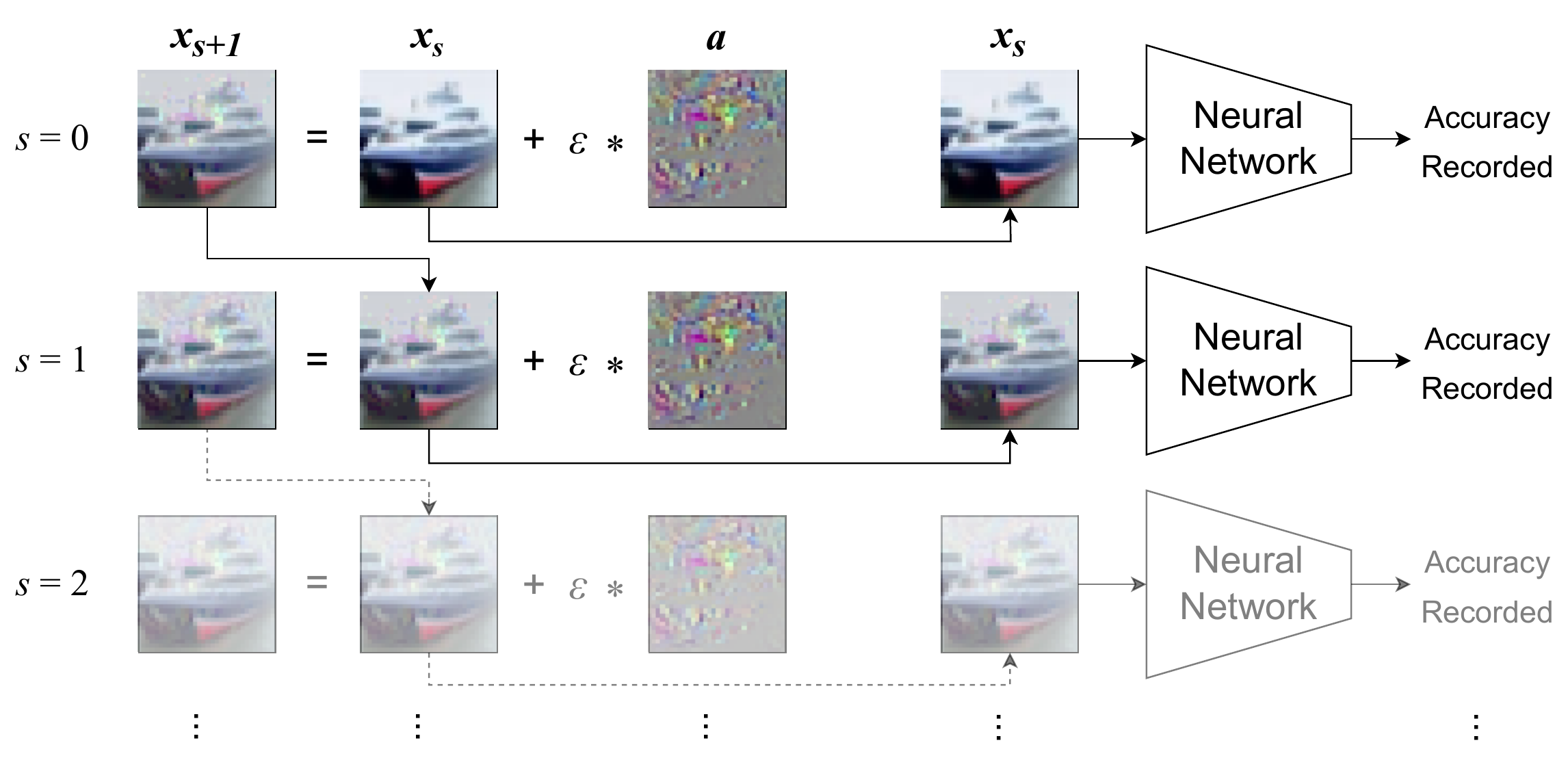}
    \caption{A schematic layout of the process of testing an attribution map using the proposed Evaluating Attributions by Adding Incrementally (EvalAttAI) method. The column labeled $\boldsymbol{x}_{s+1}$ depicts the modified input image, and $\boldsymbol{x}_s$ (s = 0) shows the clean image in the first row and the modified image from the previous iteration on all subsequent rows. The attribution map is indicated by $\boldsymbol{a}$.}
    \label{fig:faith_diagram}
\end{figure}

\begin{figure}[htpb]
    \centering
    \includegraphics[width=1.00\columnwidth]{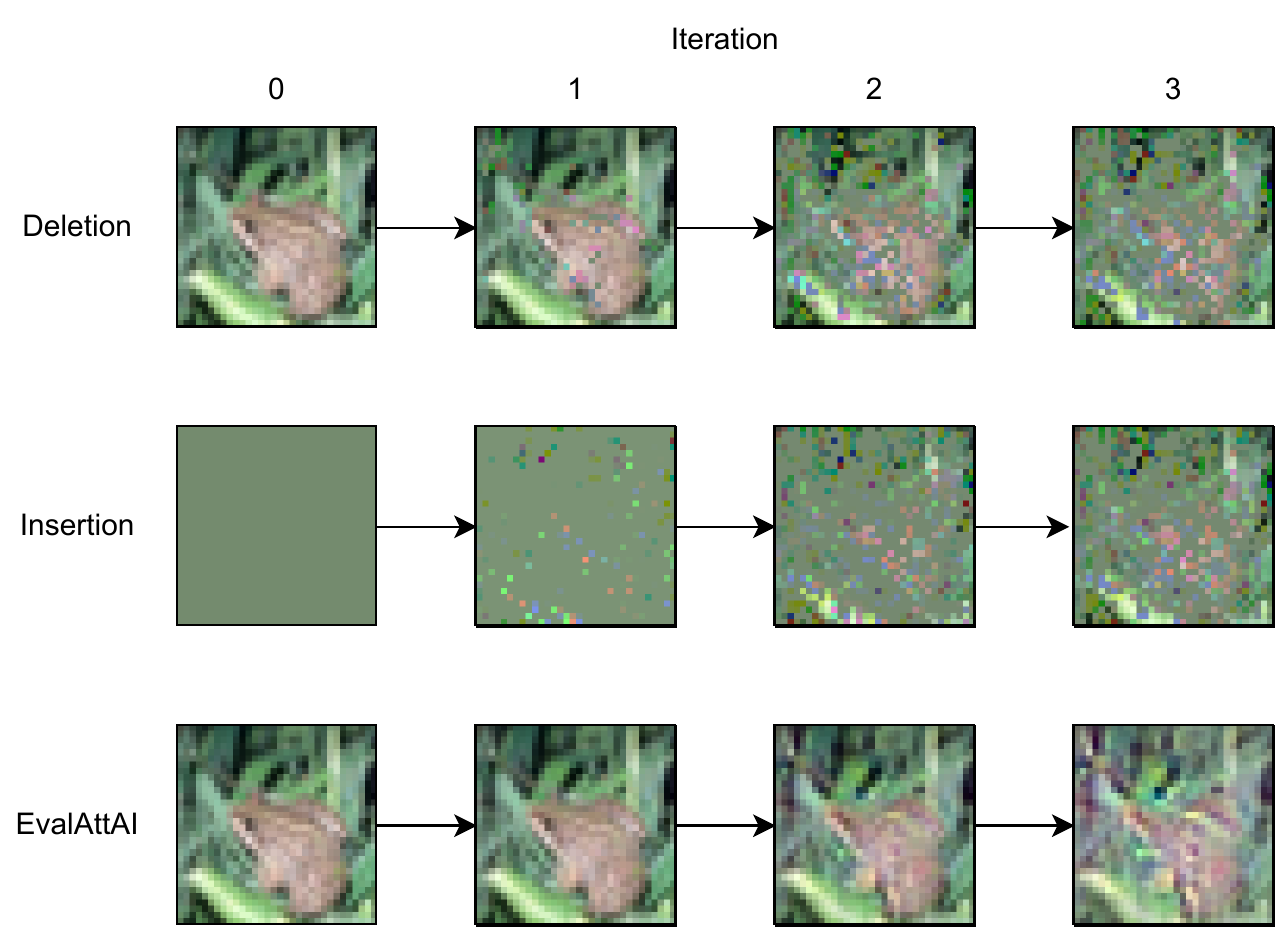}
    \caption{A visualization of each evaluation method is shown. Each image is formatted as depicted before being input into the neural network. Deletion incrementally removes and replaces pixels from the original image, and Insertion adds the original pixels back from a pixel removed image. EvalAttAI takes an entirely different approach by shifting pixel values according to the importance scores on the attribution map.}
    \label{fig:del_ins_evalattai_compared}
\end{figure}


\begin{figure*}[h]
     \centering
     \begin{subfigure}[h]{0.24\textwidth}
         \centering
         \includegraphics[width=\textwidth]{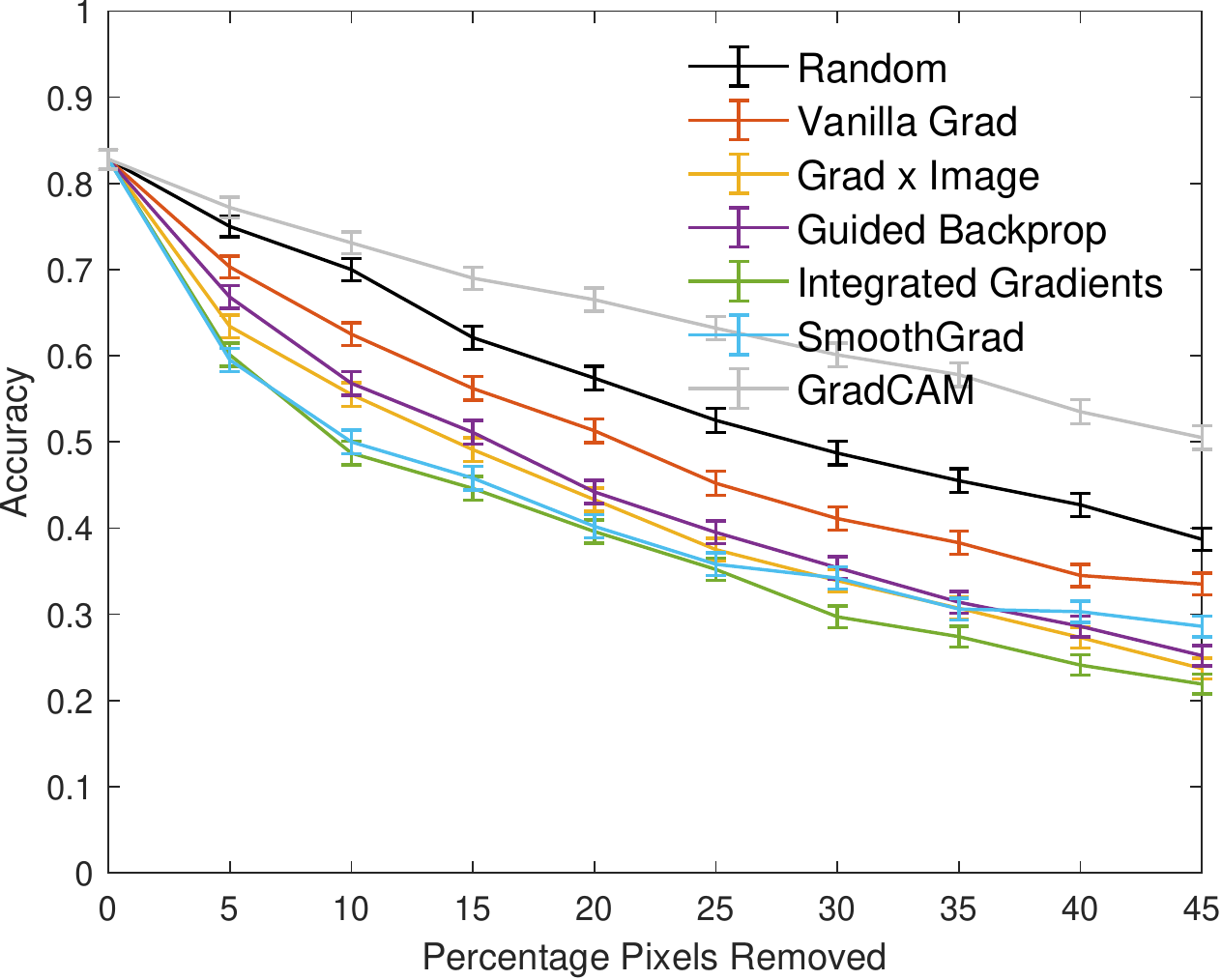}
         \caption{Deletion ResNet}
         \label{fig:del_det}
     \end{subfigure}
     \hfill
     \begin{subfigure}[h]{0.24\textwidth}
         \centering
         \includegraphics[width=\textwidth]{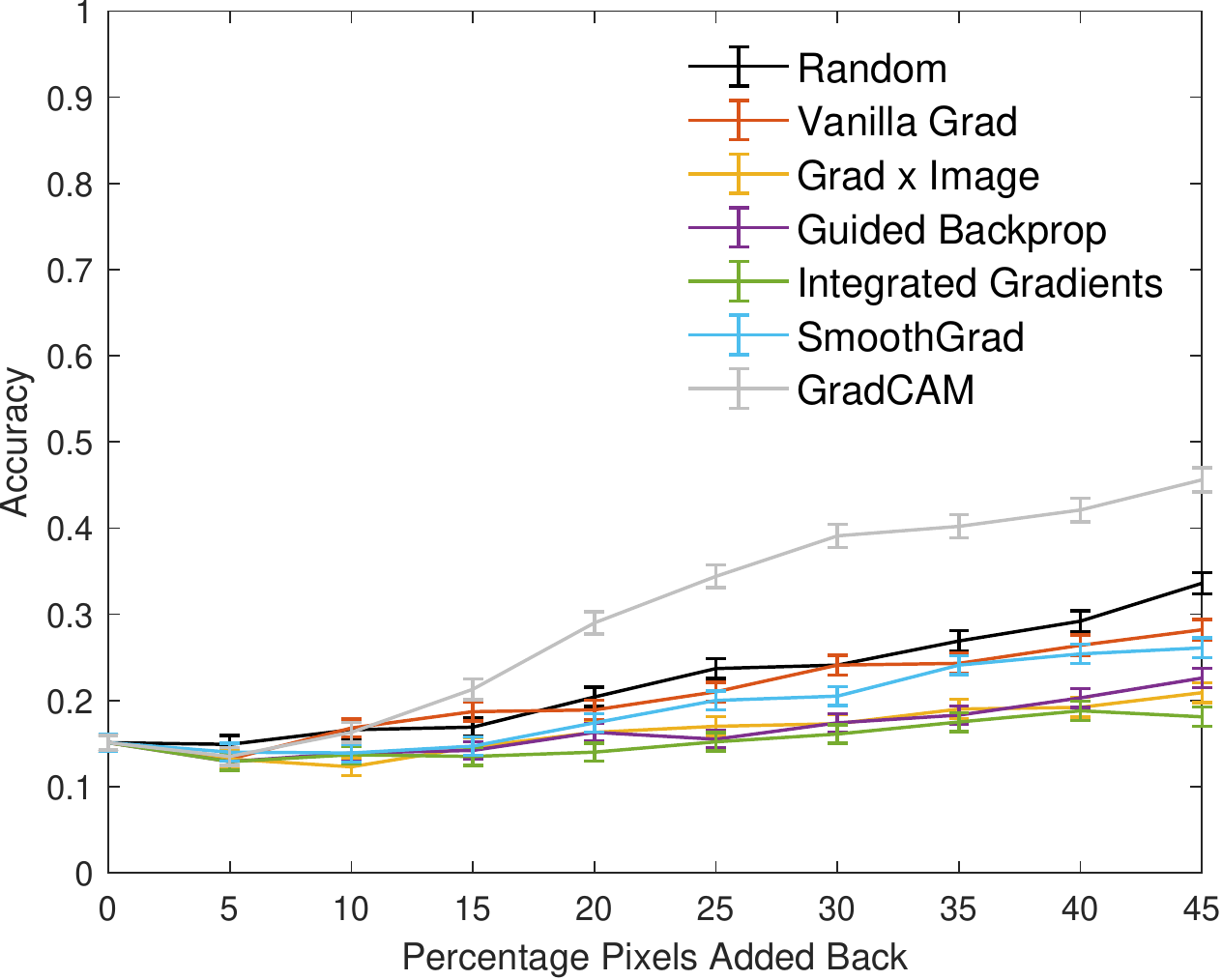}
         \caption{Insertion ResNet}
         \label{fig:ins_det}
     \end{subfigure}
     \hfill
     \begin{subfigure}[h]{0.24\textwidth}
         \centering
         \includegraphics[width=\textwidth]{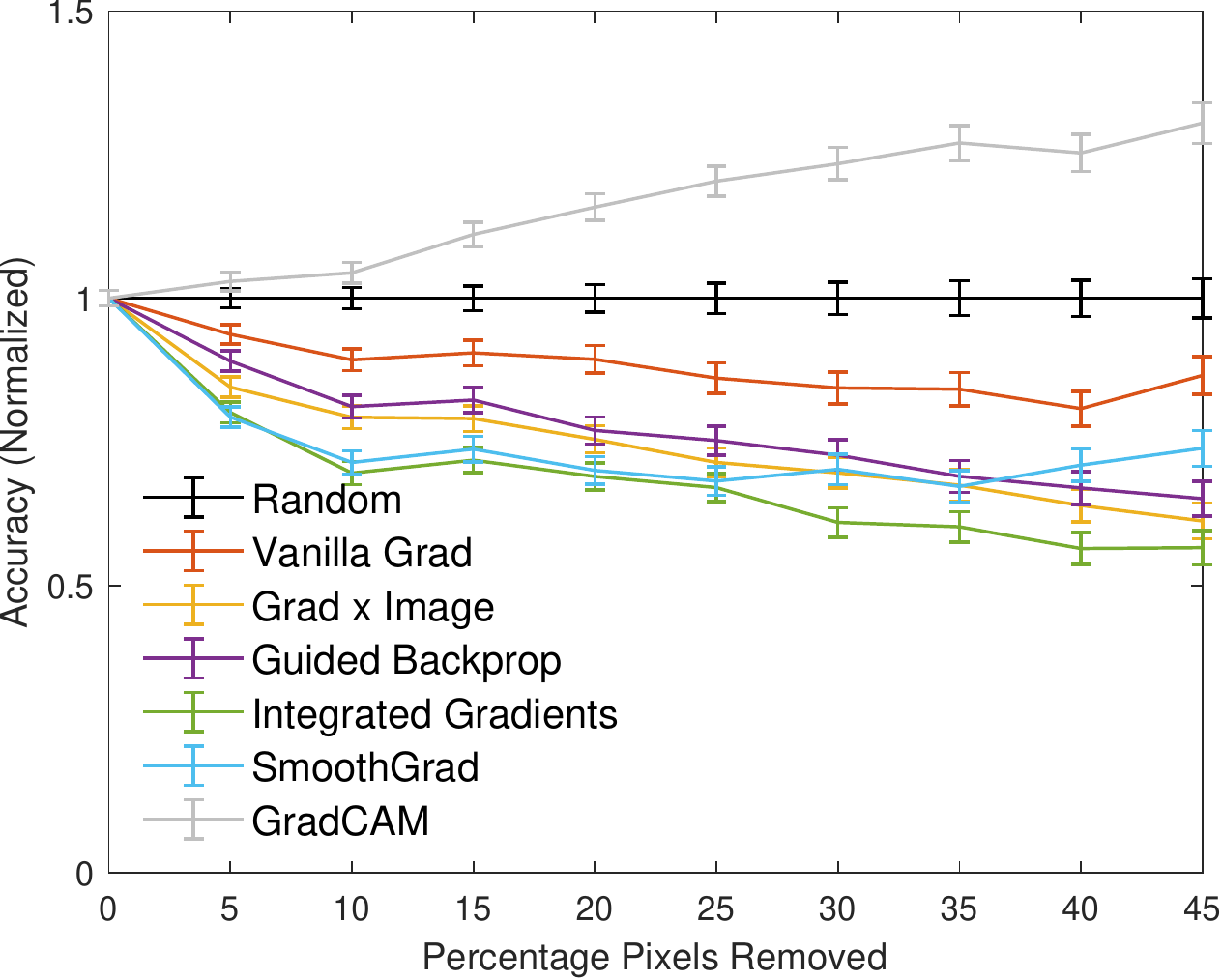}
         \caption{Deletion Norm ResNet}
         \label{fig:del_norm_det}
     \end{subfigure}
     \hfill
     \begin{subfigure}[h]{0.24\textwidth}
         \centering
         \includegraphics[width=\textwidth]{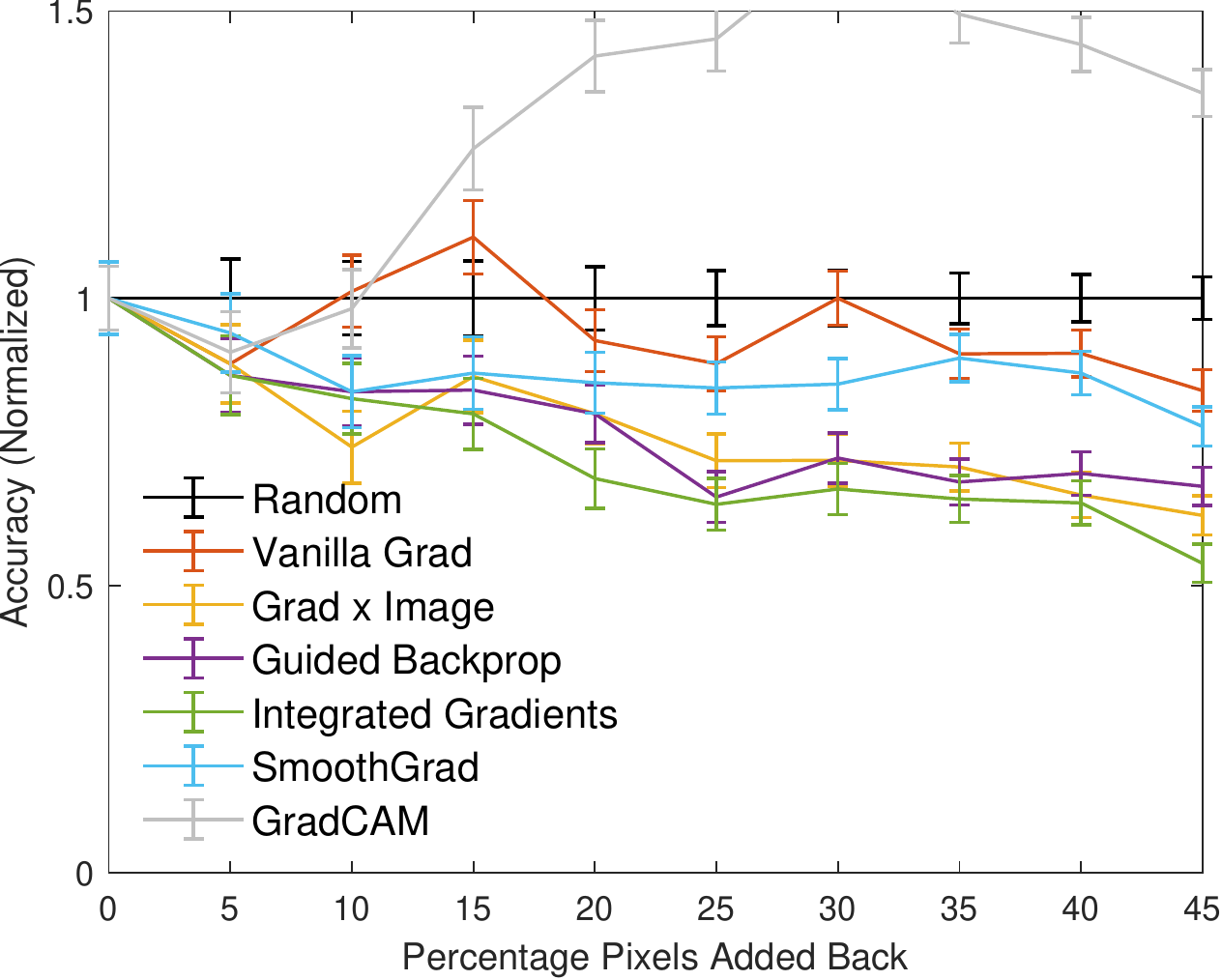}
         \caption{Insertion Norm ResNet}
         \label{fig:ins_norm_det}
     \end{subfigure}
     \hfill
     \begin{subfigure}[h]{0.24\textwidth}
         \centering
         \includegraphics[width=\textwidth]{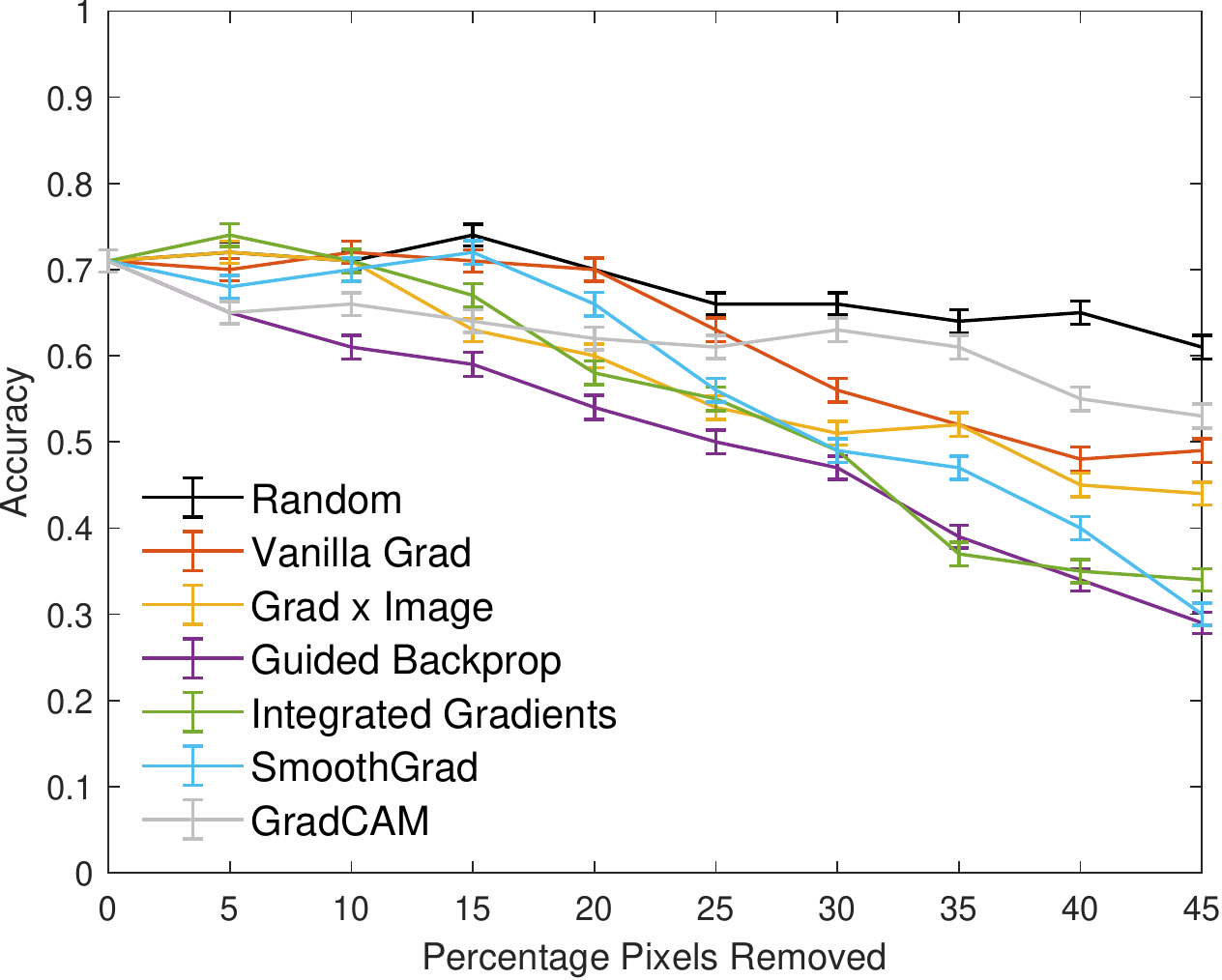}
         \caption{Deletion Robust-ResNet}
         \label{fig:del_rob}
     \end{subfigure}
     \hfill
     \begin{subfigure}[h]{0.24\textwidth}
         \centering
         \includegraphics[width=\textwidth]{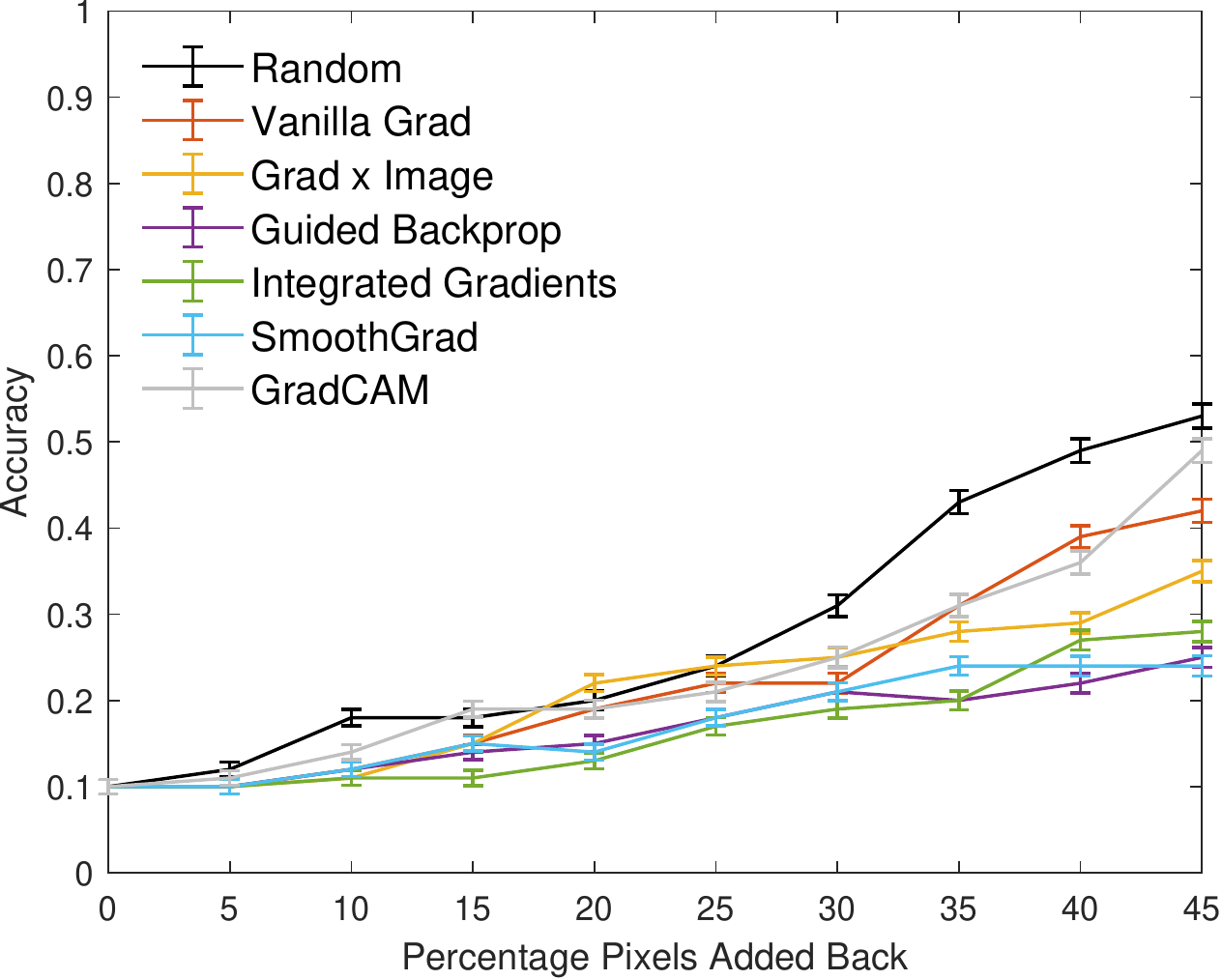}
         \caption{Insertion Robust-ResNet}
         \label{fig:ins_rob}
     \end{subfigure}
     \hfill
     \begin{subfigure}[h]{0.24\textwidth}
         \centering
         \includegraphics[width=\textwidth]{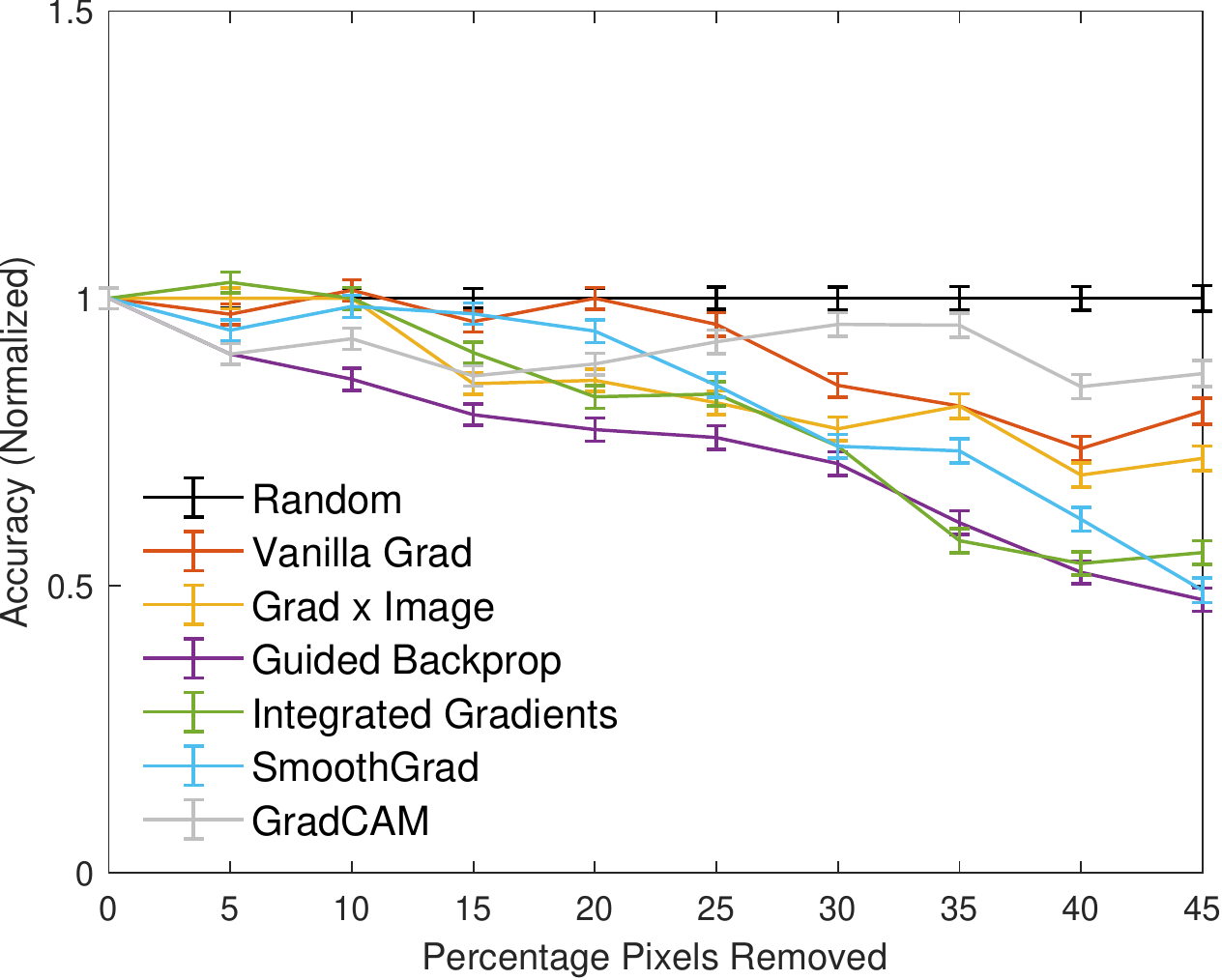}
         \caption{Deletion Norm Robust-ResNet}
         \label{fig:del_norm_rob}
     \end{subfigure}
     \hfill
     \begin{subfigure}[h]{0.24\textwidth}
         \centering
         \includegraphics[width=\textwidth]{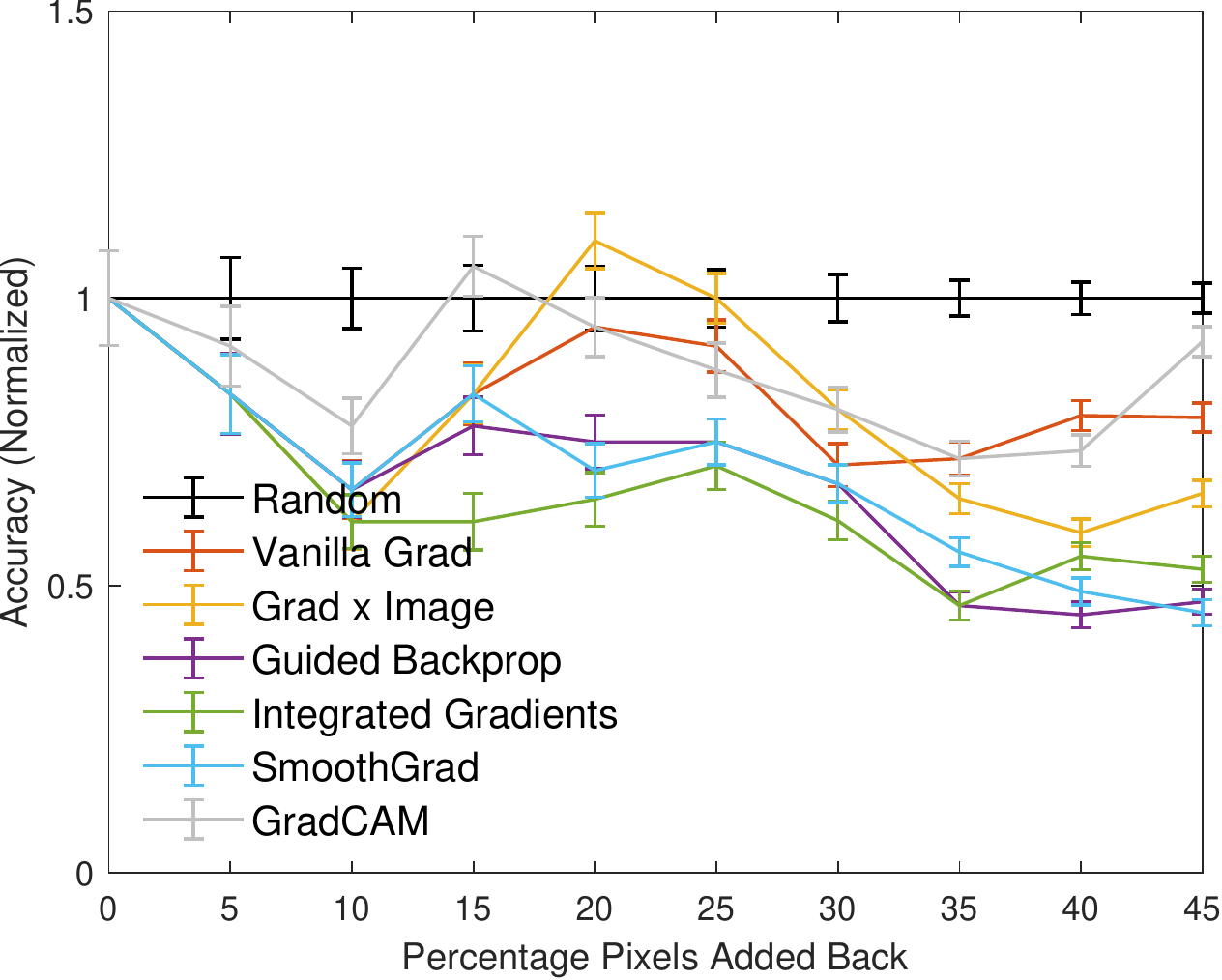}
         \caption{Insertion Norm Robust-ResNet}
         \label{fig:ins_norm_rob}
    \end{subfigure}
    \hfill
    \begin{subfigure}[h]{0.24\textwidth}
         \centering
         \includegraphics[width=\textwidth]{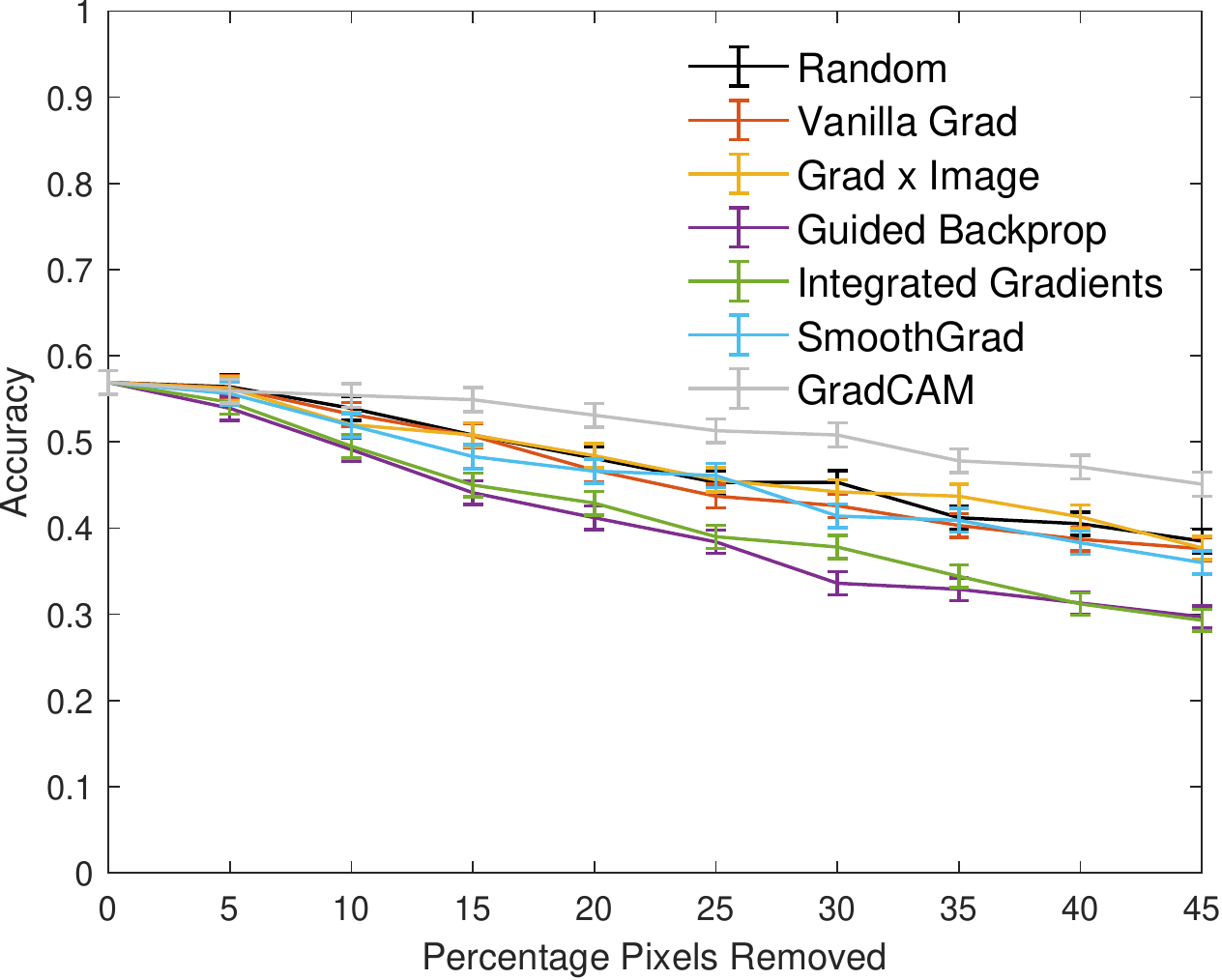}
         \caption{Deletion VDP-CNN}
         \label{fig:del_vdp}
     \end{subfigure}
     \hfill
     \begin{subfigure}[h]{0.24\textwidth}
         \centering
         \includegraphics[width=\textwidth]{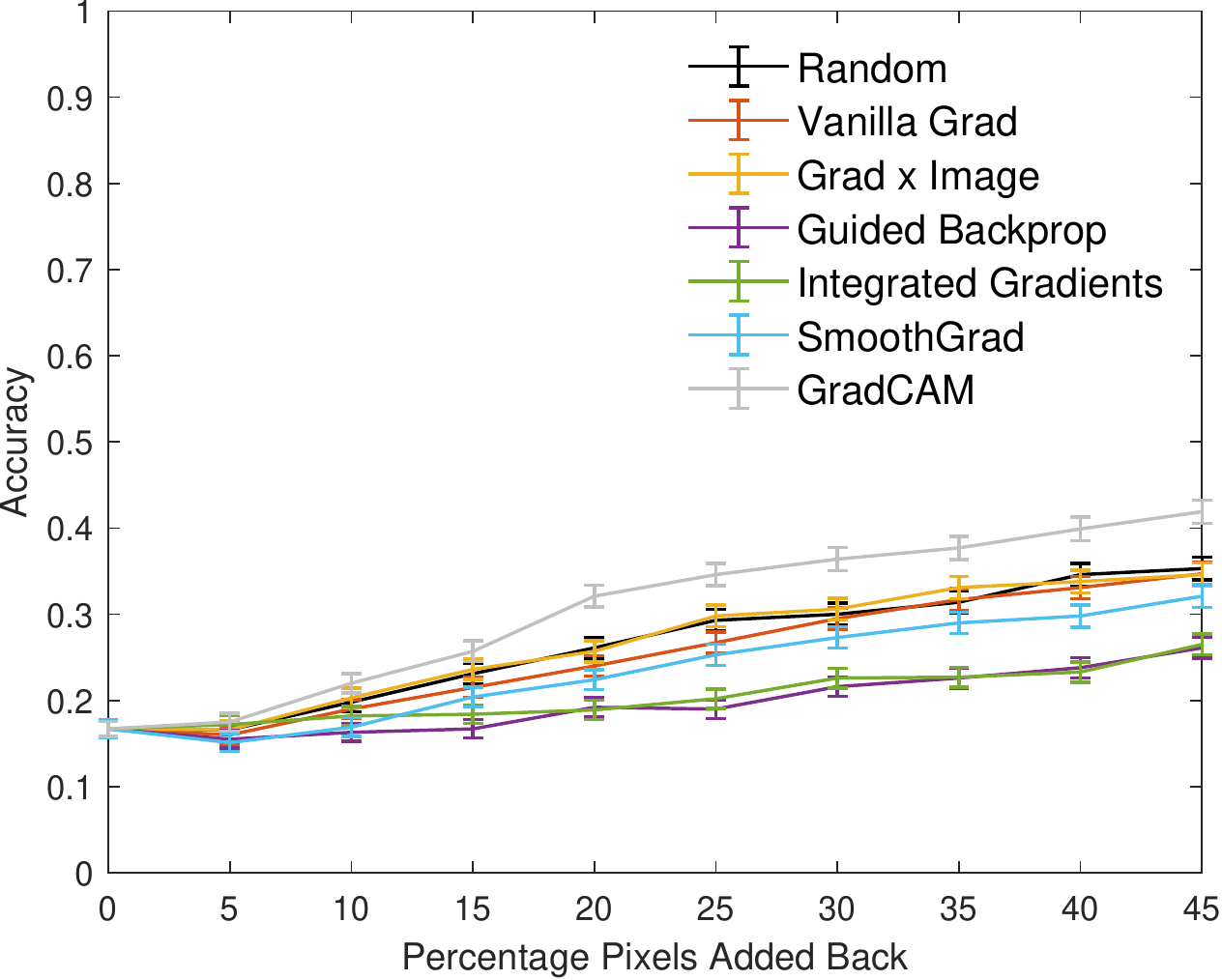}
         \caption{Insertion VDP-CNN}
         \label{fig:ins_vdp}
     \end{subfigure}
     \hfill
     \begin{subfigure}[h]{0.24\textwidth}
         \centering
         \includegraphics[width=\textwidth]{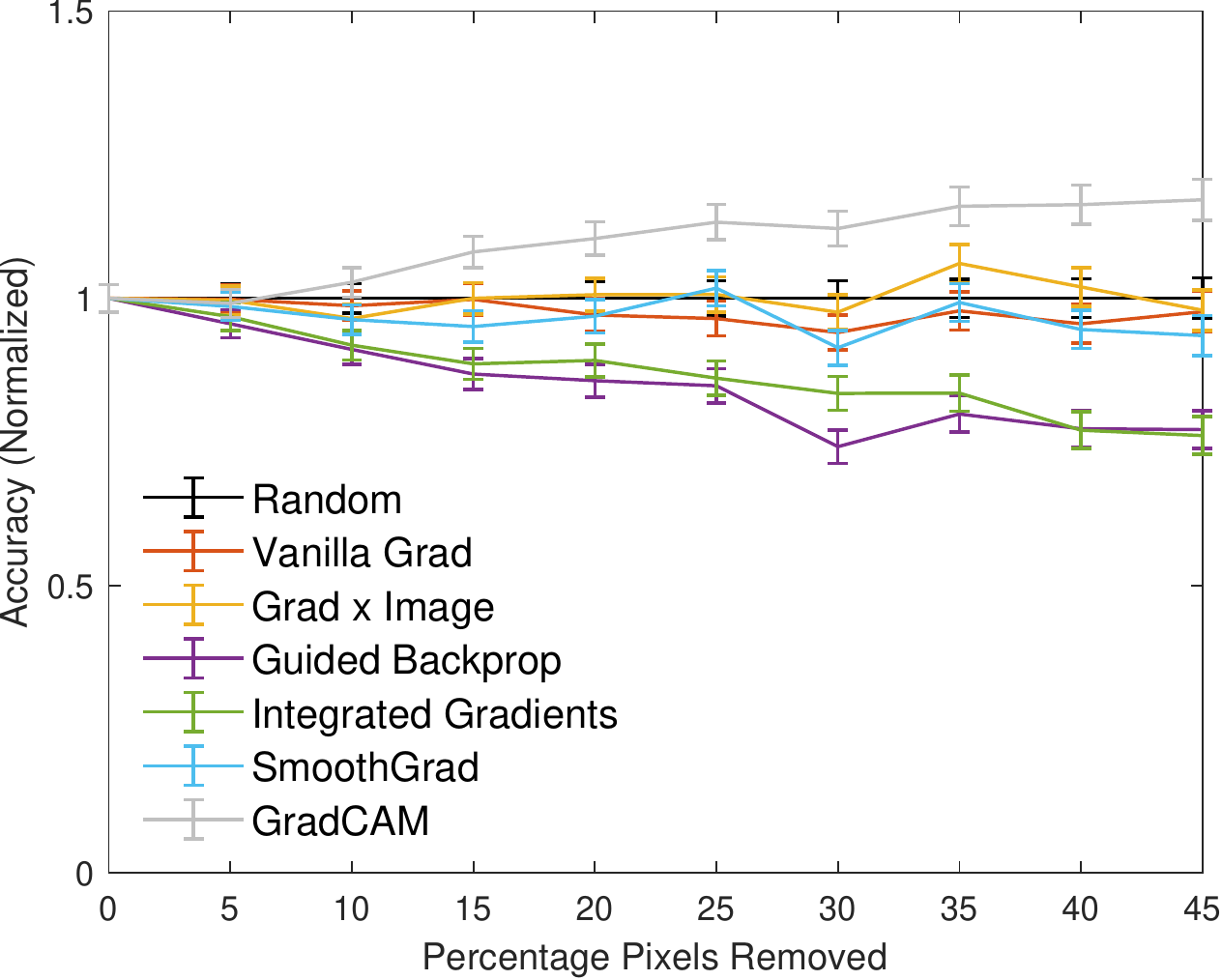}
         \caption{Deletion Norm VDP-CNN}
         \label{fig:del_norm_vdp}
     \end{subfigure}
     \hfill
     \begin{subfigure}[h]{0.24\textwidth}
         \centering
         \includegraphics[width=\textwidth]{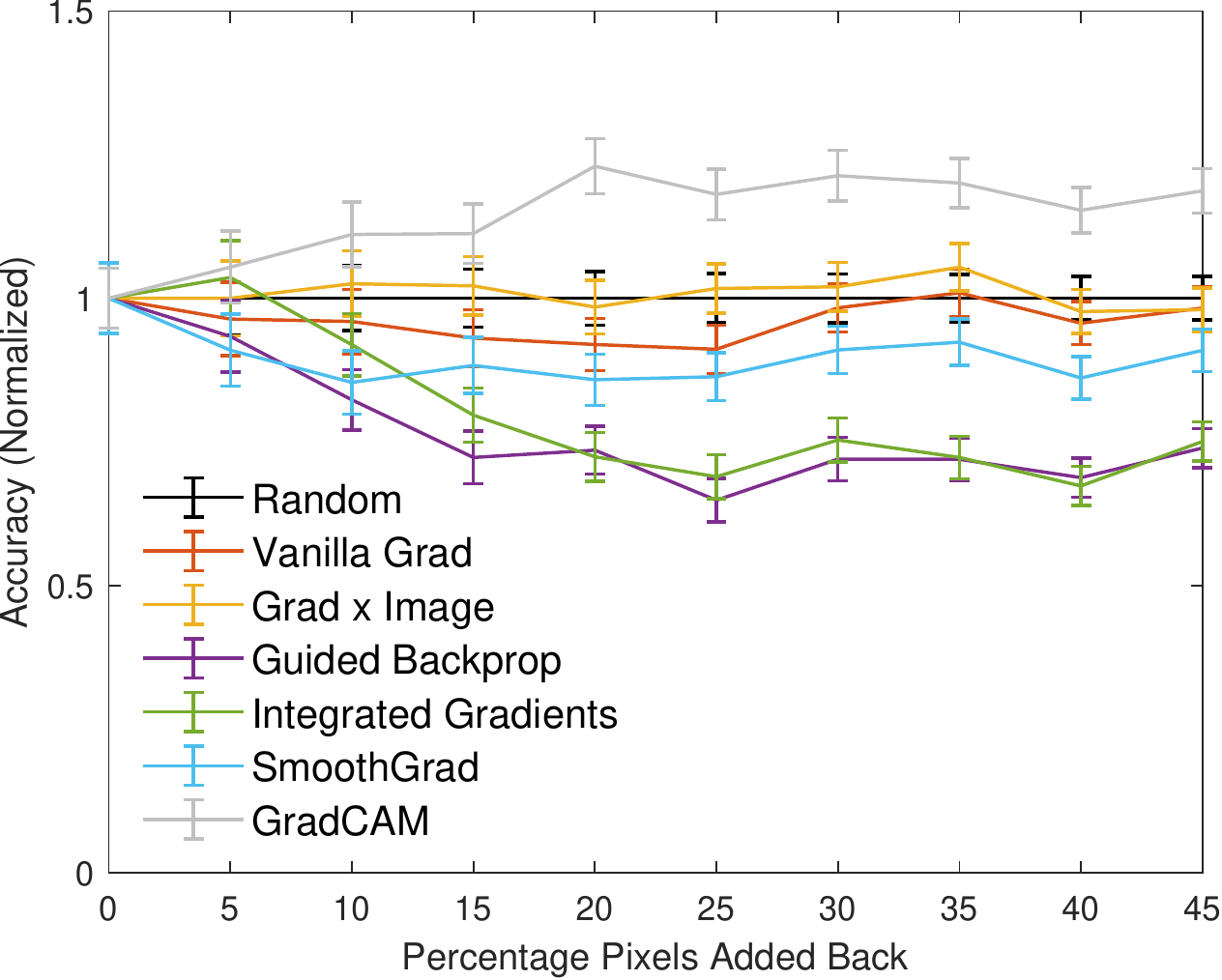}
         \caption{Insertion Norm VDP-CNN}
         \label{fig:ins_norm_vdp}
     \end{subfigure}
        \caption{Evaluating faithfulness of multiple different explainability attribution methods using Deletion and Insertion metrics. Each attribution method was evaluated using 5000 CIFAR10 images with three trained models. The top, middle, and bottom rows depict the standard trained ResNet-18 model, robustly-trained ResNet-18 (Robust-ResNet), and a Bayesian CNN trained using the VDP method respectively. The black line in all sub-figures show the behavior of the model when pixel are randomly deleted/inserted. The 95\% confidence interval is shown in each sub-figure using error bars. (Left) Deletion. (Center left) Insertion. (Center right) Deletion after normalization (abbreviated as Norm) with random baseline. (Right) Insertion after normalization with random baseline. A method is considered to be performing well on the Deletion method if the accuracy drops significantly more than random. Inversely, a method that is performing well on the Insertion metric will show the accuracy increase more than random.}
        \label{fig:del_ins_results1}
\end{figure*}

\section{Methods and Experiments}

This section describes the models, datasets and experimental protocol used in our study.

\subsection{Models and Datasets}
    
There are three models considered in our study. The first two are ResNet18 \cite{he2016deep} with different training scenarios. (1) The standardly trained ResNet18 uses a dataset with no modifications to the training data. (2) For the robustly trained ResNet18, random Gaussian noise is added to each input with a signal-to-noise ratio (SNR) of 5 dB. (3) The third model is a standardly trained Bayesian deep neural network, based on Variational Density Propagation (VDP-CNN) \cite{dera2021premium, Dera:MLSP2019, Dera:IEEERadar2019}. This model follows the same architecture as ResNet18, but propagates both the mean and the covariance of the probability distribution function defined over the model parameters through each layer of the model.

All three models were trained on the CIFAR10 dataset \cite{krizhevsky2009learning}. The ResNet18 standard model was trained until it achieved an $84\%$ validation accuracy. The robustly trained ResNet18 model achieved a validation accuracy of $70\%$. The VDP-CNN was trained on CIFAR10 to achieve a validation accuracy of $68\%$. 

The EvalAttAI method was also evaluated on three medical imaging datasets, which are part of MedMNIST. The dataset included PathMNIST, DermaMNIST, and BloodMNIST \cite{medmnistv1, medmnistv2}. All three are multi-class prediction datasets consisting of 9, 7, and 8 labels for PathMNIST, DermaMNIST, and BloodMNIST, respectively. The datasets consist of images that are classified to aid in the diagnosis of colon, skin and blood cancers. The images from the datasets used all 3 color channels. This was chosen because the ResNet architecture is designed to classify colored image data.

The standard trained ResNet18 model was trained on PathMNIST to $85\%$, DermaMNIST to $73\%$ and BloodMNIST to $92\%$ validation accuracy. The robustly trained ResNet18 was trained on PathMNIST to $86\%$, DermaMNIST to $73\%$ and BloodMNIST to $93\%$ validation accuracy. Lastly, the VDP model was trained on PathMNIST, DermaMNIST and BloodMNIST to get $76\%$, $72\%$, and $84\%$ validation accuracies, respectively.

We evaluated six attribution methods on each model including (1) Vanilla Gradient \cite{simonyan2014deep}, (2) Grad x Image \cite{ancona2019gradient}, (3) Guided Backprop \cite{springenberg2015striving}, (4) Integrated Gradients \cite{sundararajan2017axiomatic}, (5) SmoothGrad \cite{smilkov2017smoothgrad} and (6) GradCAM \cite{selvaraju2017gradcam}. For the GradCAM, we selected the first convolution layer. This means that the gradient was backpropogated to the first layer before performing global average pooling, linear combination and ReLU, resulting in the GradCAM attribution map. All attribution methods were generated using Captum \cite{kokhlikyan2020captum, kokhlikyan2019captumpytorch}, except for Vanilla Gradient and SmoothGrad which were implemented by the authors using built-in PyTorch functions.

\subsection{Deletion and Insertion Methods}

The Deletion metric evaluates how much the accuracy changes when important pixels are removed from the input image \cite{phan2022deepface, petsiuk2018rise, samek2016evaluating}. Pixels can be replaced with various values, but are often replaced with the image mean. The pixel importance is determined by the magnitude of the attribution scores on the map being evaluated. One starts with a clean image and incrementally removes pixels starting at the highest attribution score. The attribution maps are considered to be performing well if the accuracy drops significantly faster than the random baseline, which implies a smaller area under the curve (AUC). In our opinion, the AUC satisfactorily captures how well each method is performing overall by accounting for the performance over all increments using a single score. The random baseline consists of random Gaussian noise that is tested in place of an attribution map. The process of Deletion can be seen visually in Figs. \ref{fig:del_methods} and \ref{fig:del_ins_evalattai_compared}.

The Deletion experiment begins with a clean image. Then, starting from the most important according to attribution score, the pixels are removed in $5\%$ increments. The first increment $0\%$ evaluates the accuracy of the image with no pixels removed (i.e., the clean image). The next increment finds the accuracy with $5\%$ of the pixels in the image removed. This is continued until $45\%$ (almost half) of the pixels in the image are removed. The pixels are replaced with the mean of each color channel, respectively.

Insertion \cite{samek2016evaluating} uses the same premise as Deletion. The difference is that one starts with a blank image consisting of all removed pixels, and incrementally introduces the most important pixels. The more significantly the accuracy increases compared to the random baseline, the better the attribution map is performing with Insertion.
The Deletion and Insertion metrics may introduce error, since the models are not trained to be able to interpret missing and replaced pixels. In fact, the values that are chosen to replace the pixels can drastically alter the model accuracy in unintended ways. Insertion is depicted visually in Figs. \ref{fig:ins_methods} and \ref{fig:del_ins_evalattai_compared}.

For Insertion, we begin at $0\%$ with an image where all pixels are replaced by the per channel mean of the dataset. We then introduce pixels in $5\%$ increments until $45\%$ of the image is restored. The entire image can be restored, but the most meaningful change in accuracy occurs at the beginning of the replacement with the most important pixels.

 \begin{figure}[htpb]
     \centering
     \begin{subfigure}[h]{0.4\textwidth}
         \centering
         \includegraphics[width=\textwidth]{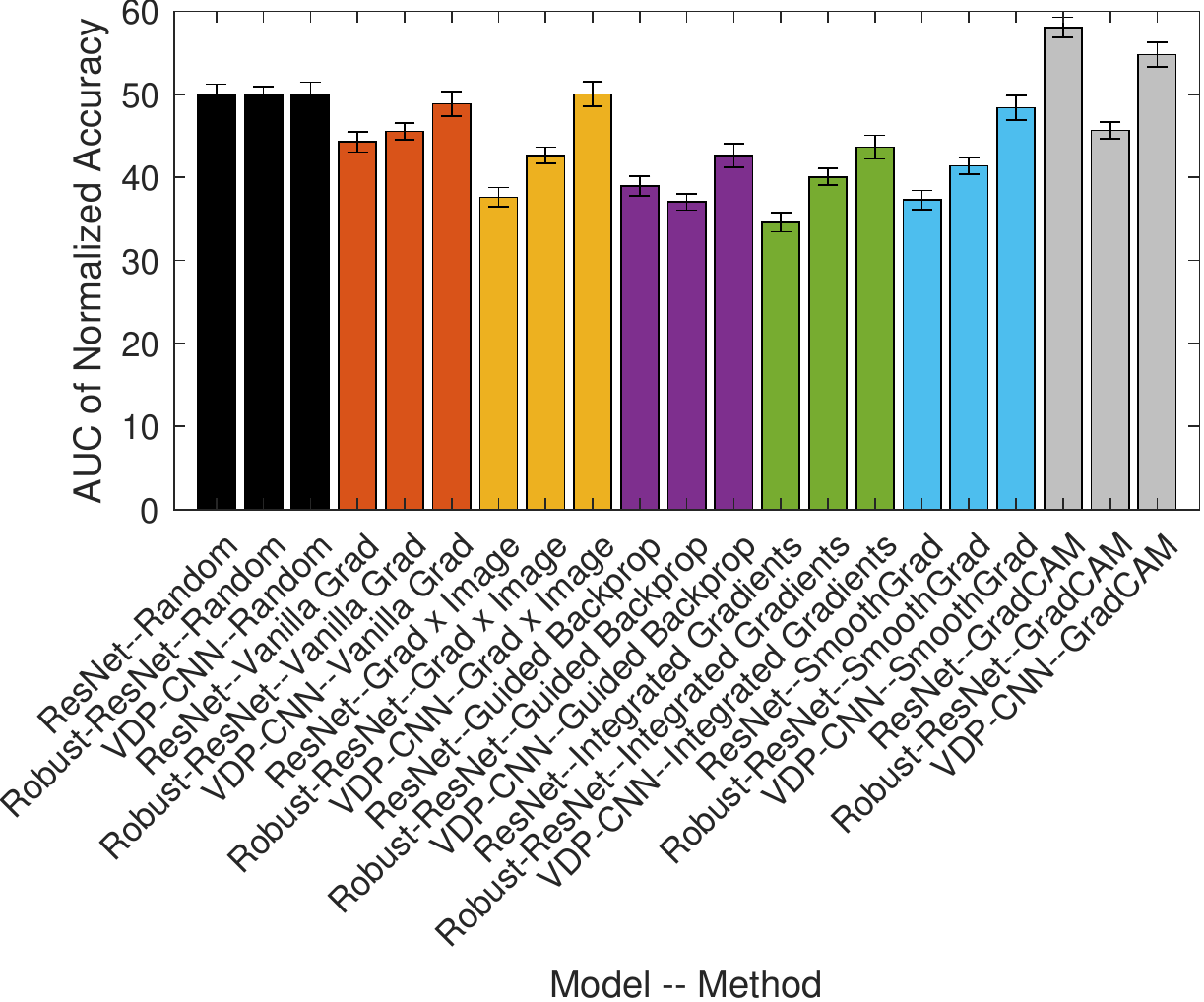}
         \caption{Deletion metric (lower is better)}
         \label{fig:auc_del}
     \end{subfigure}
     \hfill
     \begin{subfigure}[h]{0.4\textwidth}
         \centering
         \includegraphics[width=\textwidth]{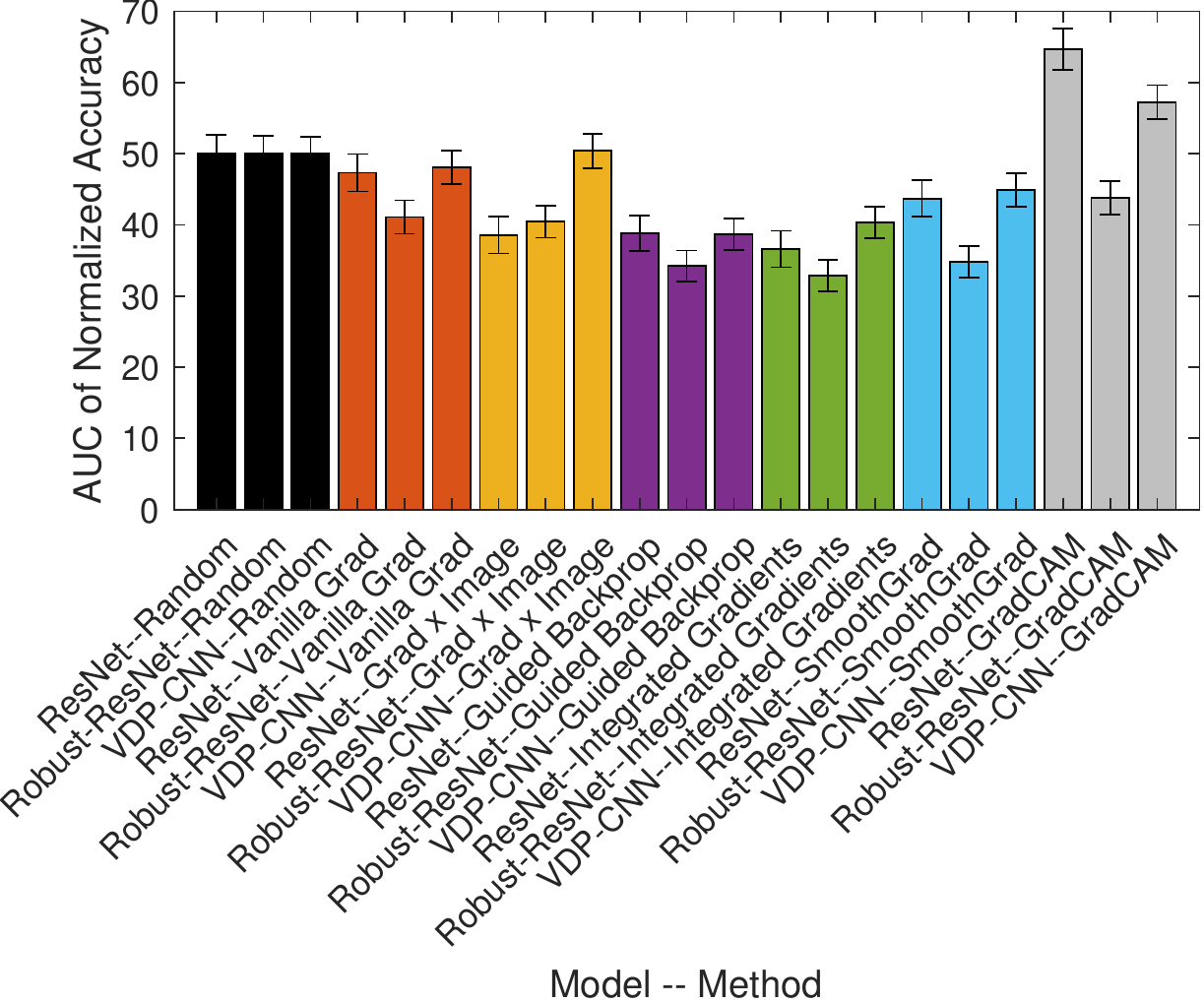}
         \caption{Insertion metric (higher is better)}
         \label{fig:auc_ins}
     \end{subfigure}
        \caption{Area under the curve (AUC) calculated using normalized accuracy curves (from Figs. \ref{fig:del_norm_det}, \ref{fig:ins_norm_det}, \ref{fig:del_norm_rob},
        \ref{fig:ins_norm_rob}, \ref{fig:del_norm_vdp} and \ref{fig:ins_norm_vdp}) for various models and attribution methods are presented. The models include a non-robust model (ResNet), a robustly-trained model (Robust-ResNet), and a Bayesian robust model (VDP-CNN). For the Deletion metric, lower values means the evaluation method and model are more explainable. For the Insertion metric, higher scores are better. The $95\%$ confidence interval is shown in each sub-figure.}
        \label{fig:del_ins_results2}
\end{figure}
 
\begin{figure*}[htpb]
     \centering
     \begin{subfigure}[h]{0.31\textwidth}
         \centering
         \includegraphics[width=\textwidth]{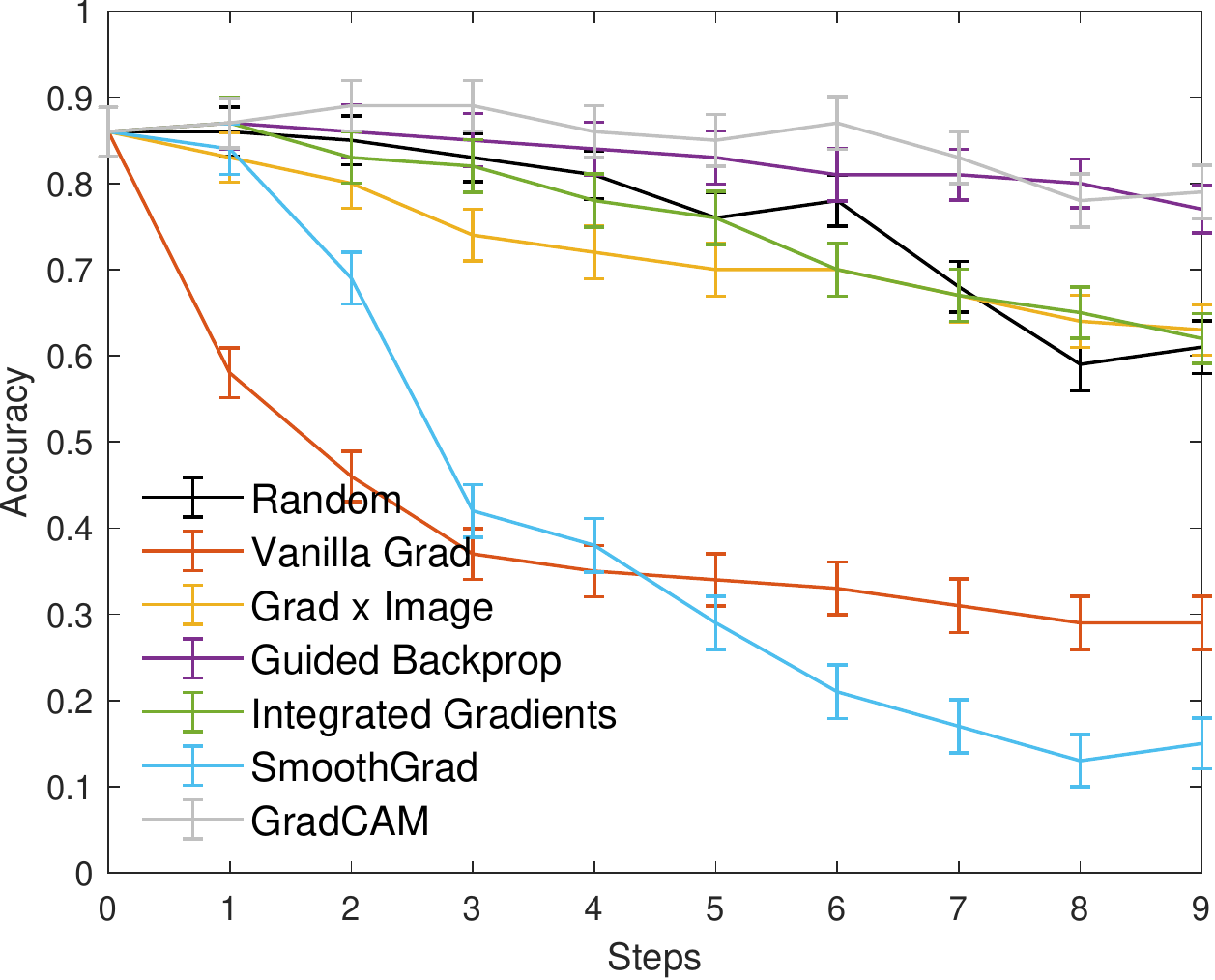}
         \caption{CIFAR10 ResNet}
         \label{fig:faith_det_cifar}
     \end{subfigure}
     \hfill
     \begin{subfigure}[h]{0.31\textwidth}
         \centering
         \includegraphics[width=\textwidth]{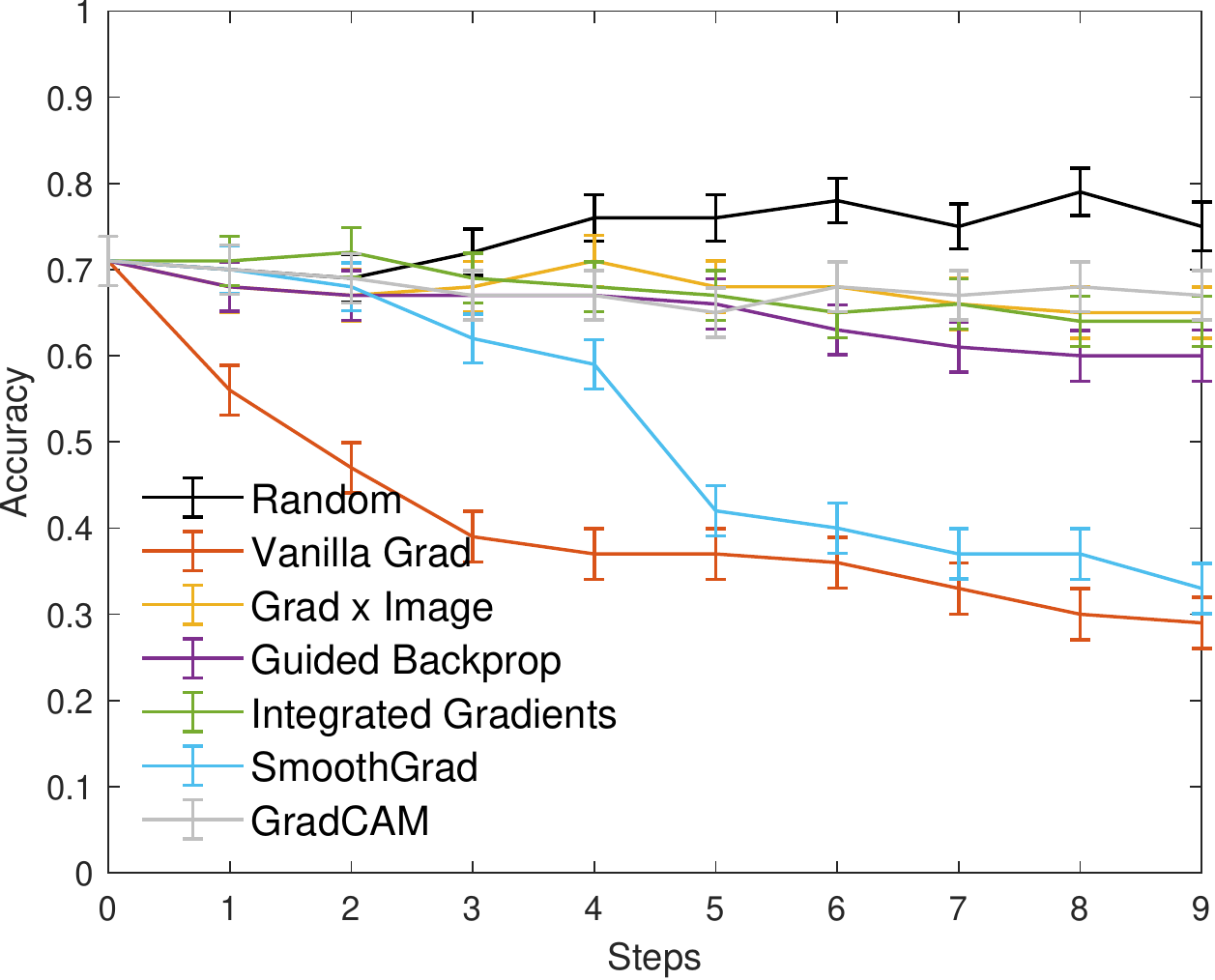}
         \caption{CIFAR10 Robust-ResNet}
         \label{fig:faith_rob_cifar}
     \end{subfigure}
     \hfill
     \begin{subfigure}[h]{0.31\textwidth}
         \centering
         \includegraphics[width=\textwidth]{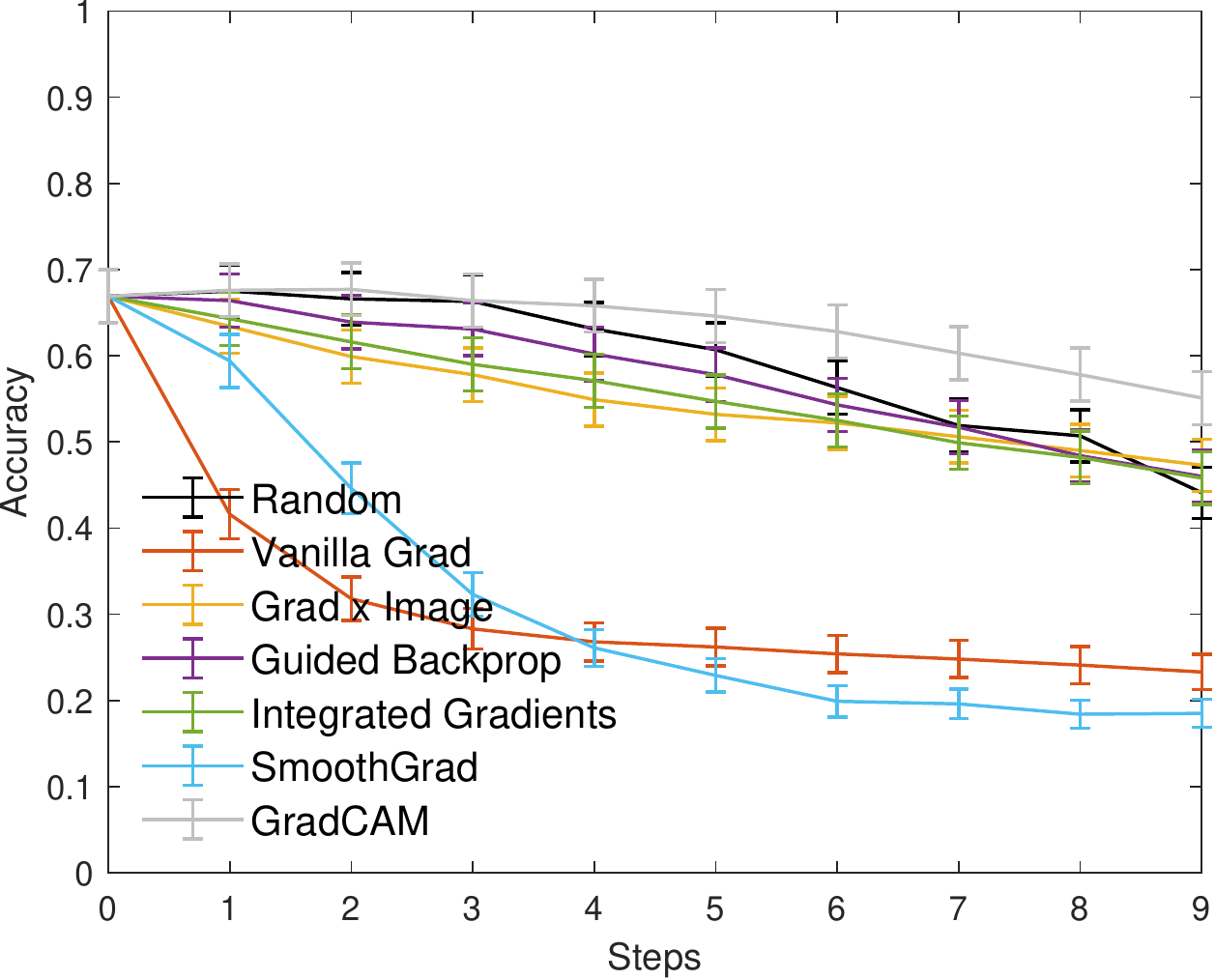}
         \caption{CIFAR10 VDP-CNN}
         \label{fig:faith_vdp_cifar}
     \end{subfigure}
     \hfill
     \begin{subfigure}[h]{0.31\textwidth}
         \centering
         \includegraphics[width=\textwidth]{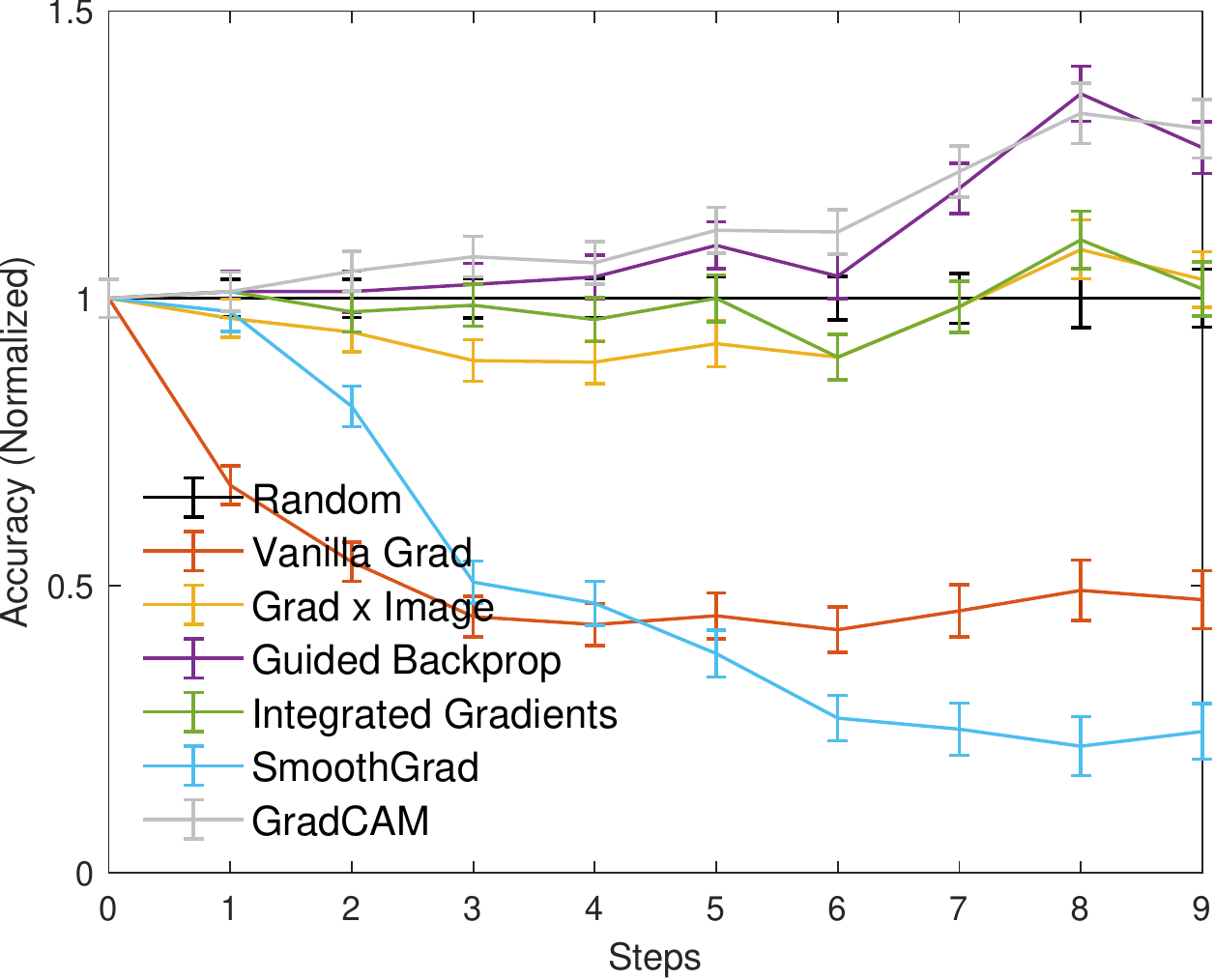}
         \caption{CIFAR10 Normalized ResNet}
         \label{fig:faith_det_norm_cifar}
     \end{subfigure}
     \hfill
     \begin{subfigure}[h]{0.31\textwidth}
         \centering
         \includegraphics[width=\textwidth]{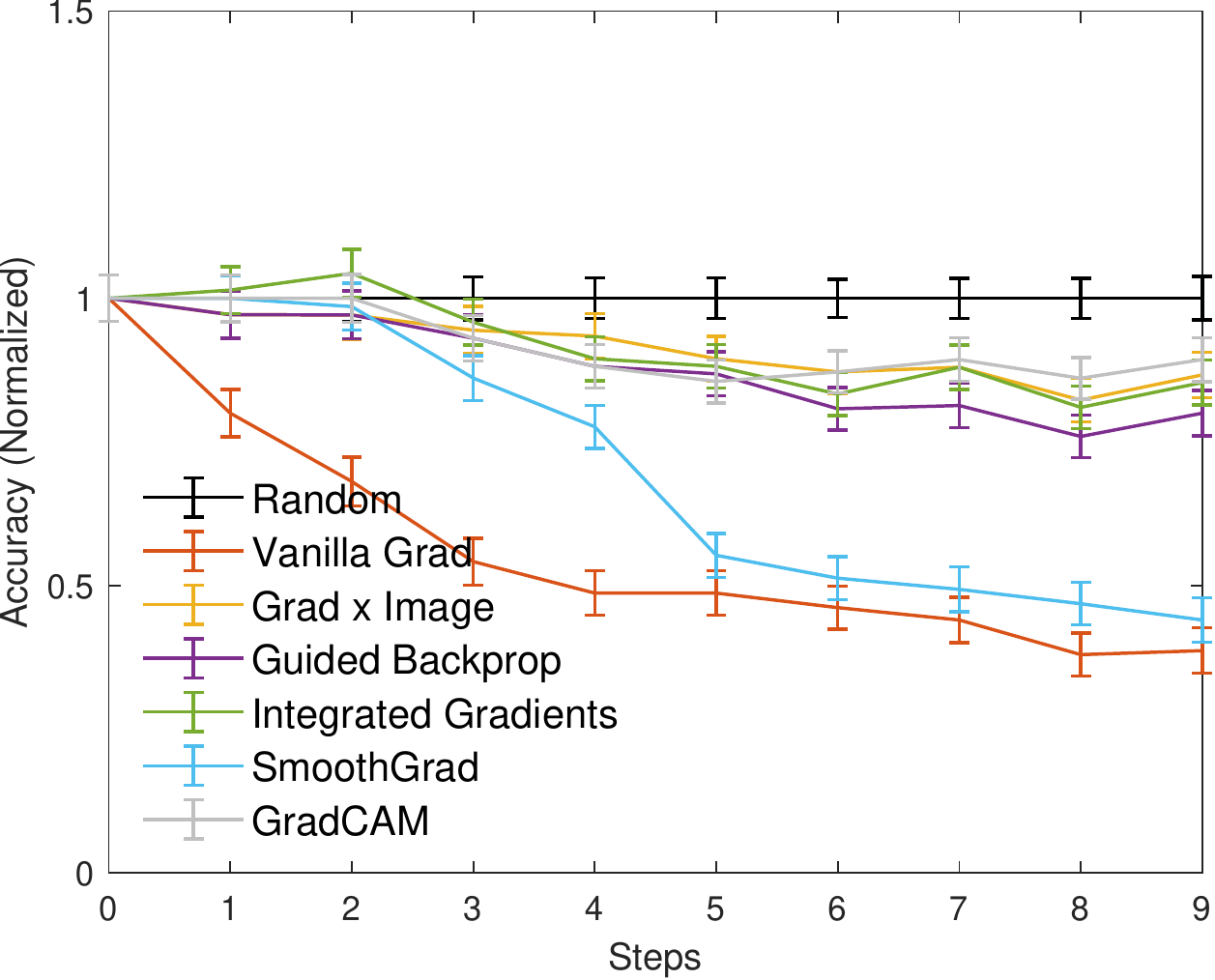}
         \caption{CIFAR10 Normalized Robust-ResNet}
         \label{fig:faith_rob_norm_cifar}
     \end{subfigure}
     \hfill
     \begin{subfigure}[h]{0.31\textwidth}
         \centering
         \includegraphics[width=\textwidth]{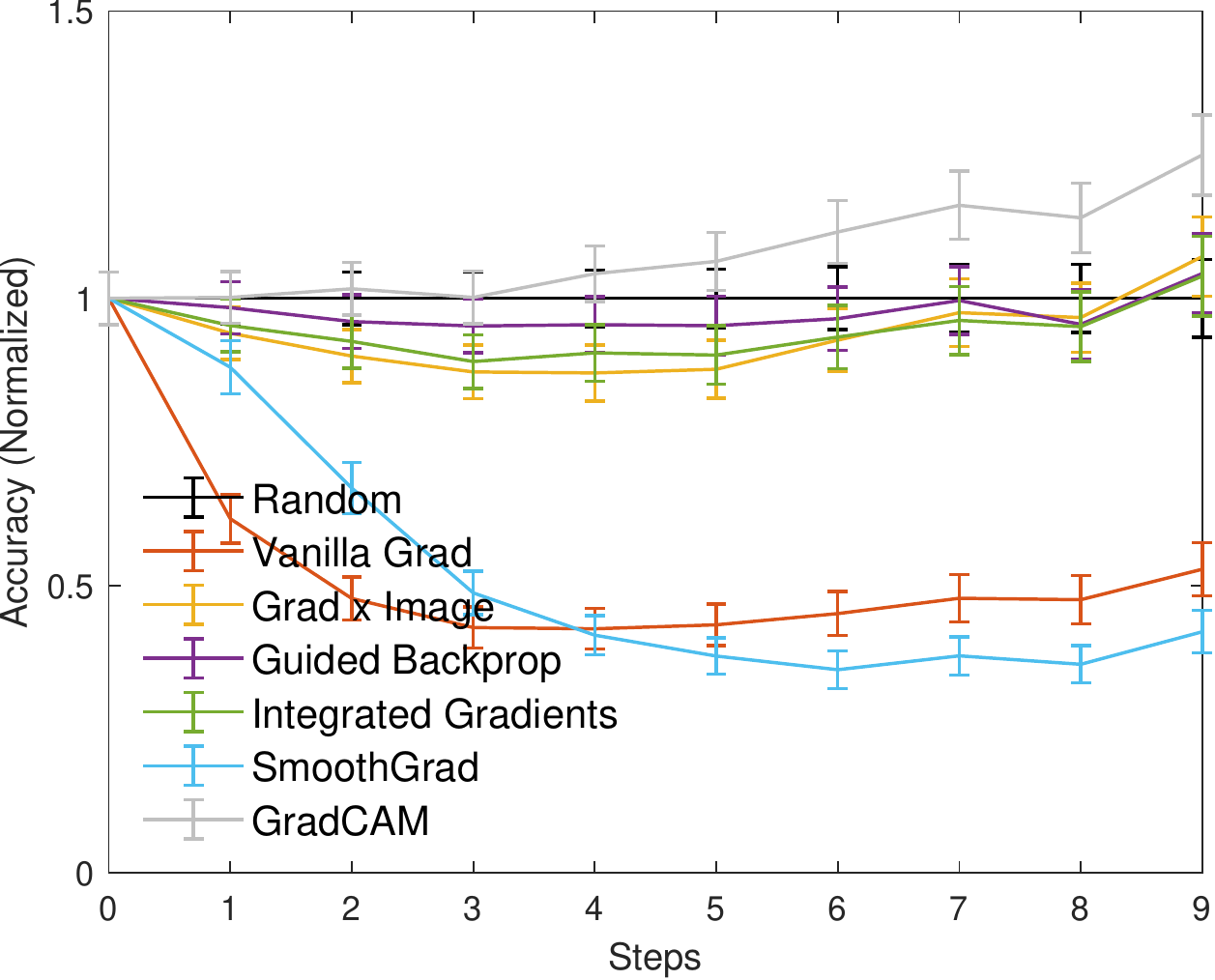}
         \caption{CIFAR10 Normalized VDP-CNN}
         \label{fig:faith_vdp_norm_cifar}
    \end{subfigure}
    \hfill
    \begin{subfigure}[h]{0.31\textwidth}
         \centering
         \includegraphics[width=\textwidth]{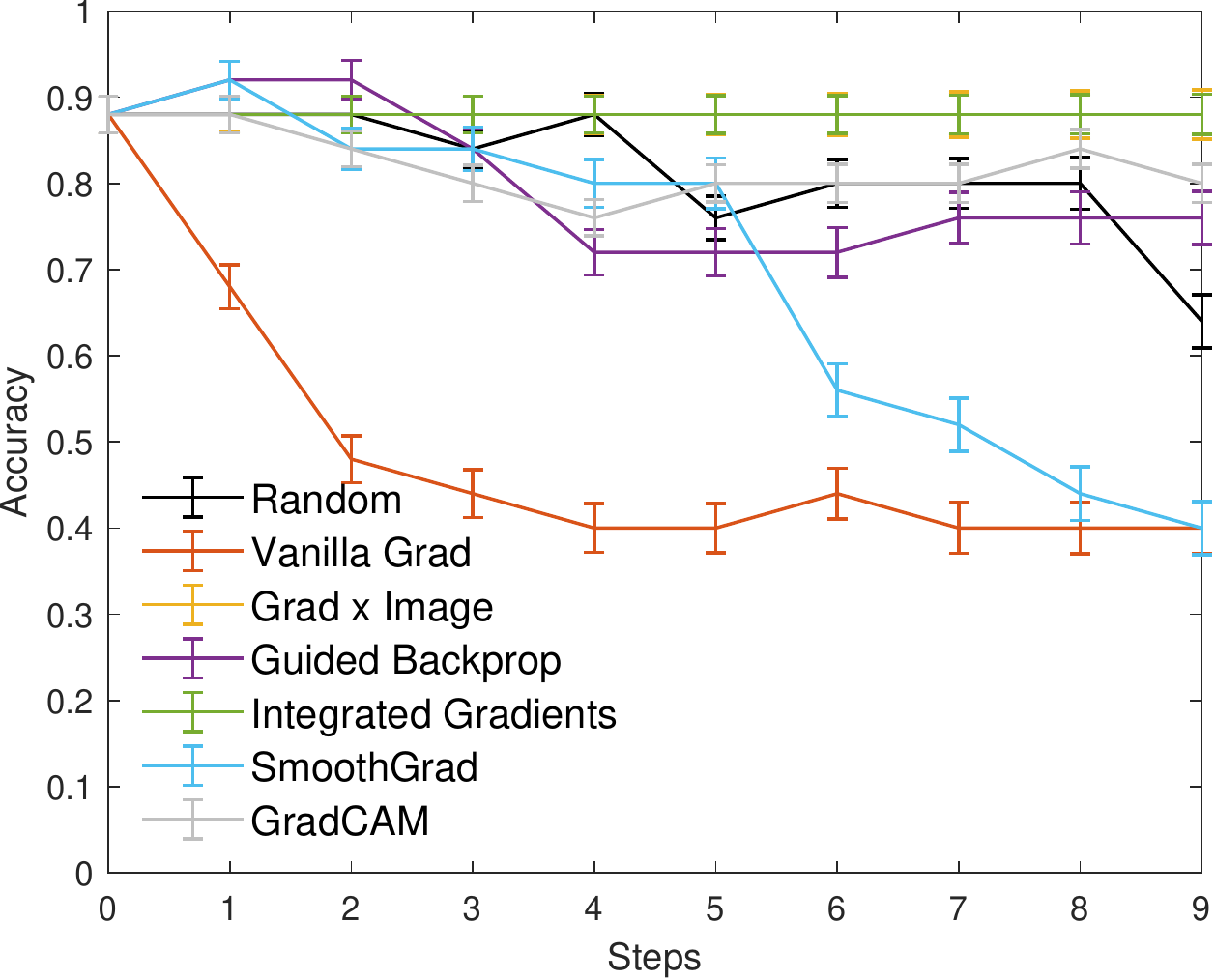}
         \caption{PathMNIST ResNet}
         \label{fig:faith_det_path}
     \end{subfigure}
     \hfill
     \begin{subfigure}[h]{0.31\textwidth}
         \centering
         \includegraphics[width=\textwidth]{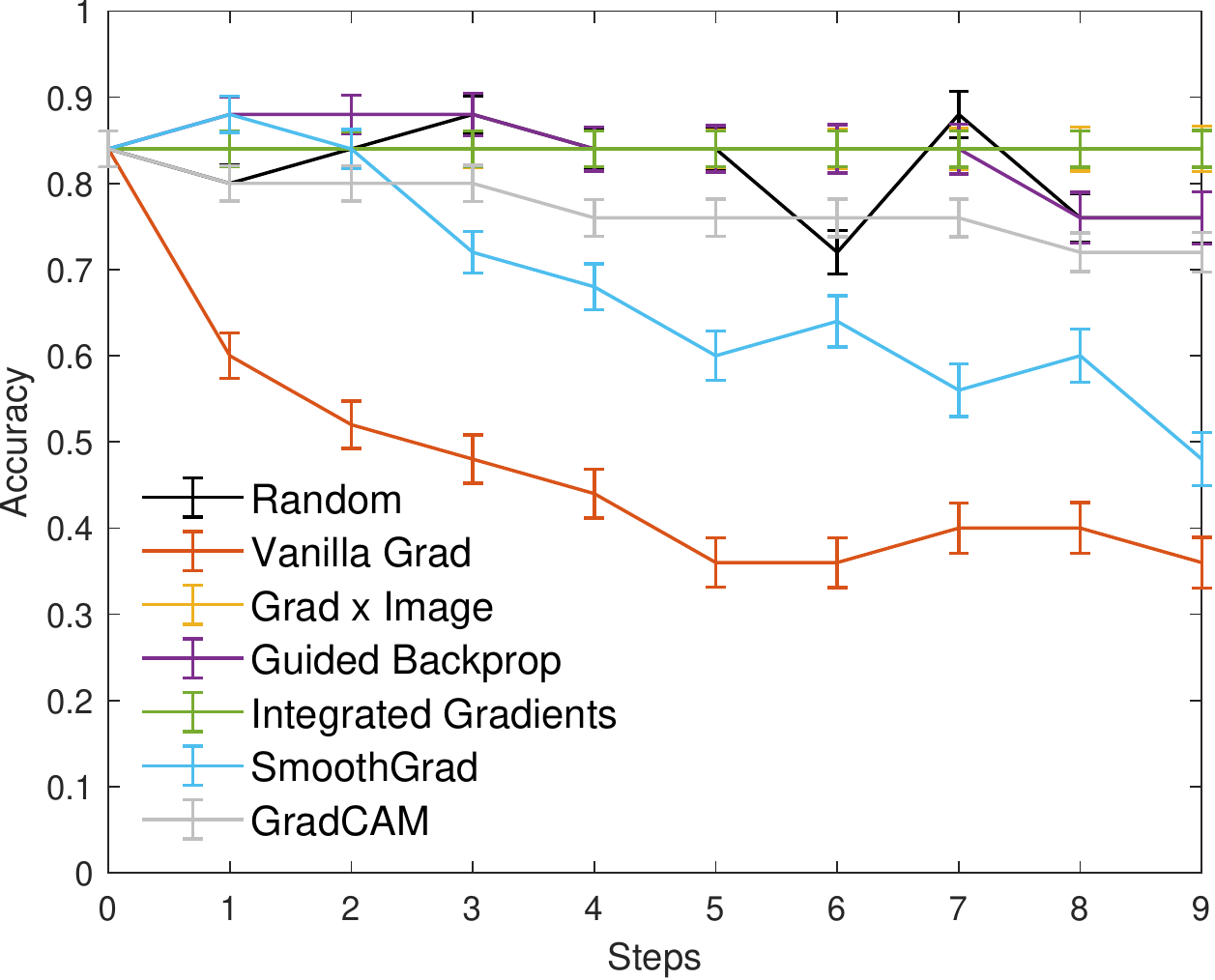}
         \caption{PathMNIST Robust-ResNet}
         \label{fig:faith_rob_path}
     \end{subfigure}
     \hfill
     \begin{subfigure}[h]{0.31\textwidth}
         \centering
         \includegraphics[width=\textwidth]{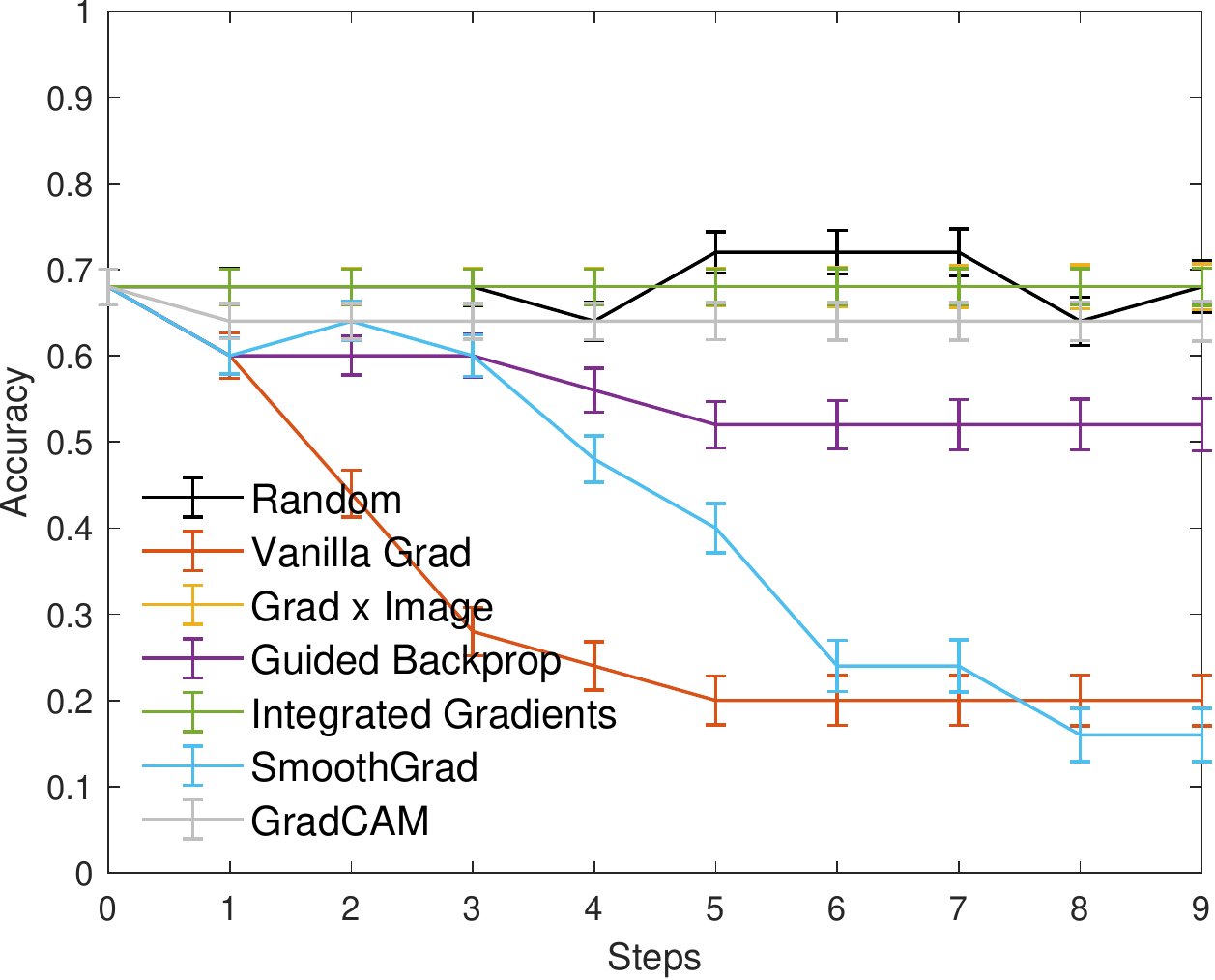}
         \caption{PathMNIST VDP-CNN}
         \label{fig:faith_vdp_path}
     \end{subfigure}
     \hfill
     \begin{subfigure}[h]{0.31\textwidth}
         \centering
         \includegraphics[width=\textwidth]{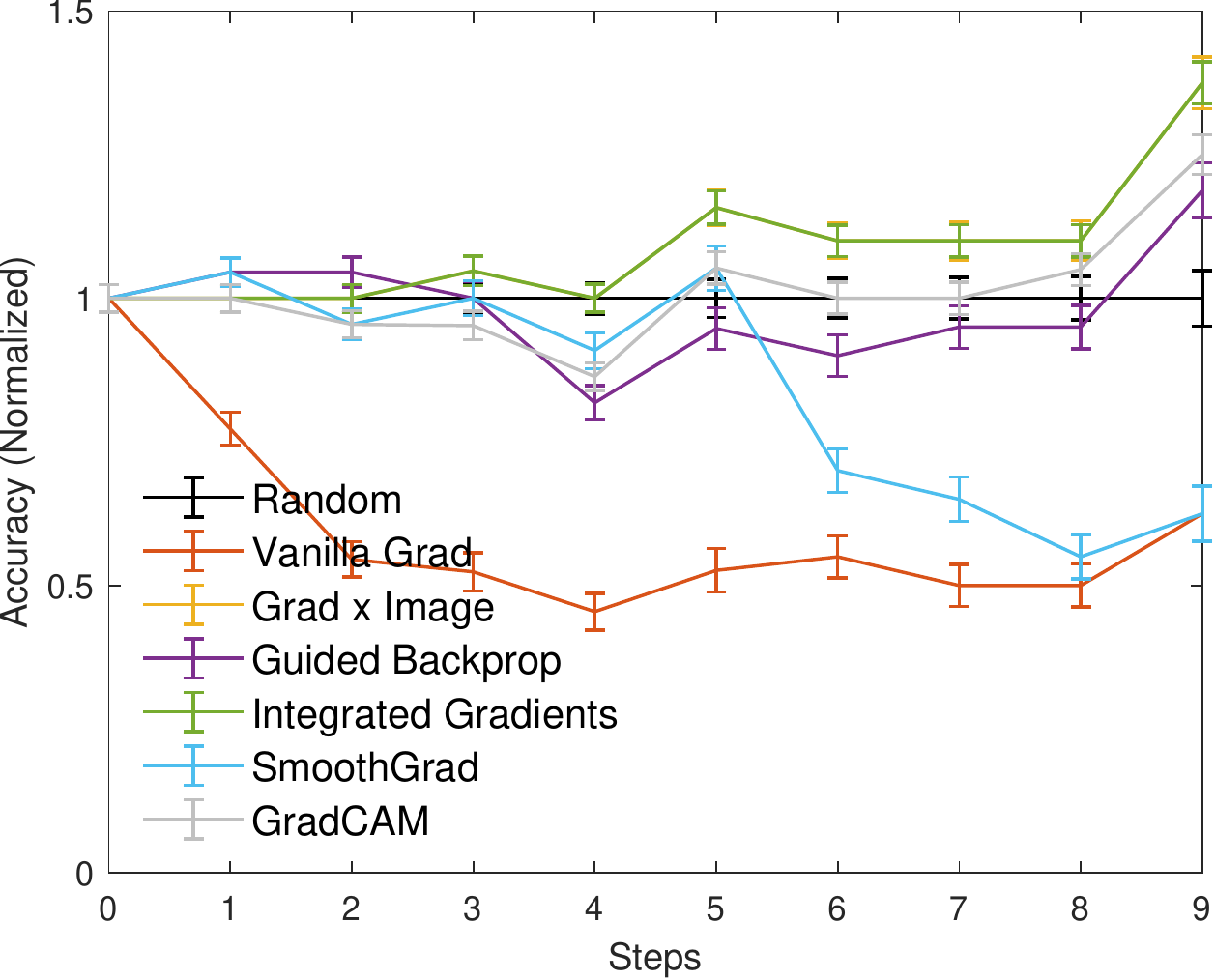}
         \caption{PathMNIST Normalized ResNet}
         \label{fig:faith_det_norm_path}
     \end{subfigure}
     \hfill
     \begin{subfigure}[h]{0.31\textwidth}
         \centering
         \includegraphics[width=\textwidth]{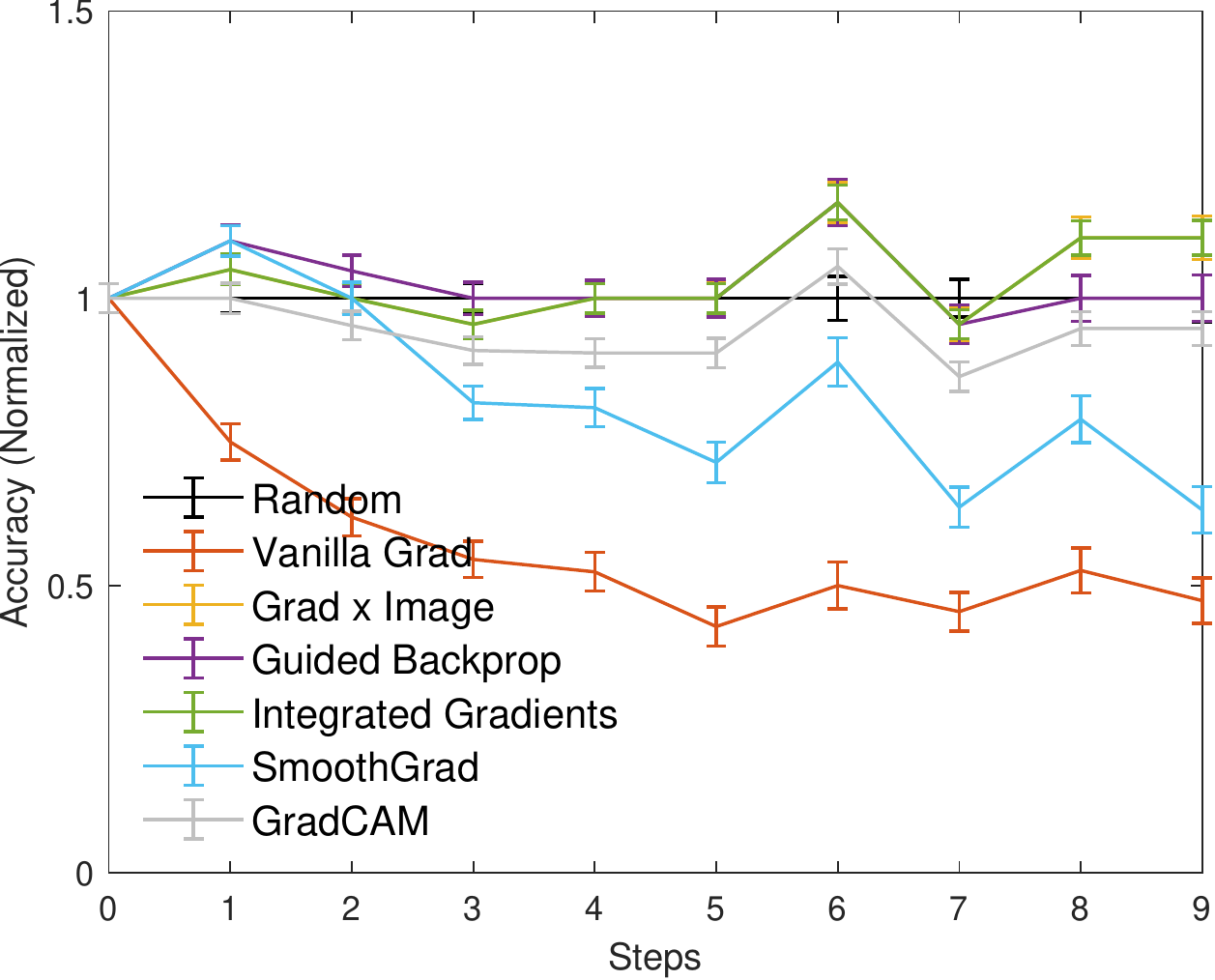}
         \caption{PathMNIST Normalized Robust-ResNet}
         \label{fig:faith_rob_norm_path}
     \end{subfigure}
     \hfill
     \begin{subfigure}[h]{0.31\textwidth}
         \centering
         \includegraphics[width=\textwidth]{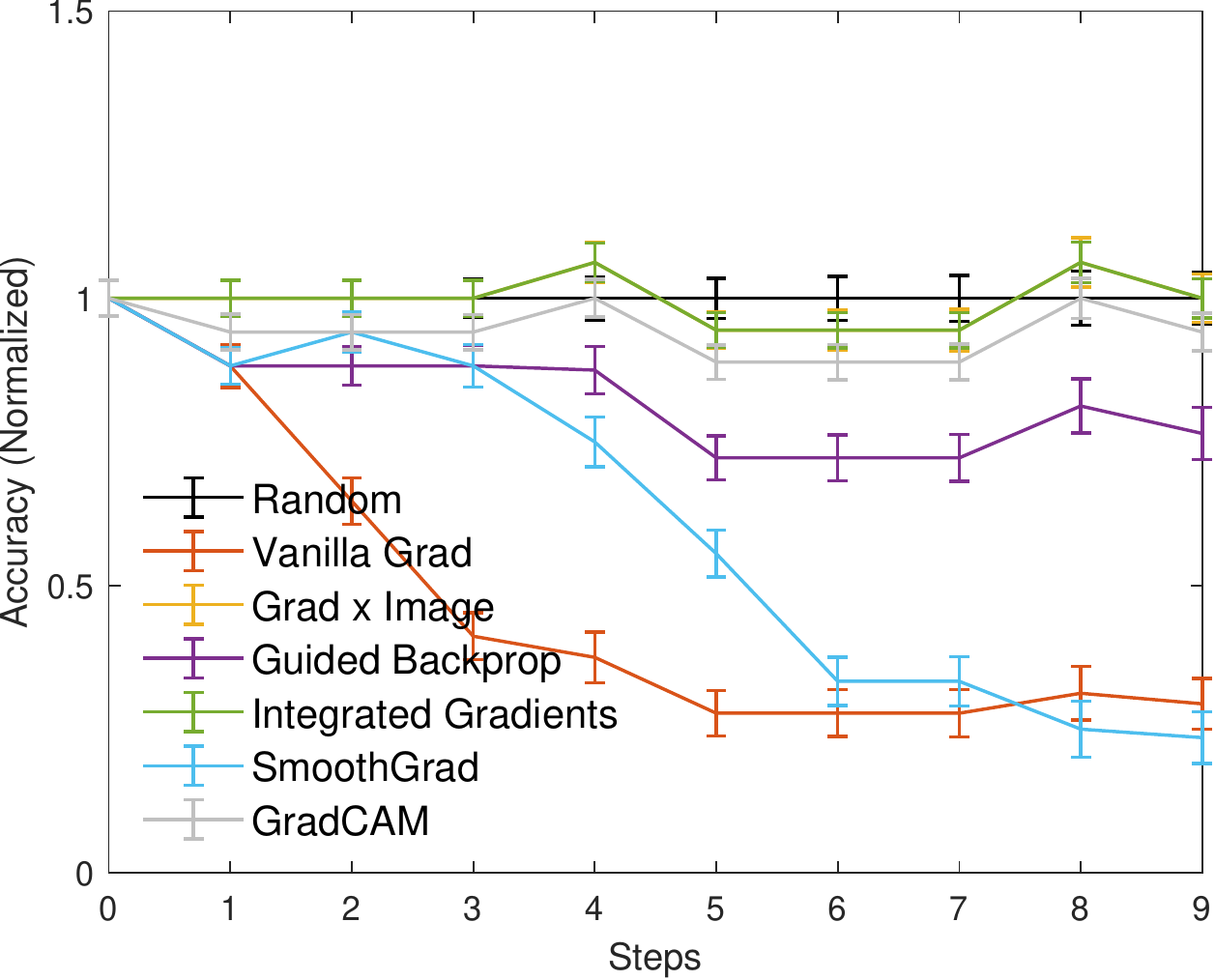}
         \caption{PathMNIST Normalized VDP-CNN}
         \label{fig:faith_vdp_norm_path}
     \end{subfigure}
        \caption{Results of the EvalAttAI metric tested using CIFAR10 and PathMNIST on three models, ResNet18, Robust-ResNet18 and VDP-CNN. The y-axis of all sub-figures shows average accuracy as attributions are added to the input image, which is captured by the x-axis. Lower accuracy values represent that the attribution method is faithful, and the metric is meant to capture this aspect. The $95\%$ confidence interval is shown in each plot using error bars. (a)-(c) and (g)-(i) Non-normalized accuracy. (d)-(f) and (j)-(l) Normalized accuracy using random as the baseline.}
        \label{fig:faith_plots1}
\end{figure*}

\begin{figure*}[htpb]
     \centering
     \begin{subfigure}[h]{0.31\textwidth}
         \centering
         \includegraphics[width=\textwidth]{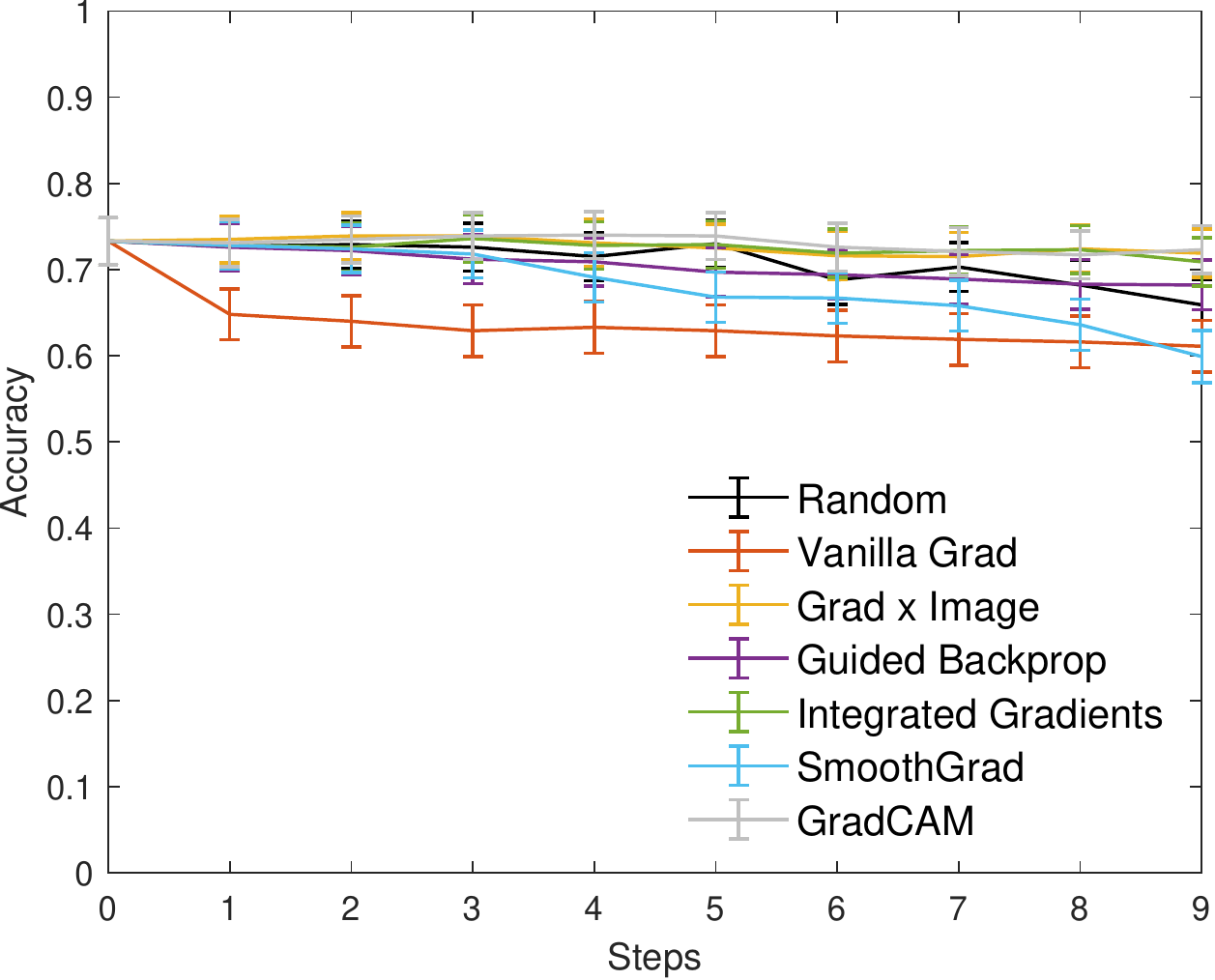}
         \caption{DermaMNIST ResNet}
         \label{fig:faith_det_derma}
     \end{subfigure}
     \hfill
     \begin{subfigure}[h]{0.31\textwidth}
         \centering
         \includegraphics[width=\textwidth]{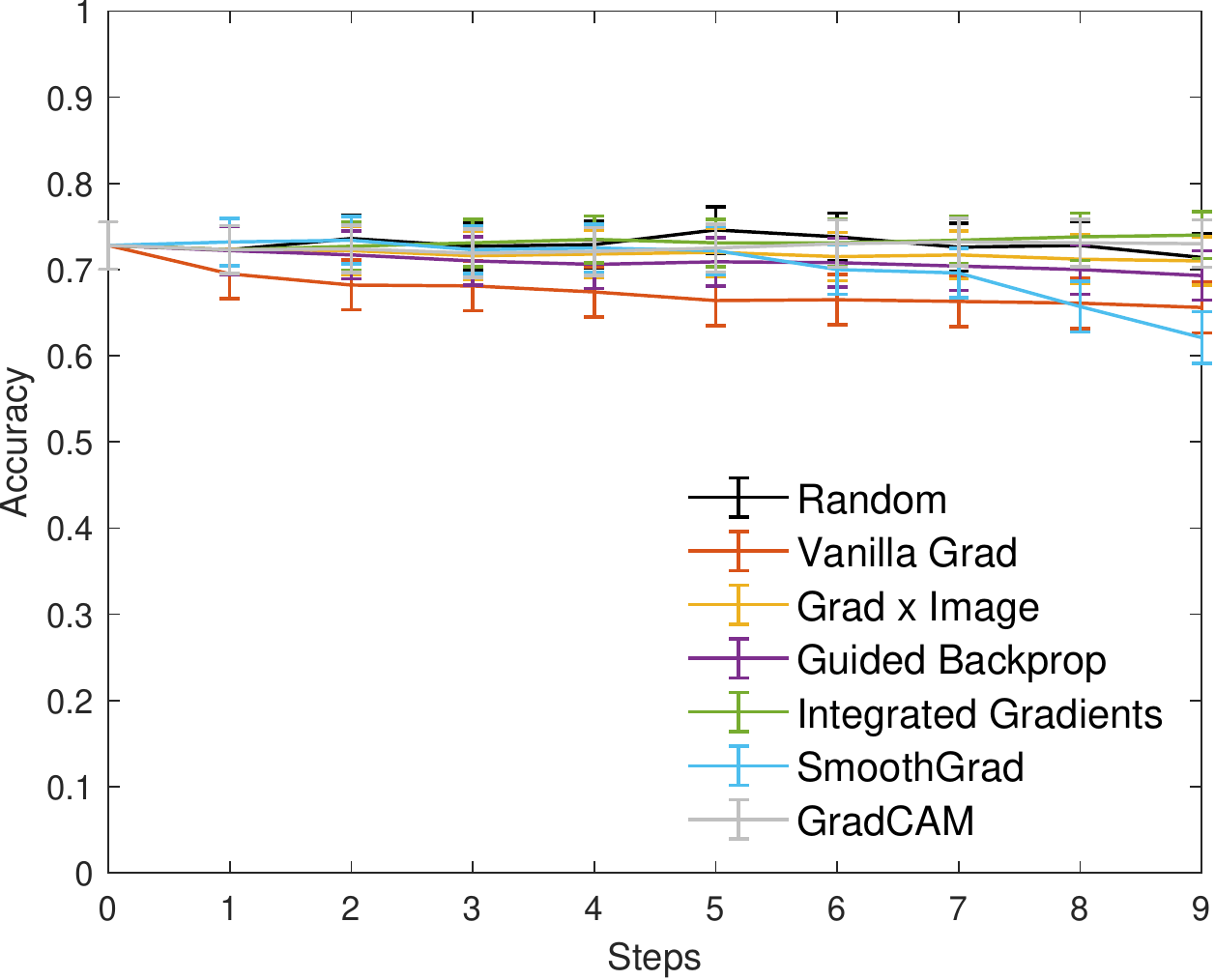}
         \caption{DermaMNIST Robust-ResNet}
         \label{fig:faith_rob_derma}
     \end{subfigure}
     \hfill
     \begin{subfigure}[h]{0.31\textwidth}
         \centering
         \includegraphics[width=\textwidth]{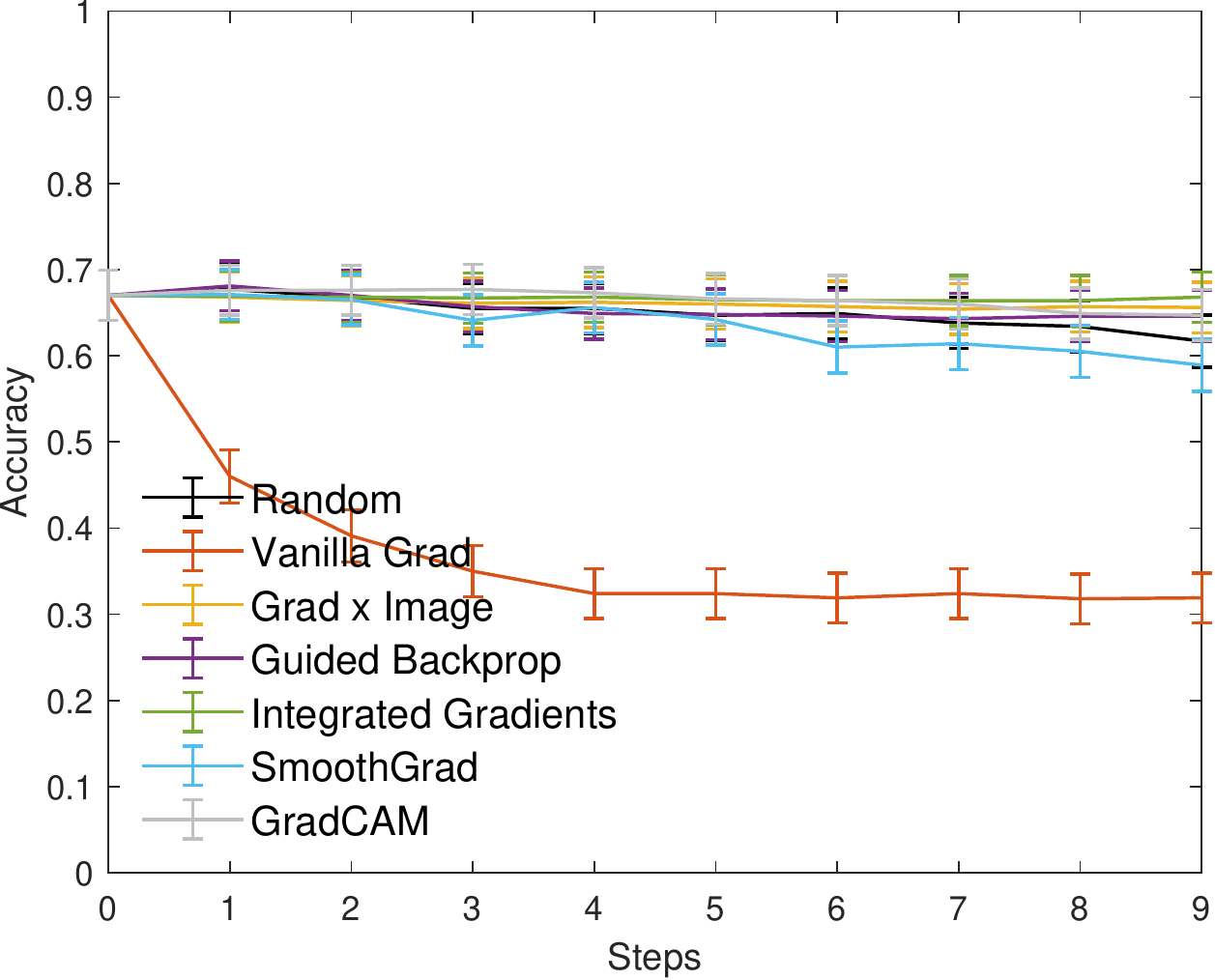}
         \caption{DermaMNIST VDP-CNN}
         \label{fig:faith_vdp_derma}
     \end{subfigure}
     \hfill
     \begin{subfigure}[h]{0.31\textwidth}
         \centering
         \includegraphics[width=\textwidth]{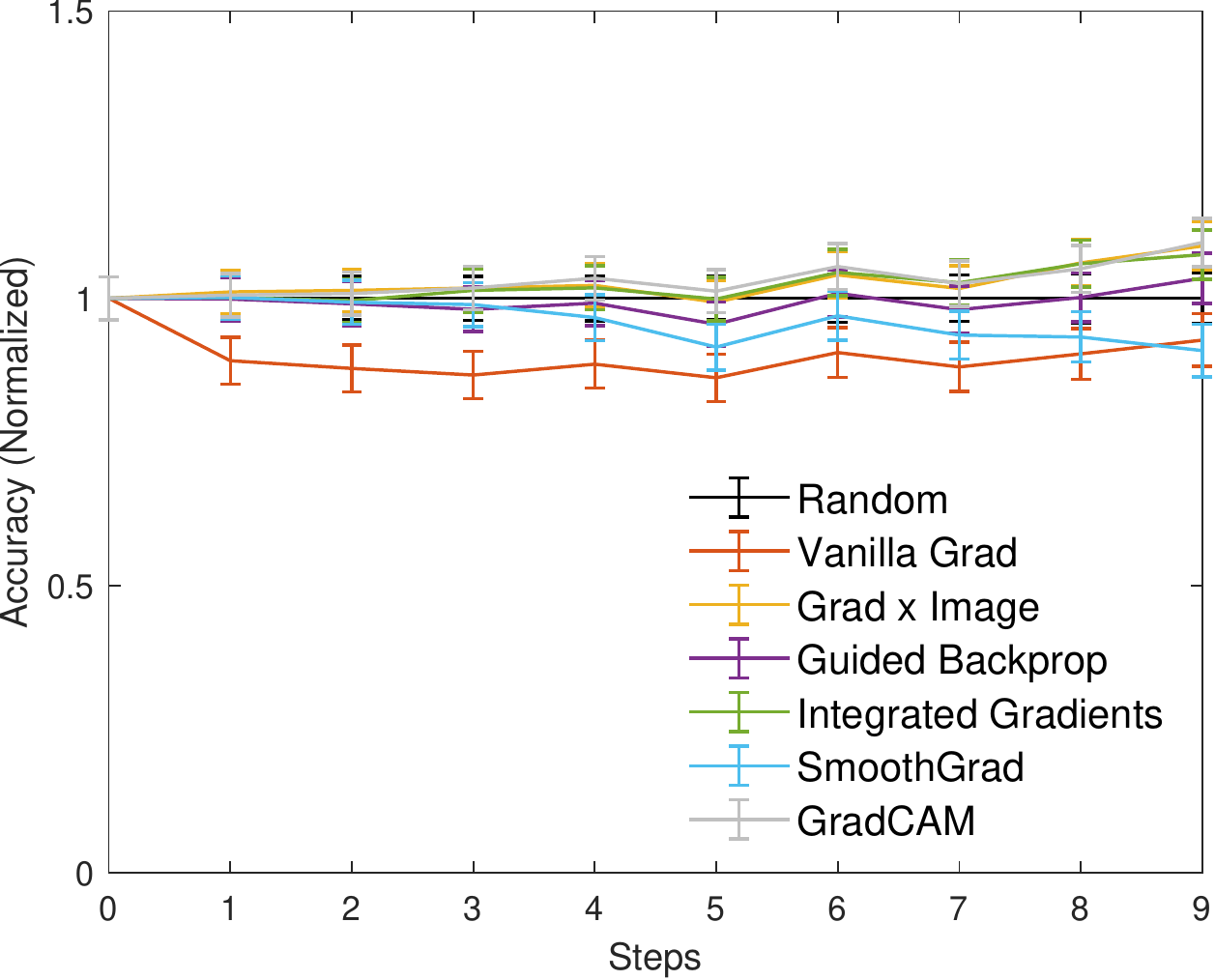}
         \caption{DermaMNIST Normalized ResNet}
         \label{fig:faith_det_norm_derma}
     \end{subfigure}
     \hfill
     \begin{subfigure}[h]{0.31\textwidth}
         \centering
         \includegraphics[width=\textwidth]{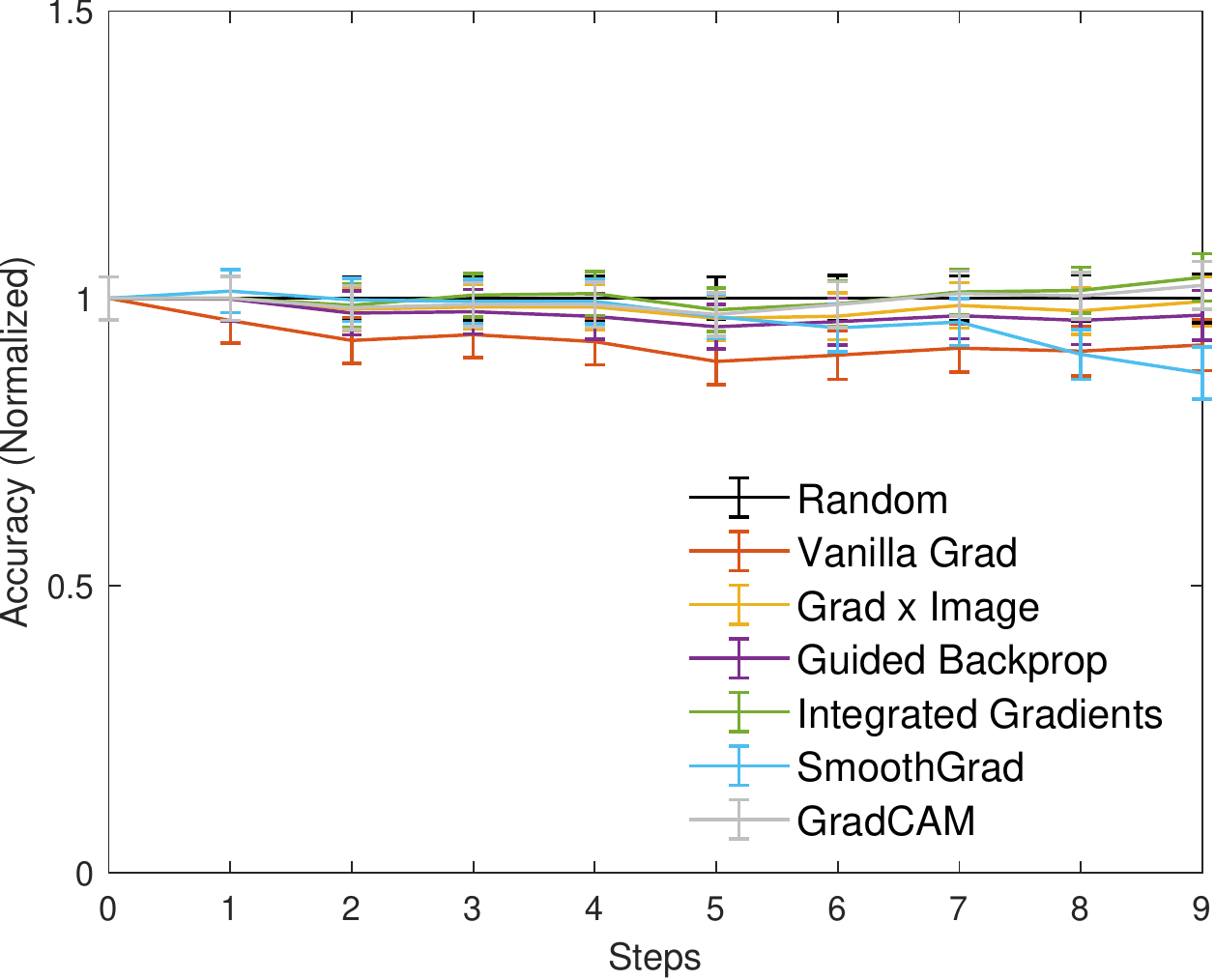}
         \caption{DermaMNIST Normalized Robust-ResNet}
         \label{fig:faith_rob_norm_derma}
     \end{subfigure}
     \hfill
     \begin{subfigure}[h]{0.31\textwidth}
         \centering
         \includegraphics[width=\textwidth]{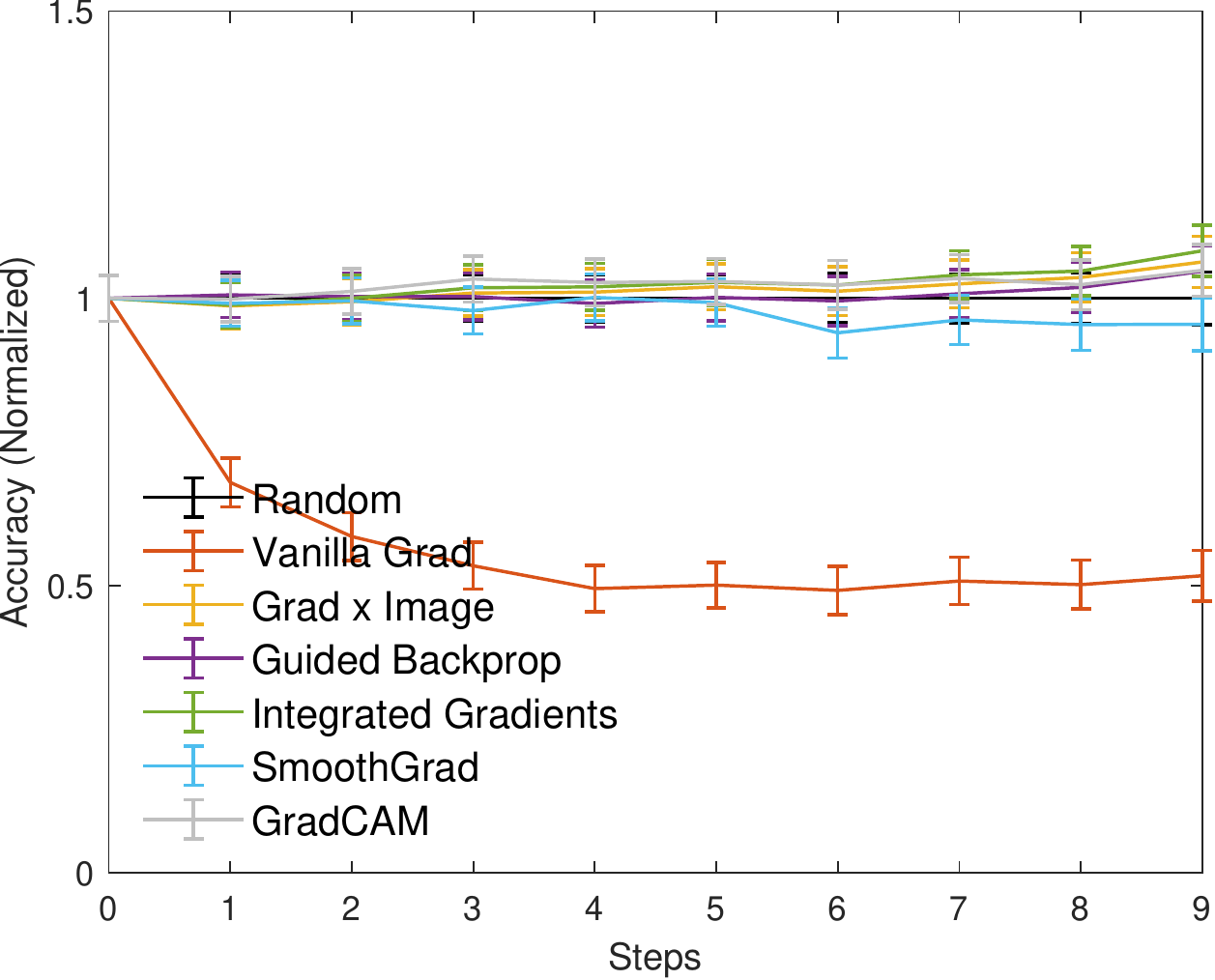}
         \caption{DermaMNIST Normalized VDP-CNN}
         \label{fig:faith_vdp_norm_derma}
    \end{subfigure}
    \hfill
    \begin{subfigure}[h]{0.31\textwidth}
         \centering
         \includegraphics[width=\textwidth]{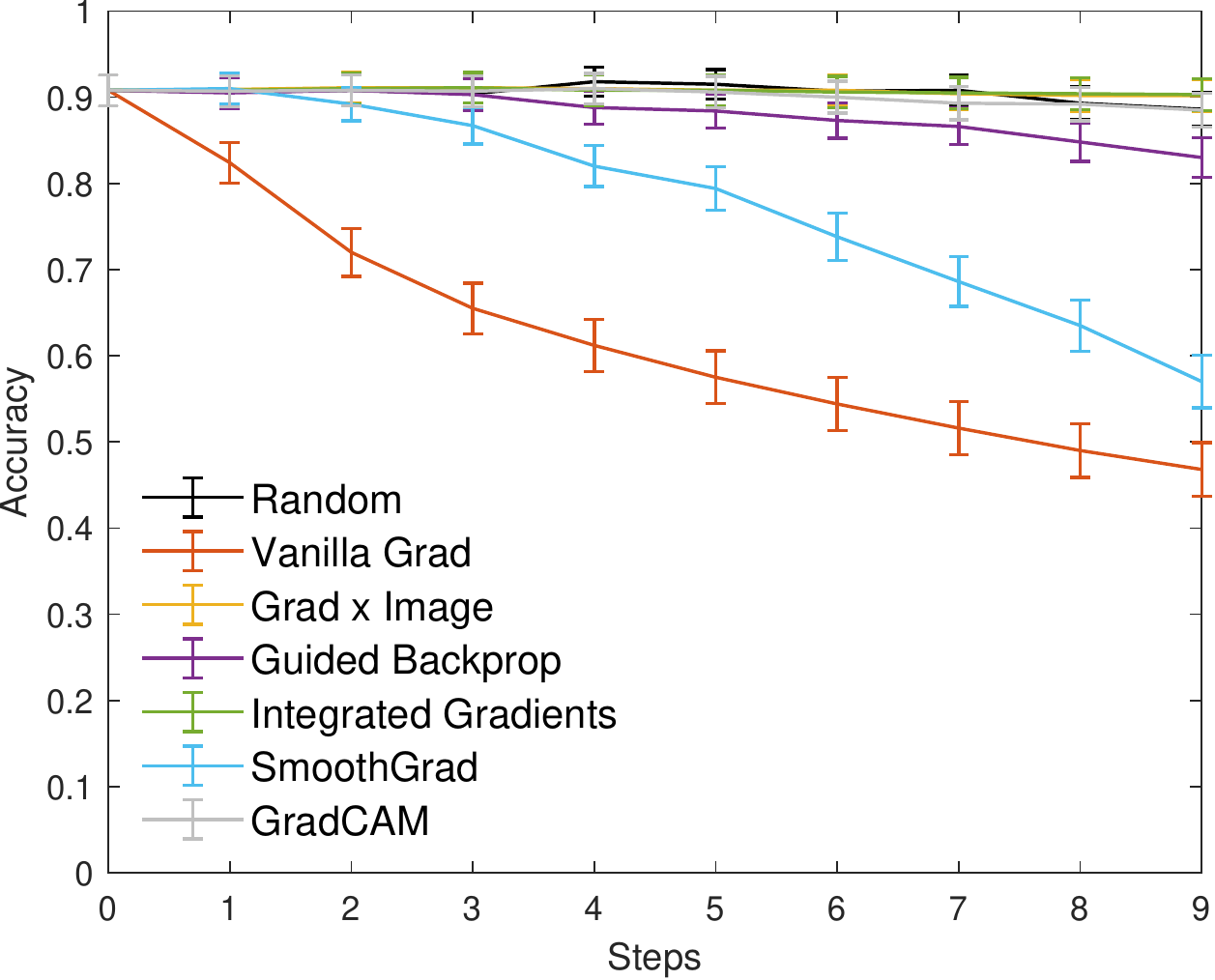}
         \caption{BloodMNIST ResNet}
         \label{fig:faith_det_blood}
     \end{subfigure}
     \hfill
     \begin{subfigure}[h]{0.31\textwidth}
         \centering
         \includegraphics[width=\textwidth]{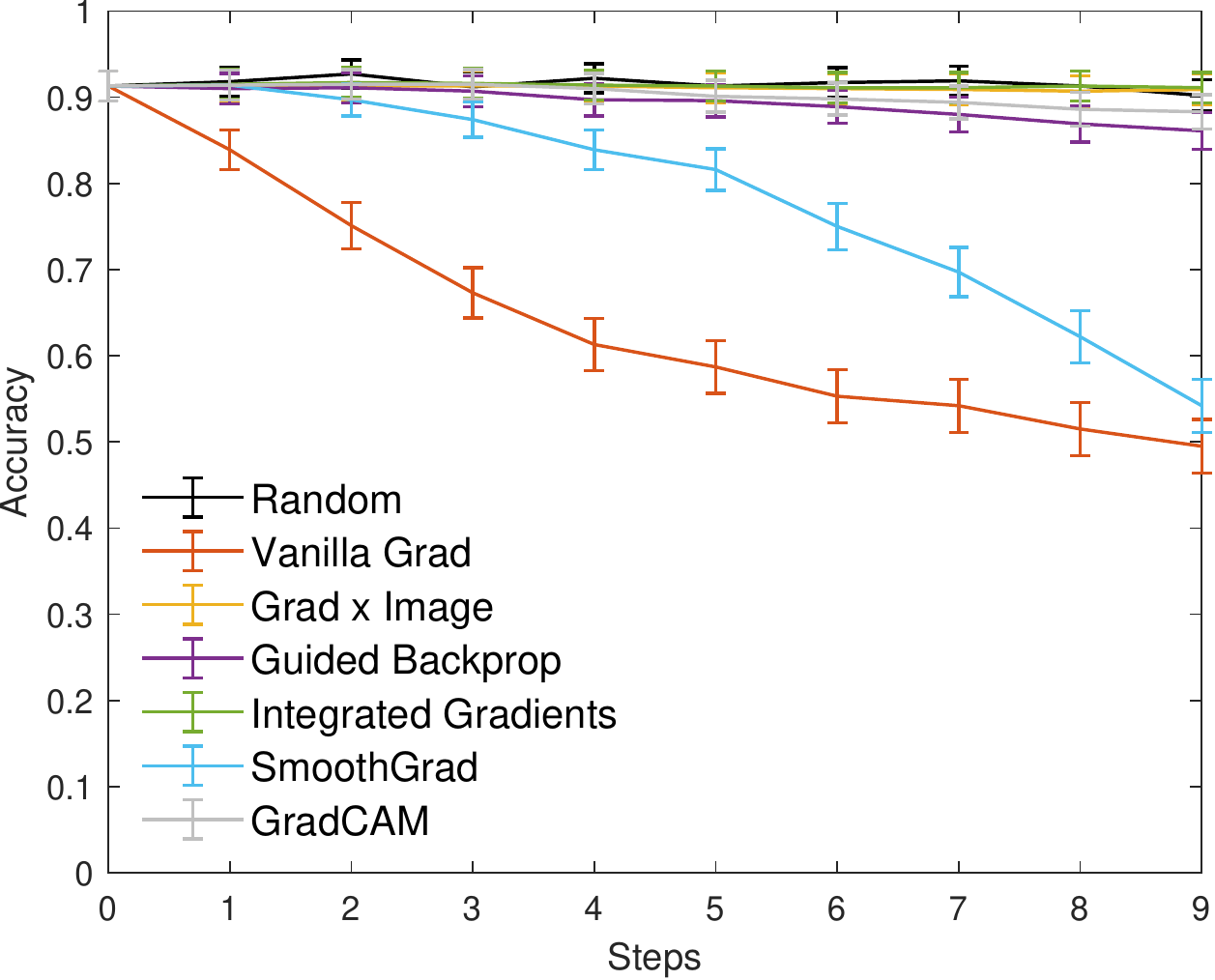}
         \caption{BloodMNIST Robust-ResNet}
         \label{fig:faith_rob_blood}
     \end{subfigure}
     \hfill
     \begin{subfigure}[h]{0.31\textwidth}
         \centering
         \includegraphics[width=\textwidth]{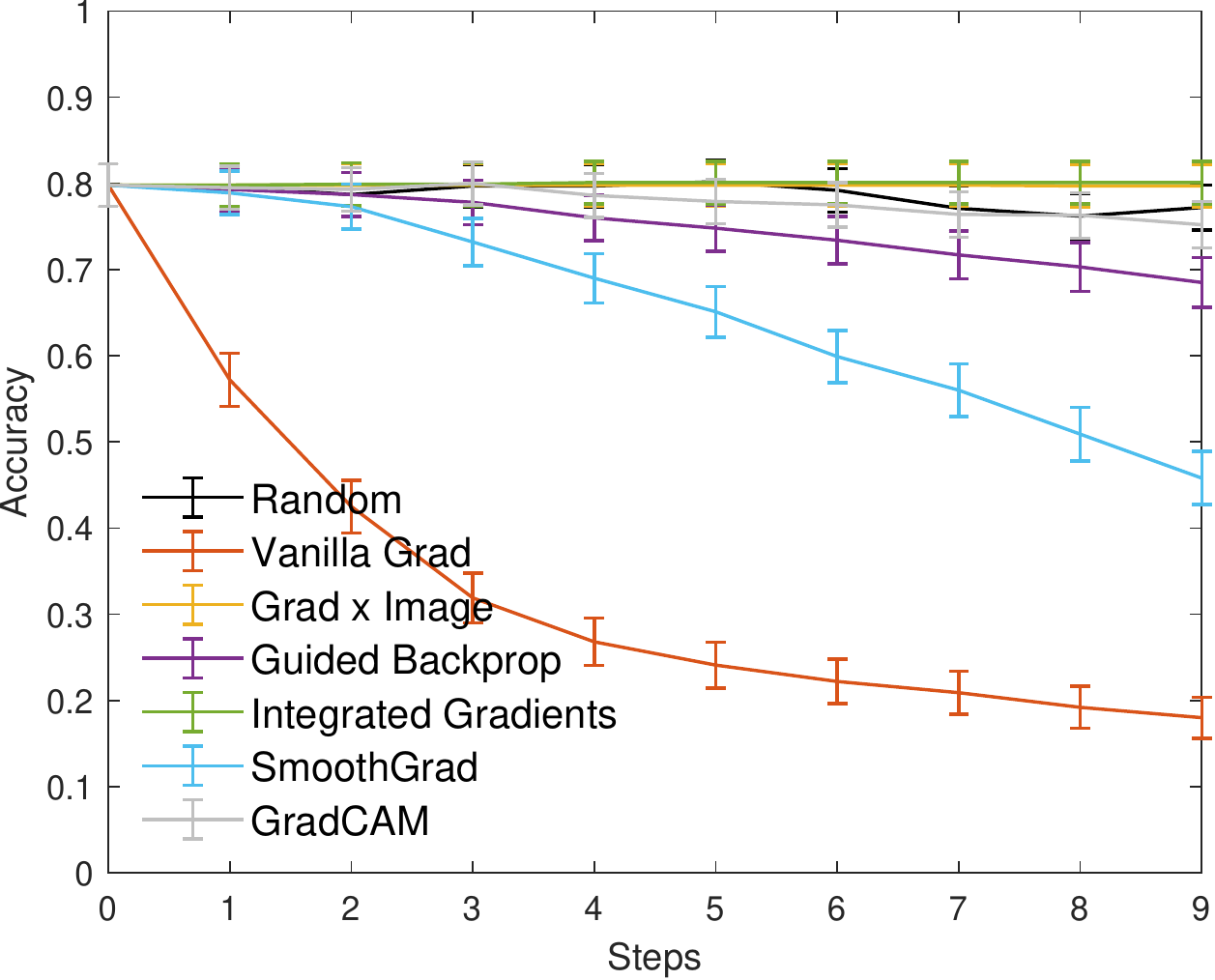}
         \caption{BloodMNIST VDP-CNN}
         \label{fig:faith_vdp_blood}
     \end{subfigure}
     \hfill
     \begin{subfigure}[h]{0.31\textwidth}
         \centering
         \includegraphics[width=\textwidth]{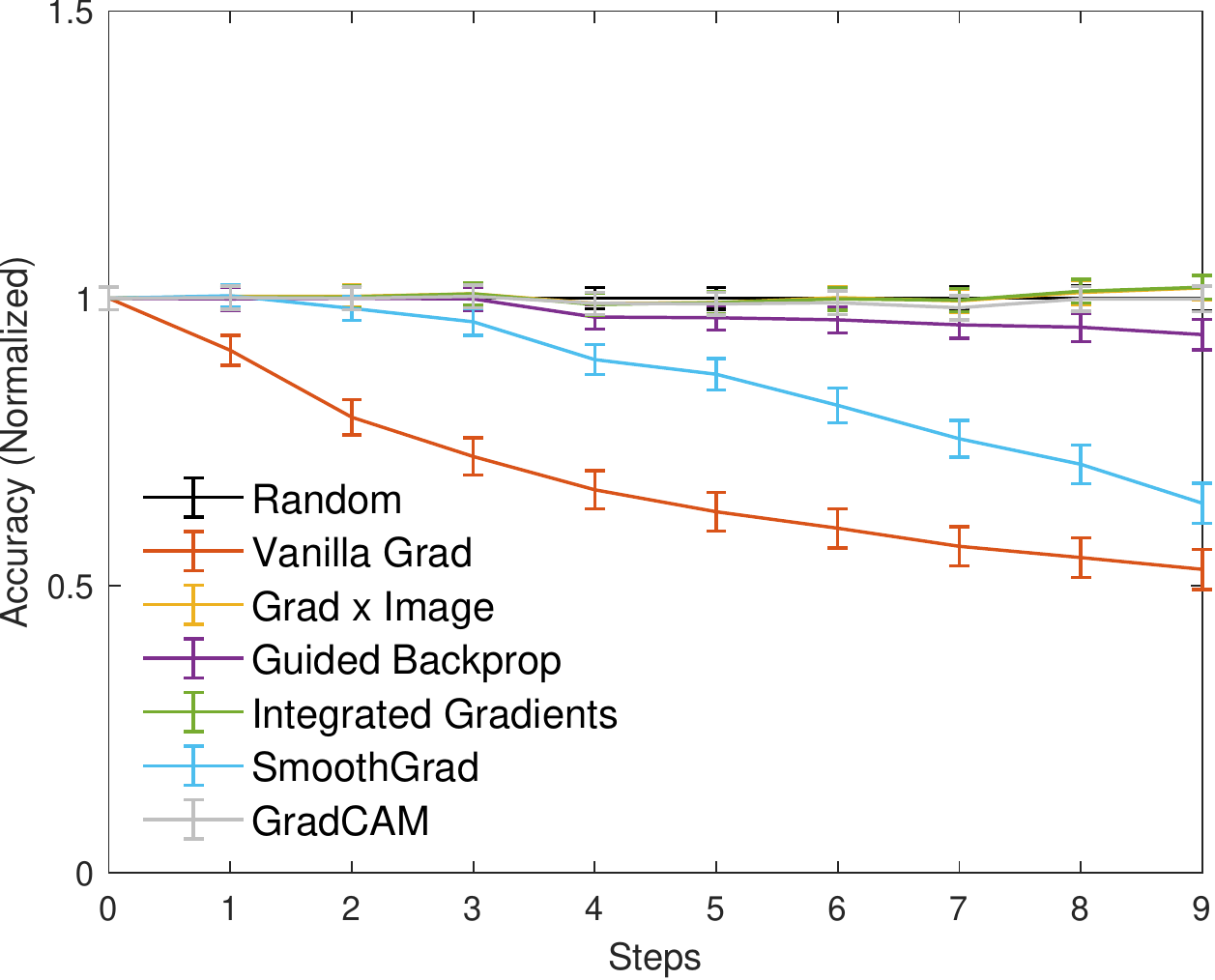}
         \caption{BloodMNIST Normalized ResNet}
         \label{fig:faith_det_norm_blood}
     \end{subfigure}
     \hfill
     \begin{subfigure}[h]{0.31\textwidth}
         \centering
         \includegraphics[width=\textwidth]{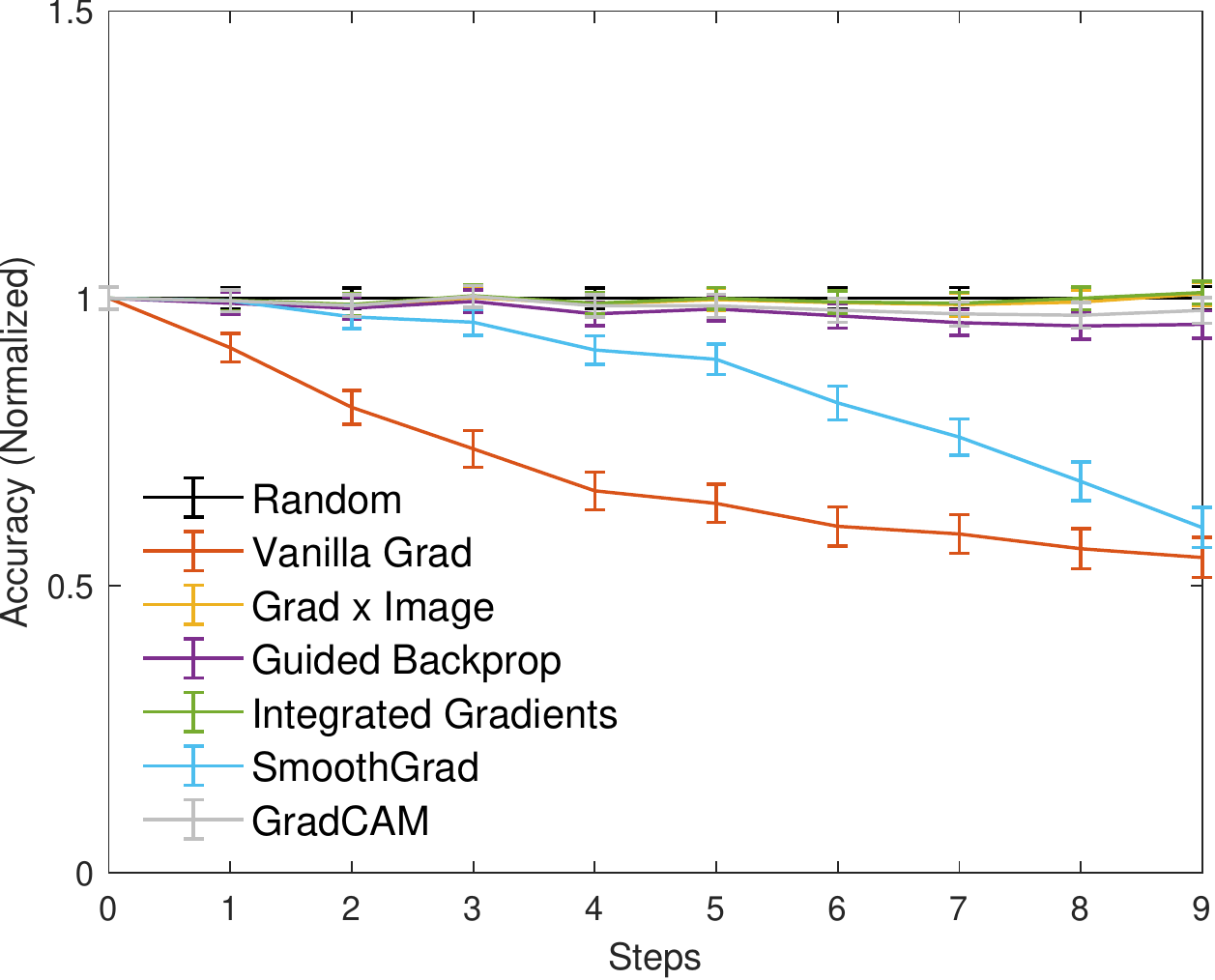}
         \caption{BloodMNIST Normalized Robust-ResNet}
         \label{fig:faith_rob_norm_blood}
     \end{subfigure}
     \hfill
     \begin{subfigure}[h]{0.31\textwidth}
         \centering
         \includegraphics[width=\textwidth]{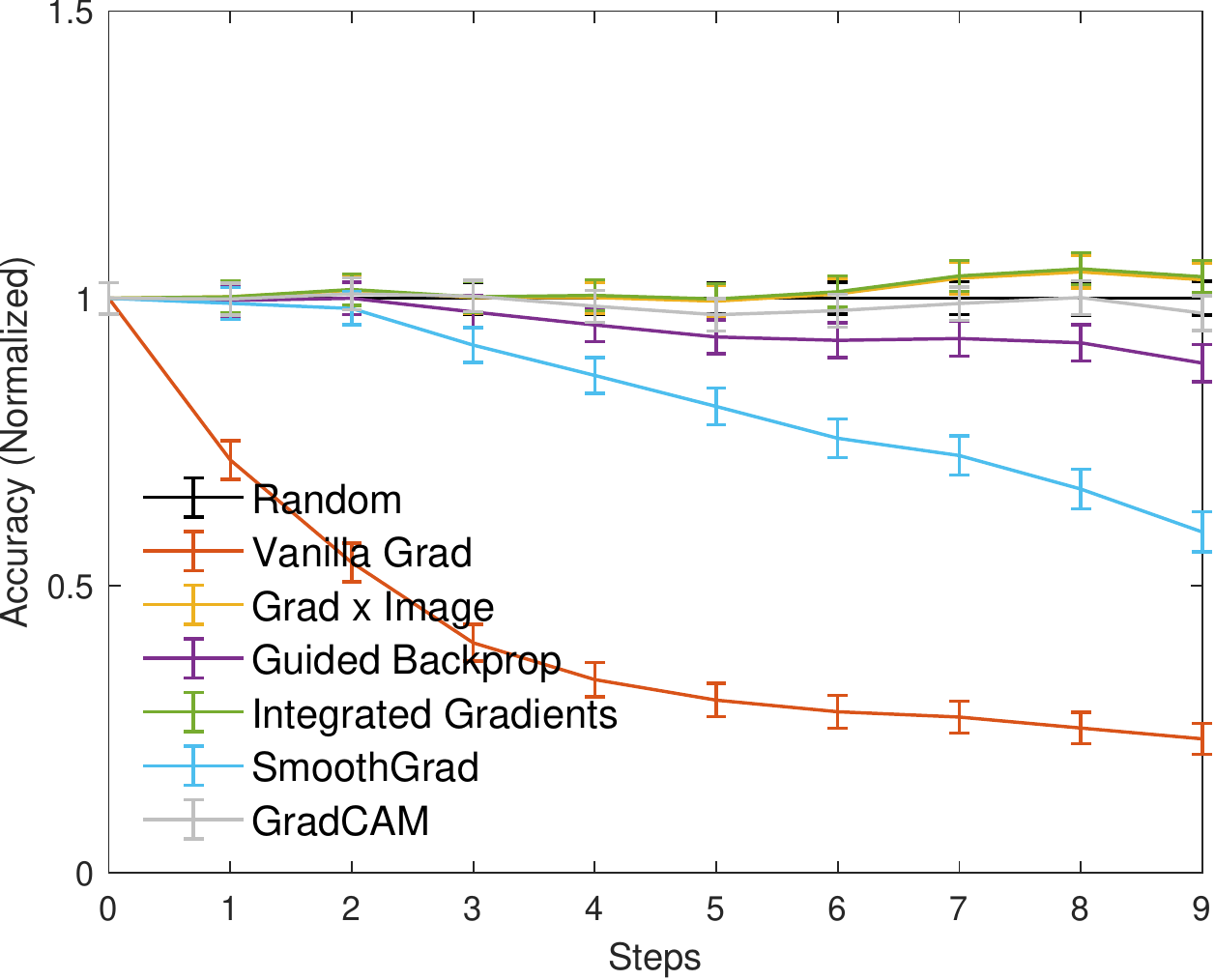}
         \caption{BloodMNIST Normalized VDP-CNN}
         \label{fig:faith_vdp_norm_blood}
     \end{subfigure}
        \caption{Results of the EvalAttAI metric tested using DermaMNIST and BloodMNIST on three models, ResNet18, Robust-ResNet18 and VDP-CNN. The y-axis of all sub-figures shows average accuracy as attributions are added to the input image, which is captured by the x-axis. Lower accuracy values correspond to the attribution method being more faithful. The $95\%$ confidence interval is shown in each plot using error bars. (a)-(c) and (g)-(i) Non-normalized accuracy. (d)-(f) and (j)-(l) Normalized accuracy using random as the baseline.}
        \label{fig:faith_plots2}
\end{figure*}

\subsection{E\lowercase{val}A\lowercase{tt}AI} 


Our method of evaluating the faithfulness of attributions takes a much different approach than Deletion and Insertion by not removing pixels at all. Our approach is to perturb pixels proportionally to the pixel scores of the attribution map. This is done by adding a scaled down attribution map to the original input image and then passing the new image through the machine learning model. The accuracy of the trained model before and after adding the perturbations are used to assess the faithfulness of each attribution method. We can see how EvalAttAI looks visually in comparison to Deletion and Insertion in Fig. \ref{fig:del_ins_evalattai_compared}. We expect that the more faithful the attribution method is, the more the accuracy will decrease after the perturbation. This implies that the maps which cause the steepest drop in accuracy are considered the most faithful. This is because highly faithful attribution maps will identify and perturb the most important pixels first, which is verified by the significant drop in the model accuracy. However, if unimportant pixels are perturbed, then the accuracy will not drop as much, perhaps not at all. This will indicate that the attribution map produced by the explainability method is less faithful. The experiments that we performed also evaluate a baseline map, which consists of random Gaussian noise with a standard deviation of $0.25$ and a mean of $0$.

The EvalAttAI method is described in Eq. \ref{eq:addattr}. In our experiments, the attribution map ($\boldsymbol{a}$) is multiplied by a scaling variable epsilon ($\varepsilon$), which we set to $0.1$, resulting in a scaled down attribution map. Starting with an image ($\boldsymbol{x}_s$), we add the scaled attribution, resulting in the modified image ($x_{s+1}$). This is done in an iterative fashion until the desired number of steps ($s$) are completed. The process always begins at $s = 0$ with the clean image ($\boldsymbol{x}_0$). Figure \ref{fig:faith_diagram} and Eq. \ref{eq:addattr} describe the process, which is repeated until the desired number of steps ($s$) are completed.

\begin{equation}
    \boldsymbol{x}_{s+1} = \boldsymbol{x}_{s} + \varepsilon * \boldsymbol{a}
\label{eq:addattr}
\end{equation}

EvalAttAI is sensitive and faithful because the most important pixels according to the attribution map will be perturbed the most. At the same time, the least important pixels, which have attribution scores closest to $0$, will not significantly alter the corresponding pixels on the input image. This method avoids the error introduced by Deletion and Insertion, which occurs due to pixels being entirely removed and replaced. Our method can also be tuned using $\epsilon$ so that pixels are altered in even smaller increments. The EvalAttAI method can be thought of as a more continuous approach to evaluating faithfulness.


\section{Results}
In this section, we present results to show how well various attribution evaluation methods including EvalAttAI perform. In order to ensure a fair comparison, we normalize the recorded output accuracy such that the random line (baseline method) is always equal to one. Any time that normalization is discussed in this paper, it is being used to explain how the results were formatted after they were collected. For instance, in Figs. \ref{fig:del_ins_results1}, \ref{fig:del_ins_results2}, \ref{fig:faith_plots1} and \ref{fig:faith_plots2}, we show the results before and after normalization of the data such that we can use the random line as a baseline across all models.

\begin{figure*}[htpb]
    \centering
    \centering
     \begin{subfigure}[h]{0.4\textwidth}
         \centering
         \includegraphics[width=\textwidth]{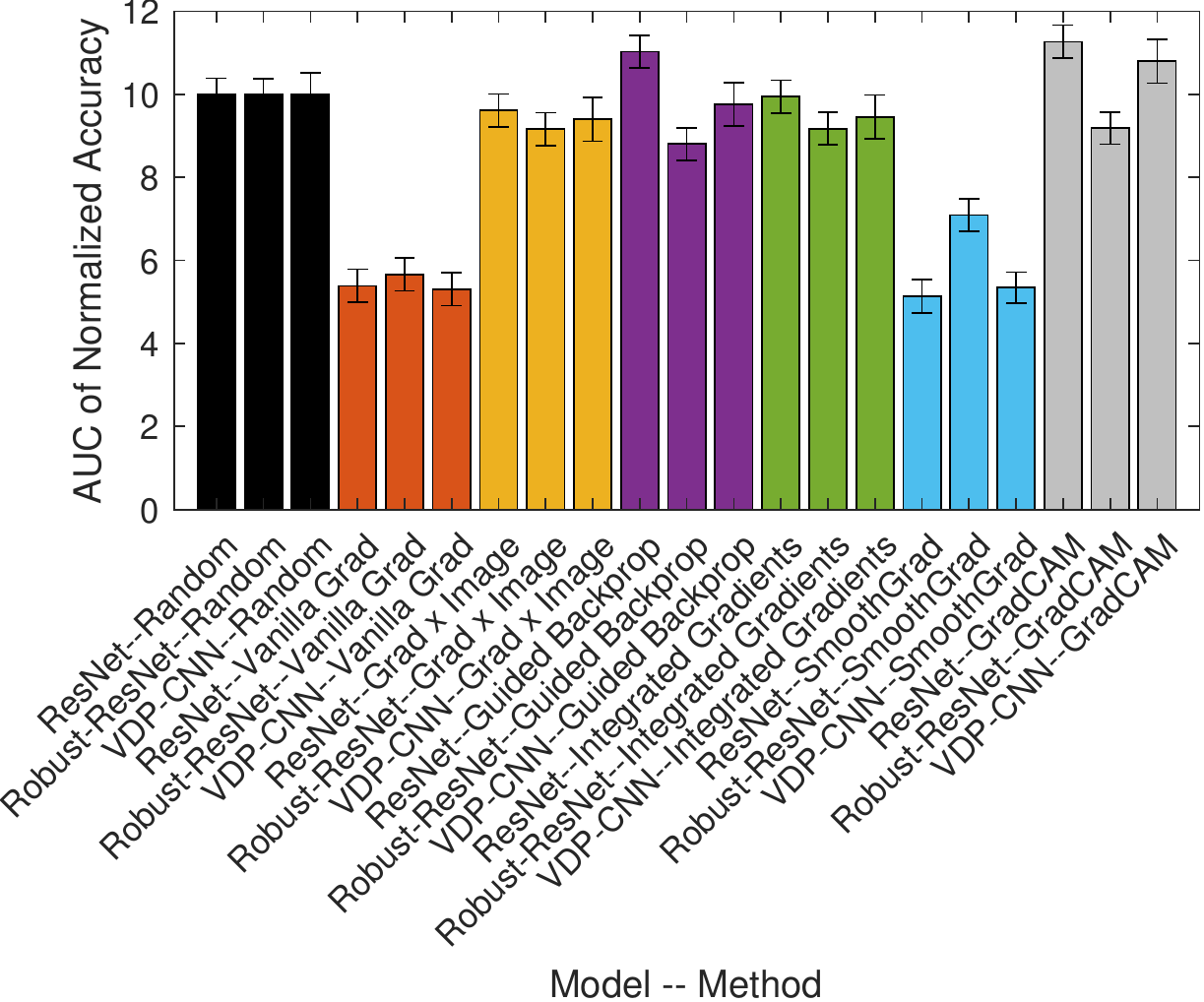}
         \caption{CIFAR10 dataset}
         \label{fig:faith_cifar}
     \end{subfigure}
     \hfill
     \begin{subfigure}[h]{0.4\textwidth}
         \centering
         \includegraphics[width=\textwidth]{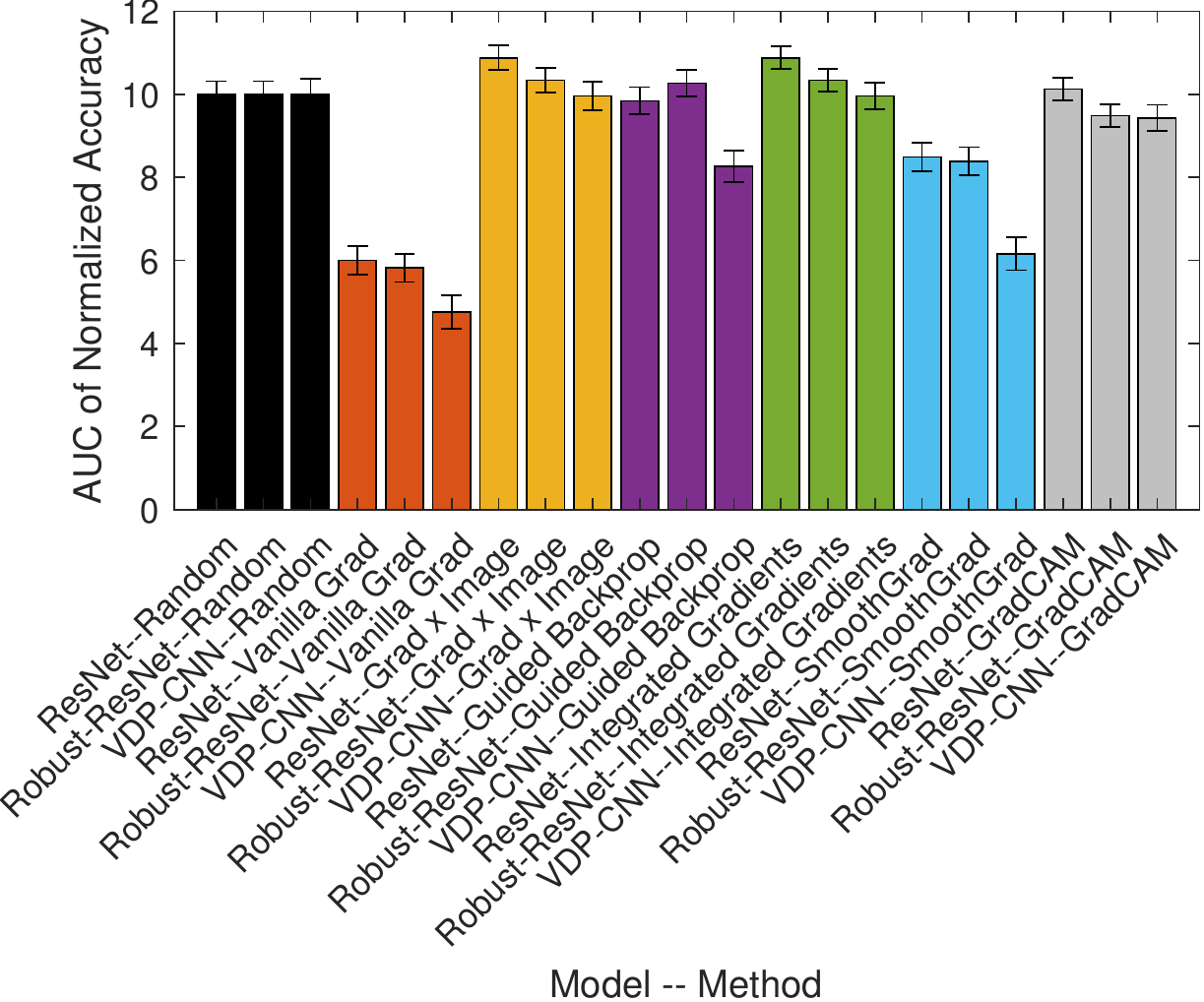}
         \caption{PathMNIST dataset}
         \label{fig:faith_path}
     \end{subfigure}
     \begin{subfigure}[h]{0.4\textwidth}
         \centering
         \includegraphics[width=\textwidth]{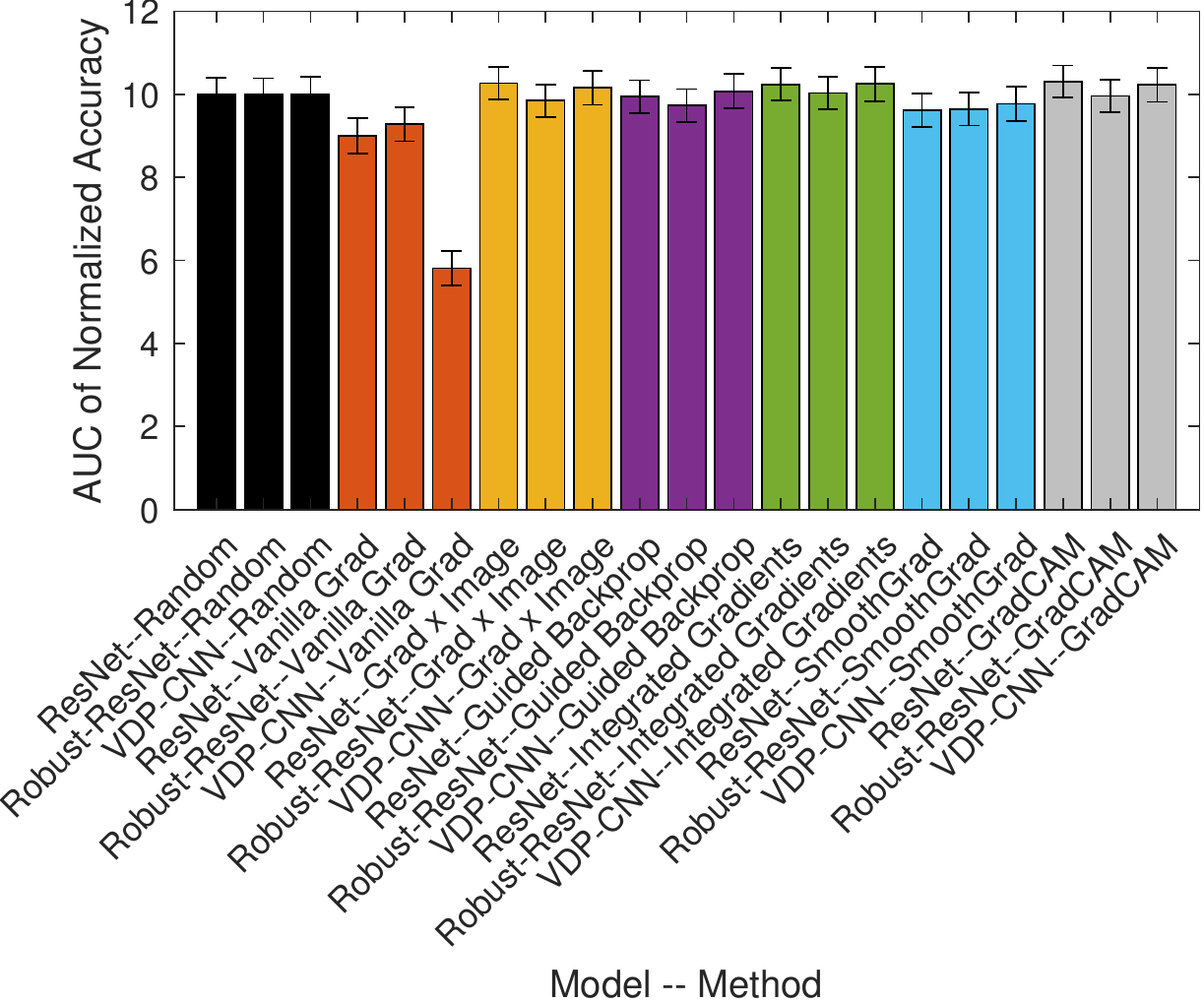}
         \caption{DermaMNIST dataset}
         \label{fig:faith_derma}
     \end{subfigure}
     \hfill
     \begin{subfigure}[h]{0.4\textwidth}
         \centering
         \includegraphics[width=\textwidth]{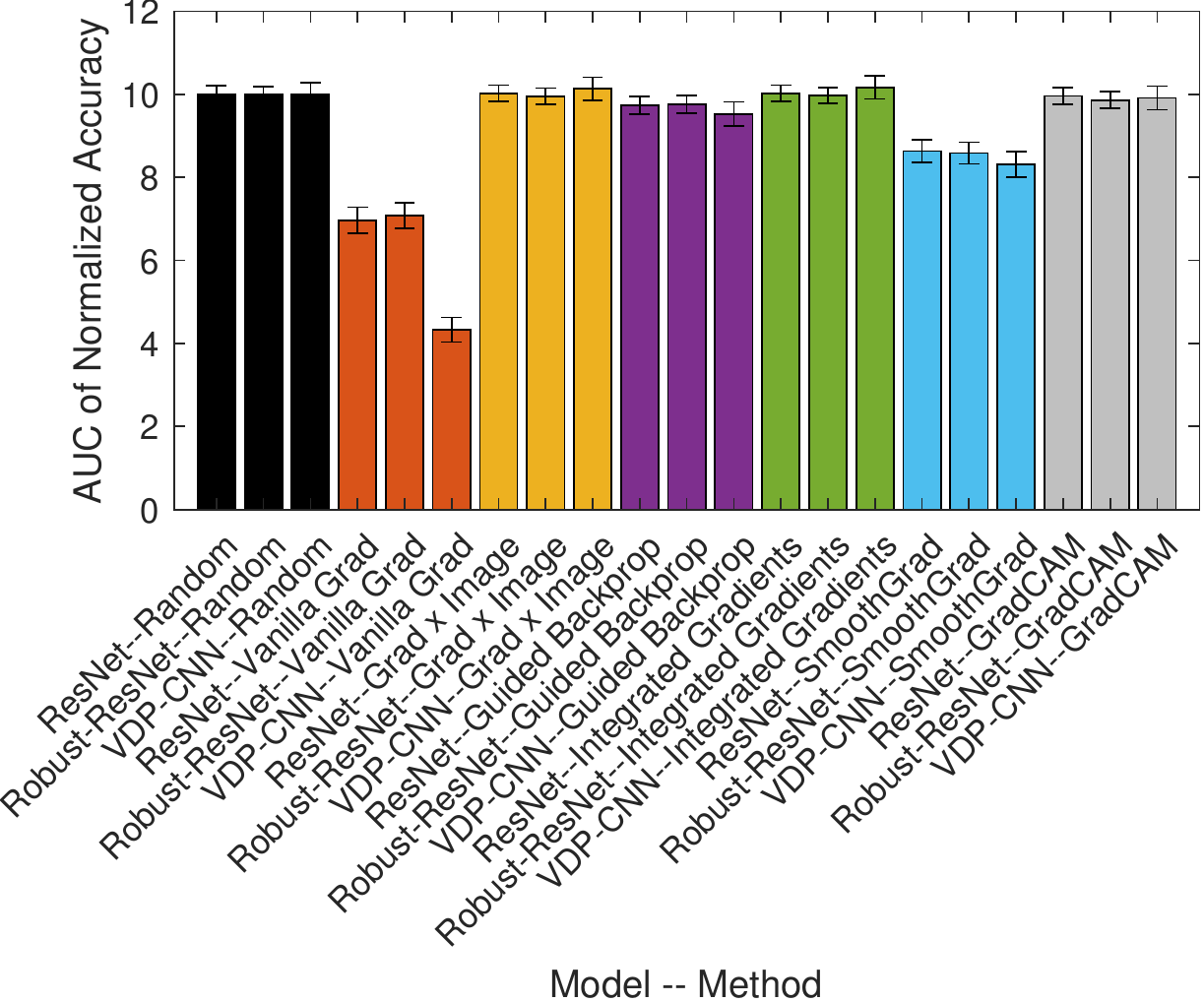}
         \caption{BloodMNIST dataset}
         \label{fig:faith_blood}
     \end{subfigure}
    \caption{Faithfulness results for EvalAttAI evaluation. Bar plots of the area under the curve (AUC) for the normalized CIFAR10, PathMNIST, DermaMNIST and BloodMNIST data (from MedMNIST dataset \cite{medmnistv1}\cite{medmnistv2}) results shown in Figs. \ref{fig:faith_plots1} and \ref{fig:faith_plots2}. Both subfigures include the 95\% confidence interval. The lower the bar, the more faithful the explanations.}
    \label{fig:faith_auc}
\end{figure*}

\subsection{Deletion and Insertion}

The goal of this first experiment is to quantify the behavior of Deletion and Insertion metrics for evaluating various types of attribution methods using three different types of trained models (ResNet, Robust-ResNet, and VDP-CNN). These models represent a non-robust model (ResNet), a robustly-trained model (Robust-ResNet), and a Bayesian model trained using the VDP technique (VDP-CNN). The test accuracy and normalized test accuracy are presented in Fig. \ref{fig:del_ins_results1}. The AUC of normalized test accuracy are presented in Fig. \ref{fig:del_ins_results2}. The error bars represent the $95\%$ confidence interval in both figures. 

The results for Deletion metric as presented in Fig. \ref{fig:del_ins_results1} ((a) and (c)) show each of the methods performing better than randomly removing pixels. We see this most clearly in Fig. \ref{fig:auc_del} where AUC is presented. The one exception to that is GradCAM, which performs worse (with statistical significance) than random. However, when we look at the Insertion results in Fig. \ref{fig:del_ins_results1} ((b) and (d)) and Fig. \ref{fig:auc_ins}, we see that GradCAM is the \textit{only} method which performs better than random (with statistical significance). In fact, when ranking the performance of the methods, we find that both metrics show contradictory behavior. Various attribution methods are ranked according to their performance and presented in Table \ref{fig:tables_eval}. This table clearly shows that the methods that are ranked best by the Deletion metric may be the worst when evaluated using the Insertion metric and vice versa.

\subsection{E\lowercase{val}A\lowercase{tt}AI}

The goal of this second experiment is two-fold. The first is to test the consistency and validity of the proposed EvalAttAI method. The second is to test whether the proposed methods capture the relationship between the faithfulness of the attribution method and the robustness of the model. All results with the EvalAttAI approach are presented in Figs. \ref{fig:faith_plots1}, \ref{fig:faith_plots2} and \ref{fig:faith_auc}.

In Figs. \ref{fig:faith_plots1} and \ref{fig:faith_plots2}, we note the slope of the accuracy drop for each attribution method. The attribution methods that cause the steepest drop, and therefore lowest AUC, are deemed to be the most faithful. The best performing methods can be clearly identified when looking at Fig. \ref{fig:faith_auc}. We can see that Vanilla Gradient and SmoothGrad are the only two that consistently perform better than random baseline, with statistical significance across all models. Other attribution methods, including  Grad x Image, Guided Backprop, Integrated Gradients, and GradCAM do not perform well as compared to the random baseline. We also observe that EvalAttAI produces consistent results for each attribution method across all models in most of the cases.

\begin{table}[htp]
    \caption{Rankings of various attribution methods for faithfulness evaluation based on AUC using CIFAR10 dataset and three different models. }
    \label{fig:tables_eval}
    \setlength{\tabcolsep}{3pt}
    \centering
    \begin{tabular}{rccc}
\hline \hline
    \multicolumn{1}{l}{}   & \multicolumn{1}{c|}{\hspace{0.5cm}ResNet\hspace{0.65cm} } & \multicolumn{1}{c|}{Robust-ResNet} & VDP-CNN \\ \hline
\hline
    \multicolumn{1}{l}{}        &  \multicolumn{1}{l}{}                          & 
    \textbf{Deletion}   &
    \multicolumn{1}{l}{} \\ 
\hline \hline
    \multicolumn{1}{r|}{1} & \multicolumn{1}{c|}{IG}     & \multicolumn{1}{c|}{GBP}           & IG                    \\ \hline
    \multicolumn{1}{r|}{2} & \multicolumn{1}{c|}{SG}     & \multicolumn{1}{c|}{IG}            & VG                    \\ \hline
    \multicolumn{1}{r|}{3} & \multicolumn{1}{c|}{IxG}    & \multicolumn{1}{c|}{SG}            & GBP                   \\ \hline
    \multicolumn{1}{r|}{4} & \multicolumn{1}{c|}{GBP}    & \multicolumn{1}{c|}{IxG}           & SG                    \\ \hline
    \multicolumn{1}{r|}{5} & \multicolumn{1}{c|}{VG}     & \multicolumn{1}{c|}{GC}            & IxG                   \\ \hline
    \multicolumn{1}{r|}{6} & \multicolumn{1}{c|}{GC}     & \multicolumn{1}{c|}{VG}            & GC                    \\ \hline
    \hline
    \multicolumn{1}{l}{}        & \multicolumn{1}{l}{}                          & \textbf{Insertion}   & \multicolumn{1}{l}{} \\ 
    \hline \hline
    \multicolumn{1}{r|}{1} & \multicolumn{1}{c|}{GC}     & \multicolumn{1}{c|}{GC}            & GC                    \\ \hline
    \multicolumn{1}{r|}{2} & \multicolumn{1}{c|}{VG}     & \multicolumn{1}{c|}{VG}            & IxG                   \\ \hline
    \multicolumn{1}{r|}{3} & \multicolumn{1}{c|}{SG}     & \multicolumn{1}{c|}{IxG}           & VG                    \\ \hline
    \multicolumn{1}{r|}{4} & \multicolumn{1}{c|}{GBP}    & \multicolumn{1}{c|}{SG}            & SG                    \\ \hline
    \multicolumn{1}{r|}{5} & \multicolumn{1}{c|}{IxG}    & \multicolumn{1}{c|}{GBP}           & IG                    \\ \hline
    \multicolumn{1}{r|}{6} & \multicolumn{1}{c|}{IG}     & \multicolumn{1}{c|}{IG}            & GBP                   \\ \hline
    \hline
    \multicolumn{1}{l}{}        & \multicolumn{1}{l}{}                          & \textbf{EvalAttAI}   & \multicolumn{1}{l}{} \\ 
    \hline \hline
    \multicolumn{1}{r|}{1} & \multicolumn{1}{c|}{SG}     & \multicolumn{1}{c|}{VG}            & VG                    \\ \hline
    \multicolumn{1}{r|}{2} & \multicolumn{1}{c|}{VG}     & \multicolumn{1}{c|}{SG}            & SG                    \\ \hline
    \multicolumn{1}{r|}{3} & \multicolumn{1}{c|}{IxG}    & \multicolumn{1}{c|}{GBP}           & GBP                   \\ \hline
    \multicolumn{1}{r|}{4} & \multicolumn{1}{c|}{IG}     & \multicolumn{1}{c|}{IxG}           & IxG                   \\ \hline
    \multicolumn{1}{r|}{5} & \multicolumn{1}{c|}{GBP}    & \multicolumn{1}{c|}{IG}            & IG                    \\ \hline
    \multicolumn{1}{r|}{6} & \multicolumn{1}{c|}{GC}     & \multicolumn{1}{c|}{GC}            & GC                    \\ \hline
    \multicolumn{4}{p{220pt}}{\small{Method names are represented as follows: IG = Integrated Gradient, SG = SmoothGrad, IxG = Input x Grad, GBP = Guided Backprop, VG = Vanilla Gradient, GC = GradCam.}}
    \end{tabular}
    \label{fig:table_EvalAttAI}
\end{table}

\section{Discussion}
We aimed to answer whether robust neural networks were more explainable, considering the potential usefulness of robust models and their visual explanations in medical imaging \cite{nielsen2022robust}. However, the question is difficult to answer due to three interrelated phenomena that need consideration. The first one is the robustness of the machine learning models, which is difficult to define and quantify \cite{dera2021premium, Dera:IEEERadar2019, Dera:MLSP2019, waqas2021exploring}. The second deals with various explainability methods and their internal complexities \cite{nielsen2022robust}. The third is related to metrics or measures employed to evaluate the plausibility and faithfulness of these explainability methods that try to explain the behavior of neural networks \cite{nielsen2022robust}. These three aspects of explainability research in deep neural networks are intimately intertwined. It may be challenging to disentangle these three (i.e., the robustness of models, internal dynamics of explainability methods, and metrics to evaluate this relationship) to understand the underlying relationship between robustness and explainability.

We started by limiting ourselves to two types of robust neural networks, (1) ResNet models trained using noisy datasets and (2) Bayesian deep neural networks trained using VDP technique. We acknowledge that the robustness of deep neural networks can be defined in many different ways, and consequently, many types of robust models can be built. However, we argue that these two methods represent a large class of robust models \cite{dera2021premium, carannante2020self, carannante2021trustworthy, Dera:MLSP2019, Dera:IEEERadar2019, ahmed2022failure}. On the other hand, for the explainability methods, we restricted ourselves to well-known gradient-based methods that are routinely used in image applications \cite{nielsen2022robust}. There are many different approaches to building explanations for elucidating the behavior of deep neural networks on test datasets. However, gradient-based methods represent a significantly large class of explainability methods. Finally, we wanted to use Insertion and Deletion metrics to evaluate the ``goodness'' or faithfulness of various attribution methods and quantify the explainability of robust neural networks. However, given the contradictory results produced by these two metrics due to their inherent nature, we proposed a new metric, EvalAttAI, to evaluate attribution methods.

Recently, Nourelahi et al. analyzed the faithfulness of various attribution methods for robust and non-robust CNNs \cite{nourelahi2022explainable}. The authors used Deletion and Insertion metrics for the evaluation of faithfulness. The faithfulness results appear to show methods and models which perform best on Deletion, perform the worst on Insertion, and vice versa \cite{nourelahi2022explainable}. We get similar results. Based on the working principles of these metrics (refer to Fig. \ref{fig:del_ins_explained}), the logical conclusion would be that both methods might quantify different things in an attribution method. Perhaps, Insertion and Deletion may not be capturing any helpful information, as evident in Figs. \ref{fig:del_ins_results1} and \ref{fig:del_ins_results2}. The contradictory findings for Deletion and Insertion are likely the result of an error introduced by removing and replacing the pixels, since the machine learning model is not trained on image data containing modified features (considering each pixel as a feature). We argue that given such a discrepancy, Deletion and Insertion may not be reliable metrics for evaluating faithfulness. The proposed EvalAttAI introduces perturbations smoothly and continuously, thus, avoiding abrupt changes in feature (pixel) values. These continuous and smooth changes are controlled using the $\varepsilon$ parameter as defined in Eq. \ref{eq:addattr}.  

A summary of our results is presented in Fig. \ref{fig:faith_auc} which compares three models (ResNet, Robust-ResNet, and Bayesian VDP-CNN) for four datasets (three of medical images and one of natural images), six attribution methods and a random baseline. In all sub-figures, the y-axes present AUC numbers calculated using test accuracy values for the proposed EvalAttAI metric and error bars represent $95\%$ confidence interval. We do not observe any significant trend showing that any model (among ResNet, Robust-ResNet, or VDP-CNN) is more explainable than others across all tested attribution methods and datasets. For some attribution methods, robust models are more explainable, but not for others. On the other hand, we note that Vanilla Grad (orange color bars) and SmoothGrad (blue color bars) consistently perform better than all other attribution methods. Amongst these two attribution methods,  we note that Bayesian neural networks (VDP-CNN) are more explainable. Based on these results, we can conclude that Bayesian CNNs (trained using the VDP technique \cite{dera2021premium}) are more explainable than the standard and robustly trained neural networks when attributions are generated using Vanilla Gradient.

We also observed that the Vanilla Gradient consistently performed better than all other methods on our evaluation metric. We argue that this is expected since this method works directly on the neural network without any alterations or modifications.

These findings expand upon the best practices presented in our previous work \cite{nielsen2022robust}, where we discussed considerations for the researchers when choosing an attribution method. We also discussed how robustness could play a significant role in generating plausible-looking explanations that may not be faithful. Based on this work, we suggest that the Vanilla Gradient should be used as the primary method of generating attributions. Instead of developing new ways to create visually appealing explanations using various operations, the community should focus on improving the robustness of the machine learning models using Bayesian approaches or other training methods.

\section{Conclusions}
In this work, we introduce a new metric for testing the faithfulness of attribution methods while showing the inconsistency and unreliability of the current state-of-the-art approaches. Our experiments are performed on both natural and medical image datasets. Our proposed faithfulness evaluation metric, EvalAttAI, shows consistent results. Our evaluation found that the Vanilla Gradient and SmoothGrad performed consistently better than all other attribution methods. We could not find compelling evidence that all robust models are more explainable across the board. However, our experiments consistently show that Bayesian CNNs (trained using the VDP framework) were more explainable than all other models when used with the best performing attribution method (Vanilla Gradient). 




\bibliographystyle{IEEEtran}
\bibliography{main}

\begin{IEEEbiography}
[{\includegraphics[width=1in, height=1.25in, clip, keepaspectratio]{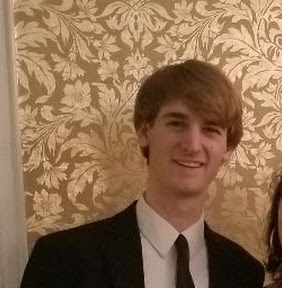}}]
{Ian E. Nielsen} is a Research Assistant, Teaching Assistant, and PhD candidate at Rowan University studying Electrical and Computer Engineering. He graduated Cum Laude and received his B.S. in Electrical and Computer Engineering at Rowan University in 2020. He is currently a recipient of the GAANN Teaching Fellowship under the U.S. Department of Education, as of January 2021. His current research is focused on robust machine learning and how it relates to explainable artificial intelligence. His work mainly focuses on computer vision and cancer diagnosis tasks. He conducts research as part of Rowan’s Artificial Intelligence Lab (RAIL). He currently coordinates a group of 19 undergraduate student researchers alongside Dr. Ravi Ramachandran though the Rowan University engineering clinic program. Since the start of his graduate education, he has taught machine learning using PyTorch and Python to dozens of students through this program. His tutorial on robust explainability was recently published in the IEEE Signal Processing Magazine.
\end{IEEEbiography}

\begin{IEEEbiography}
[{\includegraphics[width=1in, height=1.25in, clip, keepaspectratio]{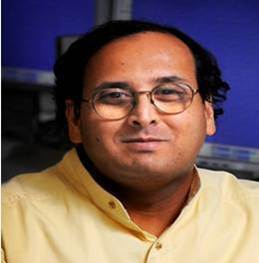}}]
{Ravi P. Ramachandran} (SM’08) received his B.Eng. degree (with great distinction) from Concordia University in 1984, his M.Eng. degree from McGill University in 1986 and his Ph.D. degree from McGill University in 1990. From October 1990 to December 1992, he worked at the Speech Research Department at AT\&T Bell Laboratories. From January 1993 to August 1997, he was a Research Assistant Professor at Rutgers University. He was also a Senior Speech Scientist at T-Netix from July 1996 to August 1997. Since September 1997, he is with the Department of Electrical and Computer Engineering at Rowan University, where he has been a Professor since September 2006. He has served as a consultant to T-Netix, Avenir Inc., Motorola and FocalCool. From September 2002 to September 2005, he was an Associate Editor for the IEEE Transactions on Speech and Audio Processing and was on the Speech Technical Committee for the IEEE Signal Processing society. From September 2000 to December 2015, he was on the Editorial Board of the IEEE Circuits and Systems Magazine.  Since May 2002, he has been on the Digital Signal Processing Technical Committee for the IEEE Circuits and Systems society. Since May 2012, he has been on the Education and Outreach Technical Committee for the IEEE Circuits and Systems Society. He is presently an Associate Editor for the journal Circuits, Systems and Signal Processing. His research interests are in digital signal processing, speech processing, biometrics, pattern recognition, machine learning and filter design.
\end{IEEEbiography}

\begin{IEEEbiography}
[{\includegraphics[width=1in, height=1.25in, clip, keepaspectratio]{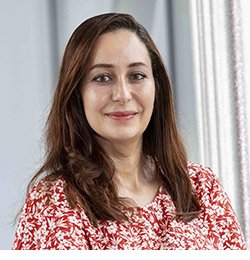}}]
{Nidhal Carla Bouaynaya} holds a Ph.D. in Electrical and Computer Engineering (ECE) and an M.S. in Pure Mathematics from the University of Illinois at Chicago. She is a Professor of ECE and the Director of Rowan’s Artificial Intelligence Lab (RAIL). She is currently serving as the Associate Dean for Research and Graduate Studies of the Henry M. Rowan College of Engineering. Previously, she was a faculty member with the University of Arkansas at Little Rock.

Her research interests are in big data analytics, machine learning, artificial intelligence and mathematical optimization. She co-authored more than 100 referred journal articles, book chapters and conference proceedings, such as IEEE Transactions on Pattern Analysis and Machine Intelligence, IEEE Signal Processing Letters, IEEE Signal Processing Magazine and PLOS Medicine. Dr. Bouaynaya won numerous Best Paper Awards, the most recent was at the 2019 IEEE International Workshop on Machine Learning for Signal Processing. She is also the winner of the Top algorithm at the 2016 Multinomial Brain Tumor Segmentation Challenge (BRATS). Dr. Bouaynaya has been honored with numerous research and teaching awards, including the Rowan Research Achievement Award in 2017 and The University of Arkansas at Little Rock Faculty Excellence Award in Research.
Her research is primarily funded by the National Science Foundation (NSF CCF, NSF ACI, NSF DUE, NSF I-Corps, NSF ECCS, NSF OAC and NSF HRD), The National Institutes of Health (NIH), US. Department of Education (USED), New Jersey Department of Transportation (NJ DoT), US. Department of Agriculture (USDA), the Federal Aviation Administration (FAA), Lockheed Martin Inc. and other industry. She is also interested in entrepreneurial endeavors. In 2017, she Co-founded and is Chief Executive Officer (CEO) of MRIMATH, LLC, a start-up company that uses artificial intelligence to improve patient oncology outcome and treatment response. MRIMath is funded by the NIH SBIR Program. 
\end{IEEEbiography}

\begin{IEEEbiography}
[{\includegraphics[width=1in, height=1.25in, clip, keepaspectratio]{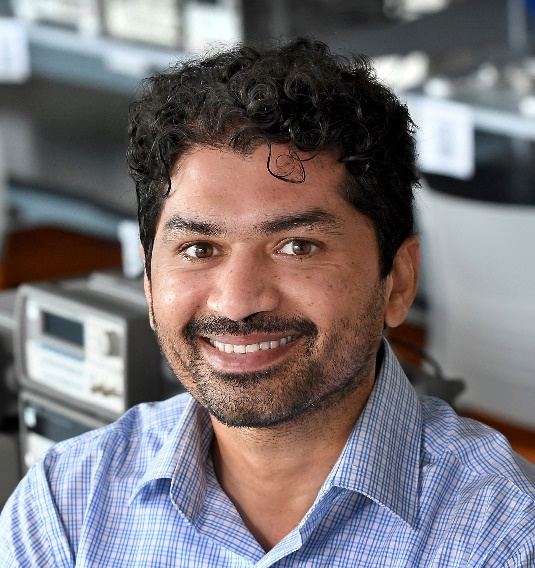}}]
{Ghulam Rasool} (M'2014) is an Assistant Member in the Department of Machine Learning at the H. Lee Moffitt Cancer Center \& Research Institute, Tampa, FL. He received a B.S. in Mechanical Engineering from the National University of Sciences and Technology (NUST), Pakistan, in 2000, an M.S. in Computer Engineering from the Center for Advanced Studies in Engineering (CASE), Pakistan, in 2010, and a Ph.D. in Systems Engineering from the University of Arkansas at Little Rock in 2014. He was a postdoctoral fellow with the Rehabilitation Institute of Chicago and Northwestern University from 2014 to 2016. Before joining Moffitt, he was an Assistant Professor at the Department of Electrical and Computer Engineering at Rowan University. His current research focuses on building trustworthy multimodal machine learning and artificial intelligence model for cancer diagnosis, treatment planning, and risk assessment. His research efforts are currently funded by two National Science Foundation awards (NSF) awards. Previously his research was supported by the National Institute of Health (NIH), U.S. Department of Education, NSF, the New Jersey Health Foundation (NJHF), Google, NVIDIA, and Lockheed Martin, Inc. His work on Bayesian machine learning won the Best Student Award at the 2019 IEEE Machine Learning for Signal Processing Workshop.
\end{IEEEbiography}


\end{document}